%% file: qixiang.tex
\documentclass[pmlr]{jmlr}% new name PMLR (Proceedings of Machine Learning Research)

\usepackage{longtable}% for long tables

\usepackage{booktabs}
\usepackage{lineno}
\usepackage{multirow}
\usepackage{enumitem}
\usepackage{tcolorbox}
\usepackage{graphicx}
\usepackage{pifont}
\usepackage{arydshln}
\definecolor{beaublue}{rgb}{1.0, 0.9, 0.9}
\definecolor{blackish}{rgb}{0.2, 0.2, 0.2}
\usepackage{hyperref}
\makeatletter
\let\Ginclude@graphics\@org@Ginclude@graphics 
\DeclareRobustCommand\onedot{\futurelet\@let@token\@onedot}
\def\@onedot{\ifx\@let@token.\else.\null\fi\xspace}

\input{definitions}

\def\eg{\emph{e.g}\onedot}

 \def\vs{\emph{vs}\onedot}

\makeatother

\jmlrvolume{260}
\jmlryear{2024}
\jmlrworkshop{ACML 2024}
\jmlrpages{-}
\renewcommand{\thepage}{}

\title[Motion meets Attention: Video Motion Prompts]{Motion meets Attention: Video Motion Prompts}

\author{\Name{Qixiang Chen} \Email{u7227010@anu.edu.au}\\
\addr Australian National University
\AND
\Name{Lei Wang\nametag{\thanks{Corresponding author.}}} \Email{lei.w@anu.edu.au}\\
\addr Australian National University \& Data61/CSIRO
\AND
\Name{Piotr Koniusz} \Email{piotr.koniusz@data61.csiro.au}\\
\addr Data61/CSIRO \& Australian National University
\AND
\Name{Tom Gedeon} \Email{tom.gedeon@curtin.edu.au}\\
\addr Curtin University
}
 
\editors{Vu Nguyen and Hsuan-Tien Lin}

\begin{document}

\maketitle

\begin{abstract}
Videos contain rich spatio-temporal information. 
Traditional methods for extracting motion, used in tasks such as action recognition, often rely on visual contents rather than precise motion features. This phenomenon is referred to as `blind motion extraction' behavior, which proves inefficient in capturing motions of interest due to a lack of motion-guided cues.
Recently, attention mechanisms have enhanced many computer vision tasks by effectively highlighting salient visual areas.
Inspired by this, we propose a modified Sigmoid function with learnable slope and shift parameters as an attention mechanism to modulate motion signals from frame differencing maps. This approach generates a sequence of attention maps that enhance the processing of motion-related video content.
To ensure temporal continuity and smoothness of the attention maps, we apply pair-wise \textit{temporal attention variation regularization} to remove unwanted motions (\eg, noise) while preserving important ones.
We then perform Hadamard product between each pair of attention maps and the original video frames to highlight the evolving motions of interest over time. These highlighted motions, termed \textit{video motion prompts}, are subsequently used as inputs to the model instead of the original video frames.
We formalize this process as a \textit{motion prompt layer} and incorporate the regularization term into the loss function to learn better motion prompts. This layer serves as an adapter between the model and the video data, bridging the gap between traditional `blind motion extraction' and the extraction of relevant motions of interest.
We show that our lightweight, plug-and-play motion prompt layer seamlessly integrates into models like SlowFast, X3D, and TimeSformer, enhancing performance on benchmarks such as FineGym and MPII Cooking 2. \href{https://q1xiangchen.github.io/motion-prompts/}{[\textbf{Project website}]} \href{https://github.com/q1xiangchen/VMPs}{[\textbf{Code}]} 
% Experimentally, we show that our plug-and-play motion prompt layer, despite its simplicity and lightweight nature, can be seamlessly integrated into existing architectures such as SlowFast, X3D and TimeSformer, achieving state-of-the-art performance on various benchmarks, including the large-scale FineGym and MPII Cooking 2 for fine-grained action recognition.
\end{abstract}


\section{Introduction}
\label{sec:intro}
Video-based research has gained popularity over the past several years due to its extensive applications in human-computer interaction, smart video surveillance, sports, and healthcare~\citep{lei_tip_2019,vidllmsurvey}. Videos contain rich information: spatially, they include visual contents such as objects, human subjects, and scene layouts; temporally, they show the dynamics of how these objects and humans interact and evolve over time.
Early works focused on using expert-designed, handcrafted descriptors to extract spatio-temporal information from videos~\citep{hof,sift_3d,3D-HOG,improved_traj}. However, while they were carefully designed, they could only handle simple contexts and were unable to generalize to other datasets even within the same domain. The major issue is that most descriptors do not focus on motions and heavily rely on visual contents. Compared to human vision systems in extracting information, they are now considered outdated, even though some are still in use~\citep{improved_traj}.

\begin{figure}[tbp]
\begin{center}
\includegraphics[trim=1.2cm 0.9cm 1.2cm 0.9cm, clip=true, width=\textwidth]{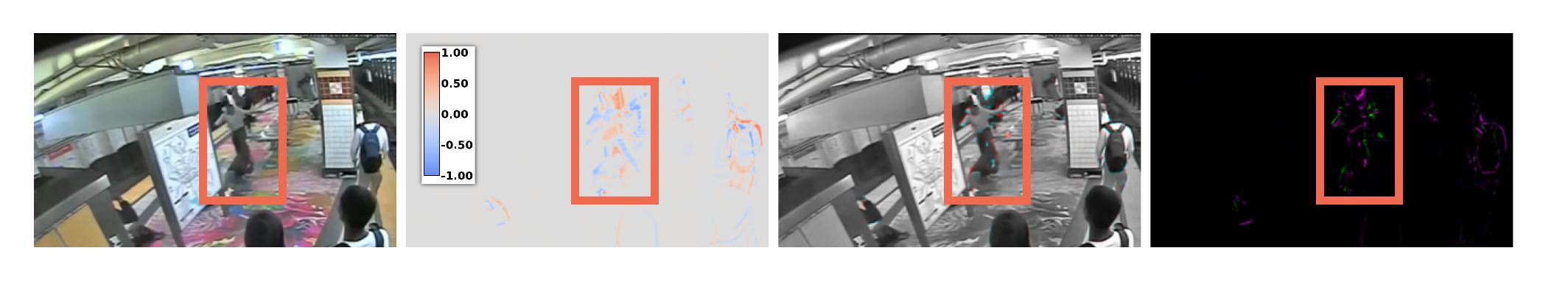}
\caption{Phenomena of `blind motion extraction'. 1st: Anomaly \textit{fighting} from UCF-Crime~\citep{sultani2018real}, highlighted by the orange bounding box. 2nd: Normalized frame differencing map, 3rd: Time-color reordering frame~\citep{kim2022capturing}, 4th: Taylor video frame~\citep{wang2024taylor}. These methods capture all motions without focusing on the anomaly. Non-motion-capturing methods focus on visual information, which is inefficient for motion-focused video processing.}\label{fig:motion-capture}
\end{center}
\vspace{-0.6cm}
\end{figure}

Deep learning has significantly advanced video-based research due to its end-to-end learnable nature~\citep{two_stream,spattemp_filters,spat_temp_resnet,i3d_net}. New architectures, such as CNNs, RNNs, and transformers~\citep{o2015introduction,sherstinsky2020fundamentals,vaswani2017attention}, along with components like normalization layers, skip connections, and dropout~\citep{ba2016layer,he2016deep,JMLR:v15:srivastava14a}, and strategies like large-scale pretraining and transfer learning~\citep{i3d_net,zhuang2020comprehensive}, have greatly supported the evolution of modern video processing techniques. The attention mechanism has recently joined convolutional layers, MLPs, and RNNs as a fundamental building block~\citep{vaswani2017attention}. Initially used within transformers in natural language processing, the attention mechanism is now effectively applied to image and video processing tasks~\citep{dosovitskiy2021an,arnab2021vivit}. It helps models focus on the most relevant parts of an image, relevant frames, spatial regions within frames, or significant scenes, thereby understanding both the visual contents and temporal dynamics.
However, there are several challenges with attention mechanisms in video processing~\citep{guo2022attention,brauwers2021general,NIU202148}: (i) they are computationally intensive due to the calculation of spatio-temporal attention weights, (ii) scalability issues when handling videos of varying lengths, and (iii) capturing temporal dependencies is challenging, as the model must focus not only on spatial features within frames but also on the temporal relationships between frames. Moreover, while attentions provide some level of interpretability by highlighting important regions or frames, interpreting why certain regions or frames receive higher attention can be difficult in complex video tasks. Additionally, attention mechanisms must be robust to variations in video quality, lighting, occlusions, and other environmental factors. Despite large-scale datasets allowing attention-driven models to capture a wide range of spatio-temporal patterns, their generalization to unseen video data remains challenging.

\begin{figure}[tbp]
\begin{center}
\includegraphics[width=\textwidth]{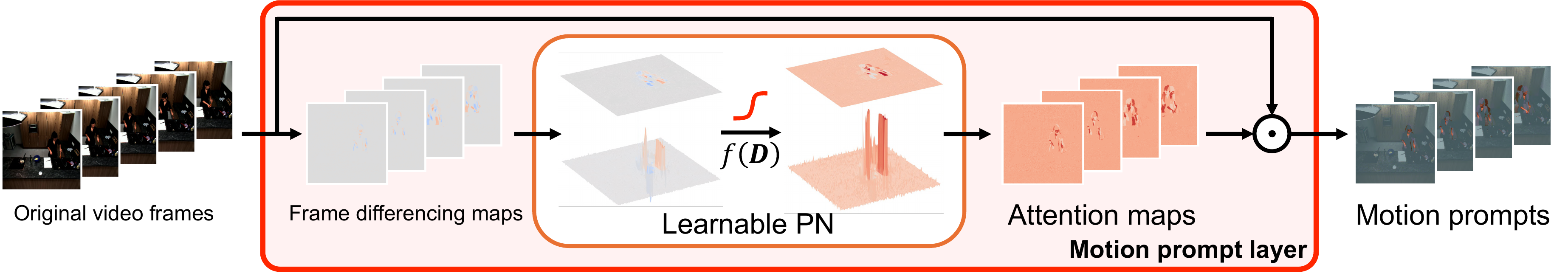}
\caption{Overview of the motion prompt layer. Learnable Power Normalization (PN) function $f(\cdot)$ modulates motion, influencing how motion is enhanced or dampened in each frame differencing map $\mathbf{D}$ to highlight relevant movements. The resulting attention maps are multiplied element-wise ($\odot$) with original video frame to produce video motion prompts. We introduce a temporal attention variation regularization term (see Sec.~\ref{sec:motion-prompt}) for smoother attention maps, ensuring better motion prompts. This layer can be inserted between the video input and backbones such as TimeSformer, serving as an adapter. Training involves optimizing both the motion prompt layer and the backbone network using a generic loss function, \eg, cross-entropy, along with the new regularization term.}
\label{fig:main-pipeline}
\end{center}
\vspace{-0.6cm}
\end{figure}

Unlike the attention mechanism, prompt engineering involves designing prompts to guide language models in producing desired responses~\citep{brown2020language,radford2021learning,kim2021vilt,zhu2021deformable}. A well-designed prompt provides instructions or contextual clues that direct the model's output~\citep{chen2023unleashing}. This allows the attention mechanism to effectively allocate focus to the relevant parts of the input, especially in tasks where there may be many frames and objects, not all of which are relevant. This highlights the potential of using prompts in video processing tasks, where identifying the relevant motions can be challenging.
In this paper, we establish a strong connection among video motion, attention and prompt engineering. We introduce the concept of motion prompts to address challenges related to efficiency, interpretability, and generalizability. We use a modified Sigmoid function with learnable slope and shift parameters as a power normalization function\footnote{Technically, ours is not a true power normalization (PN), as PN equals 0 for a 0 input. Instead, ours is a shift-enabled, PN-inspired function.} to activate the motions: the slope controls the strength of modulation, determining whether to enhance or dampen the motions, and the shift acts as a threshold. The motions to be activated and modulated are represented as a sequence of frame differencing maps computed between consecutive frames.
The activated per-frame motions can be viewed as an attention map, due to the enhancement of motion regions both spatially and temporally, as these regions evolve over time. To ensure that the generated attention maps are spatio-temporally smooth and continuous, we introduce a pair-wise temporal attention variation regularization to remove unwanted motions such as noise.
We then perform the Hadamard product between each pair of attention maps and the corresponding original video frame, resulting in a sequence of highlighted video sequences, which we refer to as video motion prompts. Thus, our motion prompts are motion-dependent, rather than being dependent solely on the dataset.
We formalize this process as a motion prompt layer, a plug-and-play component added between the model and the input data, bridging the gap between traditional, `blind motion extraction' (see Fig.~\ref{fig:motion-capture}) and the extraction of relevant motions. We show that, with only two additional learnable parameters, our motion prompt layer significantly enhances action recognition.
Our contributions are summarized as follows:
\renewcommand{\labelenumi}{\roman{enumi}.}
\begin{enumerate}[leftmargin=0.6cm]
\item We introduce video motion prompts, defined as a sequence of spatio-temporally highlighted video frames. We format the extraction of motion prompts as a plug-and-play layer that can be inserted between video data input and the video model architecture, functioning as an adapter. This adapter bridges the gap between `blind motion extraction' and the extraction of motions of interest.
\item Our motion prompts are prompt-inspired, motion-dependent, and attention-driven. To generate smooth and continuous attention maps, we introduce a temporal attention variation regularization term. This term removes unwanted motions and enhances the model's generalization ability. We incorporate this regularization term into the loss function to improve the learning of motion prompts.
\item Experimentally, we demonstrate that our motion prompt layer, despite its simplicity, can be integrated into popular video models to achieve state-of-the-art performance in tasks such as generic action recognition and fine-grained action recognition.
\end{enumerate}

% Sec. \ref{sec:related} introduces the background. Sec. \ref{sec:approach} and \ref{sec:exp} present our method and results.

In Appendix~\ref{sec:related}, we review closely related work on motion extraction, attention mechanisms, prompts, and adapter layers for video processing. We also highlight the significant differences of our approach compared to these works. Hereafter, we introduce our approach.

\section{Approach}
\label{sec:approach}

First, we present our notation. An overview of our motion prompt layer is provided in Fig.~\ref{fig:main-pipeline}. In Appendix~\ref{sec:appl-pre}, we provide preliminary information on activation functions, a mathematical view of attention mechanisms, and the power normalization family. 

\noindent\textbf{Notation.} 
Let $\idx{T}$ stand for the index set $\{1, 2, \dots, T\}$. 
Scalars are in regular fonts; 
vectors are denoted by lowercase boldface letters, \eg, $\vx$; 
matrices by uppercase boldface, \eg, \boldsymbol{$\mX$}; tensors by calligraphic letters, \eg, $\tX$. 
Let $\tX\in\mbr{d_1\times d_2\times d_3}$ denote a third-order tensor, using the Matlab convention, we refer to its $t$-th slice as $\tX_{:,:,t}$, which is a $d_1\!\times\!d_2$ matrix.

\subsection{Learnable Power Normalization}
\label{sec:learnable-pn}

\begin{figure*}[tbp]
    \centering
    \subfigure[PNs \vs ours.]{\includegraphics[trim=0.35cm 0cm 0.2cm 0.0cm, clip=true, height=2.25cm]{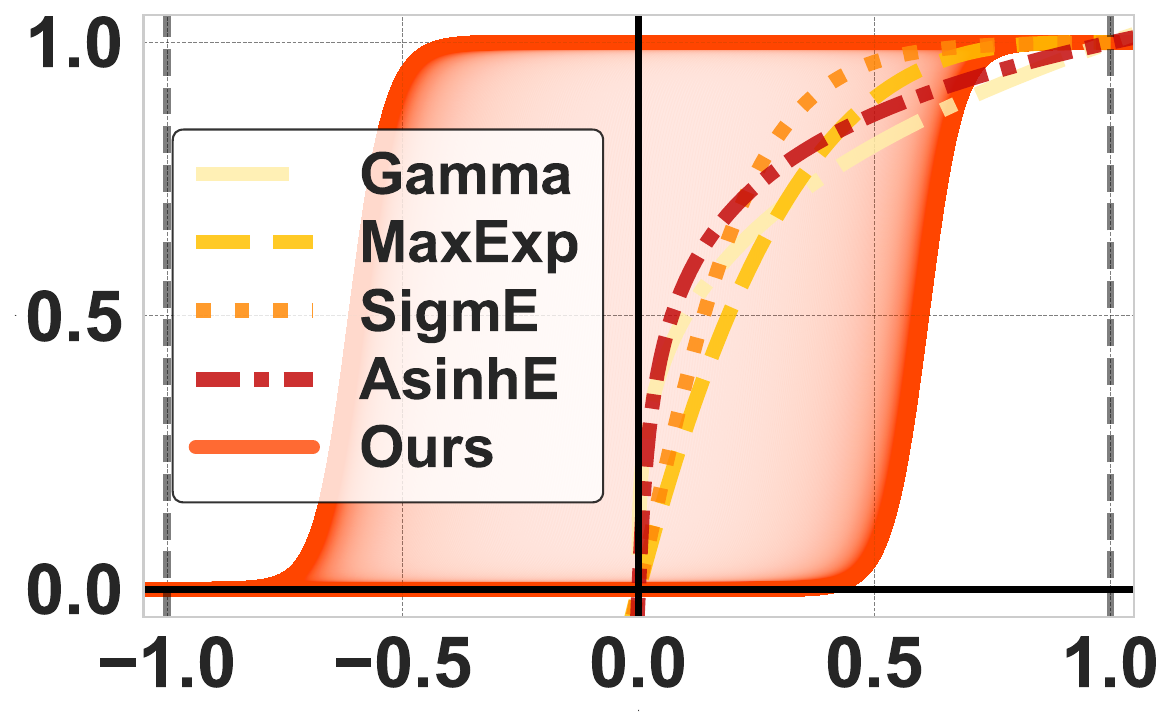}} 
    \subfigure[$a(m)$ as slope.]{\includegraphics[trim=0.1cm 0cm 0.1cm 0.0cm, clip=true, height=2.25cm]{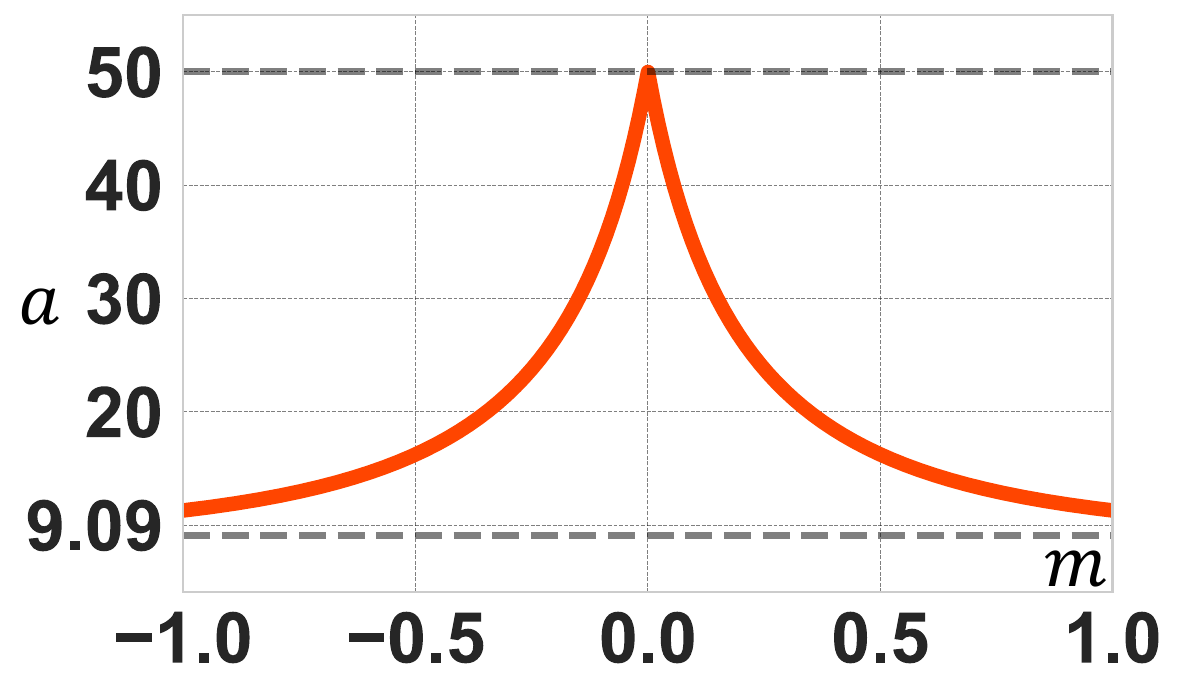}} 
    \subfigure[$b(n)$ as shift.]{\includegraphics[trim=0.1cm 0cm 0.1cm 0.0cm, clip=true, height=2.25cm]{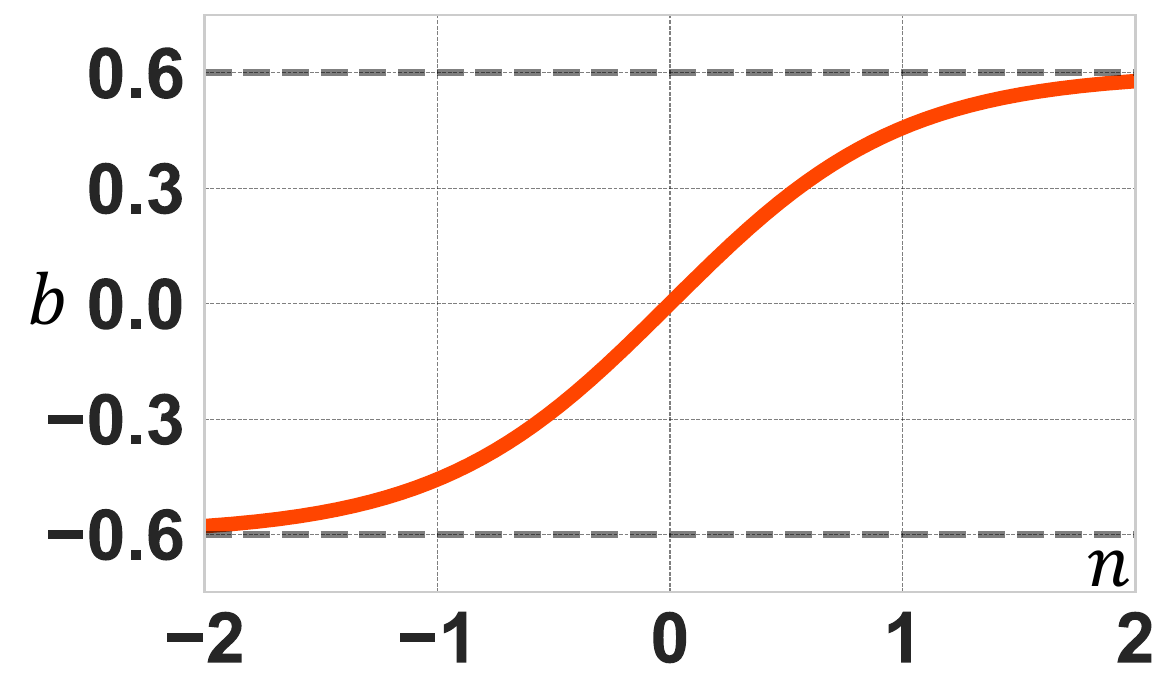}}
    \subfigure[Before \& after PN]{\includegraphics[trim=0.1cm 0cm 0.1cm 0.0cm, clip=true, height=2.25cm]{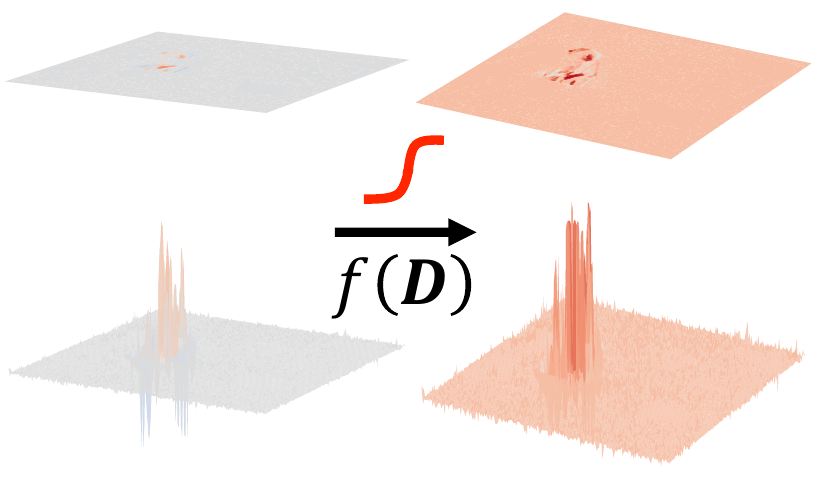}}
    \caption{(a) Comparison of existing well-behaved Power Normalization (PN) functions~\citep{koniusz2021power} and our learnable PN function (Eq.~\eqref{eq:final-sig}). Our PN function is depicted in orange with shadows, showing its learnable nature and potential shifts. % It covers a wide range and is particularly suitable for normalizing the pixel values of frame differencing maps within the range $[-1, 1]$ (bounded by vertical dashed lines). 
    The learnable slope function $a(m)$ in (b) controls the steepness of PN, determining the degree of enhancement or dampening of motions, while the learnable shift function $b(n)$ in (c) determines the threshold for PN, affecting whether the motions are enhanced or dampened. (d) shows a surface plot of pixel value changes before and after applying our PN function.}
    \label{fig:pn-comp}
    \vspace{-0.6cm}
\end{figure*}

\noindent\textbf{Frame differencing maps.} 
For a $T$-frame video $\tX=[\tF_1, \tF_2, \dots, \tF_T]\in\mbr{H\times W\times 3\times T}$ where $\tF_t \in \mbr{H\times W\times 3}$ ($t \in \idx{T}$), $H$ and $W$ denote the frame height and width, respectively, we first convert it into a grayscale video sequence $\tX'=[\mF'_1, \mF'_2, \dots, \mF'_T]\in\mbr{H\times W\times T}$. After normalizing the pixel values between 0 and 1, we compute the frame differencing maps between consecutive frames, resulting in $\tD=[\mD_1, \mD_2, \dots, \mD_{T-1}]\in\mbr{H\times W\times {(T-1)}}$, where $\mD_t=\mF'_{t+1}-\mF'_t$ ($t\in\idx{T-1}$). Note that the pixel values in $\mD_t$ can be either positive or negative. Positive values indicate areas where the pixel intensity has increased from frame $t$ to frame $t+1$, while negative values indicate areas where the pixel intensity has decreased. Note that the pixel values in frame differencing maps are in the range of $[-1, 1]$. 

The frame differencing maps $\tD$, record a sequence of motions between consecutive video frames, capturing both foreground and background motions, as well as noisy patterns. Depending on the task, we aim to enhance the motions of interest while suppressing the rest. For example, in human action recognition, we want to amplify the motions associated with human actions and reduce irrelevant motions. 
Below, we introduce our learnable Sigmoid activation function as a Power Normalization (PN) function for motion modulation. Fig.~\ref{fig:pn-comp} (a) compares existing PN functions with our learnable PN function. Appendix~\ref{sec:appl-vis} presents visual comparisons of the motions modulated by these PN functions, including ours. We observe that our PN captures different motions across various video types.

\noindent\textbf{Learnable slope and shift parameters.}
We opt for a modified Sigmoid function with learnable slope $a$ and shift $b$ as a PN function (see Fig.~\ref{fig:pn-comp} (a)) on frame differencing maps. For a given frame differencing map $\mD$, we define:
\begin{equation}
    f(\mD) = \frac{1}{1 + e^{-a \left(\mD - b\right)}}
    \label{eq:learned-sigmoid}
\end{equation}
This element-wise function maps the pixel values of frame differencing maps from $[-1, 1]$ to $[0, 1]$. For simplicity, we set $a > 0$ so that $f(\cdot)$ is a monotone increasing function. 
The parameter $a$ controls the slope of the Sigmoid function\footnote{The value of $a$ determines the sensitivity of the function $f(\mD)$ to changes in $\mD$ around the threshold $b$: large $a$ results in a steep slope with a sharp transition and high sensitivity to changes, whereas small $a$ results in a gentle slope with a smooth transition and lower sensitivity to changes.
%The value of $a$ determines the sensitivity of the function $f(\mD)$ to changes in $\mD$ around the threshold $b$. Large $a$: steep slope, sensitive to changes, sharp transition; small $a$: gentle slope, less sensitive to changes, smooth transition.
}, and it influences how sharply the function transitions from its minimum to its maximum value (see Fig.~\ref{fig:pn-comp}(b)).
The parameter $b$ acts as the threshold or the point of inflection of the Sigmoid function\footnote{The value of $b$ determines the horizontal position of the Sigmoid function: larger $b$ shifts the curve to the right, requiring higher $\mD$ values to reach the midpoint; smaller $b$ shifts the curve to the left, requiring lower $\mD$ values to reach the midpoint.
% The value of $b$ determines the horizontal position of the Sigmoid function. Larger $b$: shifts the curve to the right, higher $\mD$ values are needed to reach the midpoint; smaller $b$: shifts the curve to the left, lower $\mD$ values are needed to reach the midpoint.
}, and it determines the position of the Sigmoid curve along the $\mD$ axis.
We allow $b$ to shift either left or right, hence it can be positive or negative (see Fig. \ref{fig:pn-comp}(c)).
However, directly learning the parameters $a$ and $b$ presents challenges, such as initialization sensitivity and uncertainty in the parameter search space.
To address these issues, we design the following two mapping functions to learn $m$ and $n$ instead:
\begin{equation}
\left\{
\begin{aligned}
    & a(m)= \frac{\alpha}{\beta\left|\tanh(m)\right| + \epsilon}\\
    & b(n)=\gamma \tanh(n)
    \label{eq:ab-sigmoid}
\end{aligned}
\right.
\end{equation}
Here, $\tanh(\cdot)$ is the hyperbolic tangent function, which maps any given $m$ and $n$ to values between $-1$ and $1$. This alignment with the pixel value range in frame differencing maps facilitates better adjustment of the slope and shift for motion modulation. The symbol $|\cdot|$ denotes the absolute value operation, and $\epsilon$ is a small constant for numerical stabilization (we set $\epsilon=0.1$). Parameters $\alpha>0$, $\beta>0$, and $\gamma>0$ control the characteristics of our PN function: $\alpha$ and $\beta$ adjusts the rate at which Eq.~\eqref{eq:learned-sigmoid} transitions between 0 and 1, while $\gamma$ in $b(n)$ controls the shifts of input before it is processed by the exponential function.
This setup allows for unrestricted learning of $m$ and $n$, avoiding potential issues caused by the bounded pixel values in $\mD$. We prove the boundedness and differentiability of both $a(m)$ and $b(n)$ in Appendix~\ref{sec:appl-proof}.
Combining Eq.~\eqref{eq:learned-sigmoid} and~\eqref{eq:ab-sigmoid} results in our learnable PN function:
\begin{equation}
    f(\mD) = \frac{1}{1 + e^{-\left(\frac{\alpha}{\beta\left|\tanh(m)\right| + \epsilon}\right) \left(\mD - \gamma\tanh(n)\right)}}.
    \label{eq:final-sig}
\end{equation}
 % Both $a$ and $b$ are bounded differentiable real functions. 
We provide a detailed analysis of parameter constraints and sensitivity concerning $\alpha$, $\beta$, and $\gamma$, their impact on the learning of $m$ and $n$, and the rationale behind selecting these three parameters in Appendix~\ref{sec:suppl-proof}. Specifically, we set $\alpha=5$, $\beta=0.45$, and $\gamma=0.6$.
Below we show that for pixel values within interval $[-1, 1]$ in $\mD$, our function $f(\mD)$ is always bounded within $[0, 1]$. Additionally, we show that Eq.~\eqref{eq:final-sig} is a well-behaved PN function.

\begin{tcolorbox}[ width=1.0\linewidth, colframe=blackish, colback=beaublue, boxsep=0mm, arc=3mm, left=1mm, right=1mm, top=1mm, bottom=1mm]
\noindent\textbf{Upper and lower bound.} Consider the term inside the exponential: $\mA = a(m)(\mD - b(n))$, we first examine the derivative of $\mA$ with respect to $\mD$:
$\frac{\partial\mA}{\partial\mD} = a(m) = \frac{\alpha}{\beta|\tanh(m)| + \epsilon}$. Since this derivative is always positive, $\mA$ is monotonically increasing with respect to $\mD$. 
The scaling factor $a(m)$ ranges from $\frac{\alpha}{\beta+\epsilon}$ to $\frac{\alpha}{\epsilon}$ and $\mD - b(n)$ ranges from $-1-\gamma$ to $1+\gamma$; hence, the lower and upper bounds of $\mA$ are $-\frac{\alpha(1+\gamma)}{\beta+\epsilon}$ and $\frac{\alpha(1+\gamma)}{\epsilon}$, respectively.

The Sigmoid function $\sigma(\mA) = \frac{1}{1+e^{-\mA}}$ approaches 0 as value in $\mA$ becomes large and negative (\eg, $\sigma(-\frac{5(1+0.6)}{0.45+0.1})\approx0.0$), and 1 as value in $\mA$ becomes large and positive (\eg, $\sigma(\frac{5(1+0.6)}{0.1})\approx1.0$). Consequently, for values in $\mD$ in the interval $[-1, 1]$, the function $f(\mD)$ is always bounded within $[0, 1]$.

\noindent\textbf{Well-behaved power normalization.}
Eq.~\eqref{eq:final-sig} is continuous and smooth for all real values of $\mD$, $m$, and $n$. The scaling factor $a(m)$ is always positive, and the exponential term is well-defined, producing a valid, finite value for all input values. The Sigmoid function $\sigma(\mA) = \frac{1}{1+e^{-\mA}}$ maps any real number to the interval $[0,1]$, ensuring that $f(\mD)$ produces values within this range and thereby maintaining proper normalization. 

Therefore, Eq.~\eqref{eq:final-sig} is a well-behaved PN function given: (i) It is continuous and smooth. (ii) The output values are properly normalized within the range $[0, 1]$. (iii) The Sigmoid function ensures that the function maps real numbers to a bounded interval, maintaining normalization. 
\end{tcolorbox}

\subsection{Motion Prompt Layer: An Adapter}
\label{sec:motion-prompt}

\noindent\textbf{Video motion prompts.} 
Eq.~\eqref{eq:final-sig} modulates the motions in frame differencing maps via the learnable $m$ and $n$, resulting in a sequence of normalized frame differencing maps with pixel values ranging from 0 to 1. This element-wise PN process can also be viewed as activating the motions of interest, guided by the generic loss function, \eg, cross-entropy, hence we call this the attention map. The PN process produces a sequence of attention maps: $\tA' = [f(\mD_1), f(\mD_2), \dots, f(\mD_{T-1})] \in \mbr{H \times W \times (T-1)}$, spatially highlighting regions where motions are of interest (\eg, 1) and dampening motions that are not of interest (\eg, 0); temporally showing the evolution of attention maps over time. We then duplicate each attention map three times, resulting in $\tA'^{(3)} = [f^{(3)}(\mD_1), f^{(3)}(\mD_2), \dots, f^{(3)}(\mD_{T-1})] \in \mbr{H \times W \times 3 \times (T-1)}$, where $f^{(3)}(\mD_t) \in \mbr{H \times W \times 3}$ and $t\in\idx{T-1}$. We perform a channel-wise Hadamard product between each duplicated attention map $f^{(3)}(\mD_t)$ and the original video frame $\mF_{(t+1)}$ so that attention attends to each channel of the original video frame, resulting in a sequence of highlighted video frames:
\begin{align}
    & \tZ = \tA'^{(3)} \odot \tX_{:,:,2:} \nonumber \\
    & \quad\!\! = [f^{(3)}(\mD_1), f^{(3)}(\mD_2), \dots, f^{(3)}(\mD_{T-1})] \odot [\tF_2, \dots, \tF_T] \nonumber \\
    & \quad\!\! = [f^{(3)}(\mD_1) \odot \tF_2, f^{(3)}(\mD_2) \odot \tF_3, \dots, f^{(3)}(\mD_{T-1}) \odot \tF_T],
    \label{eq:motion-prompt}
\end{align}
where $\odot$ denotes the Hadamard (element-wise) product. 
$\tZ \in \mbr{H \times W \times 3 \times (T-1)}$ denotes the newly generated video, referred to as Video Motion Prompts (VMPs), which are motion-dependent, attention-driven, and provide rich motion cues.
$f^{(3)}(\mD_{t}) \odot \tF_{(t+1)} \in \mathbb{R}^{H \times W \times 3}$ shows the motion prompt for the $t$-th frame.
Below, we show that our video motion prompt generation process is, in fact, connected to the attention mechanism.
\begin{tcolorbox}[ width=1.0\linewidth, colframe=blackish, colback=beaublue, boxsep=0mm, arc=3mm, left=1mm, right=1mm, top=1mm, bottom=1mm]
\noindent\textbf{Connecting to attention mechanism.} We rewrite $f^{(3)}(\mD_{t})$ using the Sigmoid function $\sigma^{(3)}(\cdot)$, replace the frame differencing map $\mD_{t}$ with the grayscale conversion function $h(\cdot)$ as $h(\tF_{t+1}) - h(\tF_t)$, and rewrite Eq~\eqref{eq:motion-prompt} in the form of per-frame motion prompt (we use Eq.~\eqref{eq:ab-sigmoid} for the scaling factor and shift parameter and omit $m$ and $n$ for simplicity):
\begin{align}
    & \tZ_t = f^{(3)}(\mD_{t}) \odot \tF_{(t+1)} \nonumber \\
    % & \quad = \sigma^{(3)}(a(\mD_{t} - b)) \odot \tF_{(t+1)} \nonumber \\ 
    & \quad = \sigma^{(3)}(a[h(\tF_{t+1}) - h(\tF_t) - b]) \odot \tF_{(t+1)}
\end{align}
where $\mA = a[h(\tF_{t+1}) - h(\tF_t) - b]$ can be viewed as an attention matrix with the shifting parameter $b$ modulating the pixel intensity changes between each pair of grayscaled consecutive frames. 
The Sigmoid (denoted as $\sigma^{(3)}$) outputs are similar to the Softmax function outputs in the standard attention mechanism (as in Eq.~\eqref{eq:softm} of Appendix~\ref{sec:appl-pre}), which transforms raw attention scores to highlight the most important parts of the $h$-transformed frame differencing maps. 
Since we operate on each pixel in frame differencing maps to highlight all relevant motion pixels, there is no need to satisfy the criterion that attention weights sum up to 1, as required by the Softmax function. % Instead, we simply use the Sigmoid function.
$\tF_{(t+1)}$ is analogous to the value matrix $\mV$ in Eq.~\eqref{eq:attn-v} of Appendix~\ref{sec:appl-pre}. This shows that our motion prompt generation process is analogous to the standard attention mechanism.
\end{tcolorbox}

Unlike the traditional attention mechanism, our attention mechanism offers (i) lightweight computation with only two learnable parameters, (ii) interpretability, as the learnable scaling factor and shift parameter have well-explainable functionalities in the motion modulation process, and (iii) generalizability, as $\mA$ is motion-dependent, relying on frame differencing maps rather than being dataset-dependent.

If attention scores in $\sigma^{(3)}(\mA)$ are all 1, \eg, $\mA$ becomes large and positive due to motion modulation via slope $a$ and shift $b$, thus reverting to the use of original video frames.

\noindent\textbf{Temporal attention variation regularization.} To ensure temporally the smoothness and continuity of generated attention maps, we introduce temporal attention variation regularization on pair-wise attention maps:
\begin{equation}
\mathcal{V} = \frac{1}{T-2}\sum_{t=1}^{T-2}||f(\mD_{t+1})-f(\mD_t)||_F^2,
\label{eq:reg}
\end{equation}
where $||\cdot||_F$ denotes the Frobenius norm. Eq.~\eqref{eq:reg} reduces pixel variations between consecutive attention maps, ensuring temporal smoothness while preserving key motion regions.

\begin{figure*}[tbp]
    \centering
    \subfigure[$a$ \vs epoch]{\includegraphics[height=2.553cm]{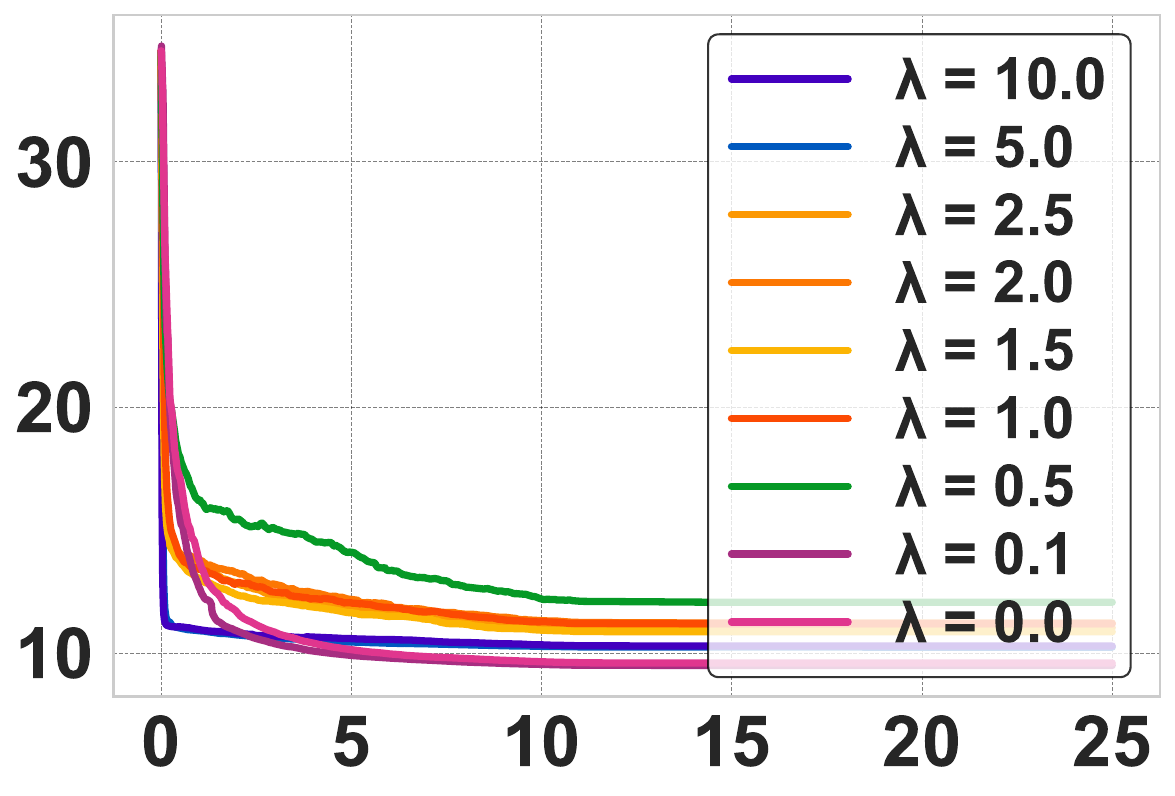}}
    \subfigure[$b$ \vs epoch]{\includegraphics[height=2.553cm]{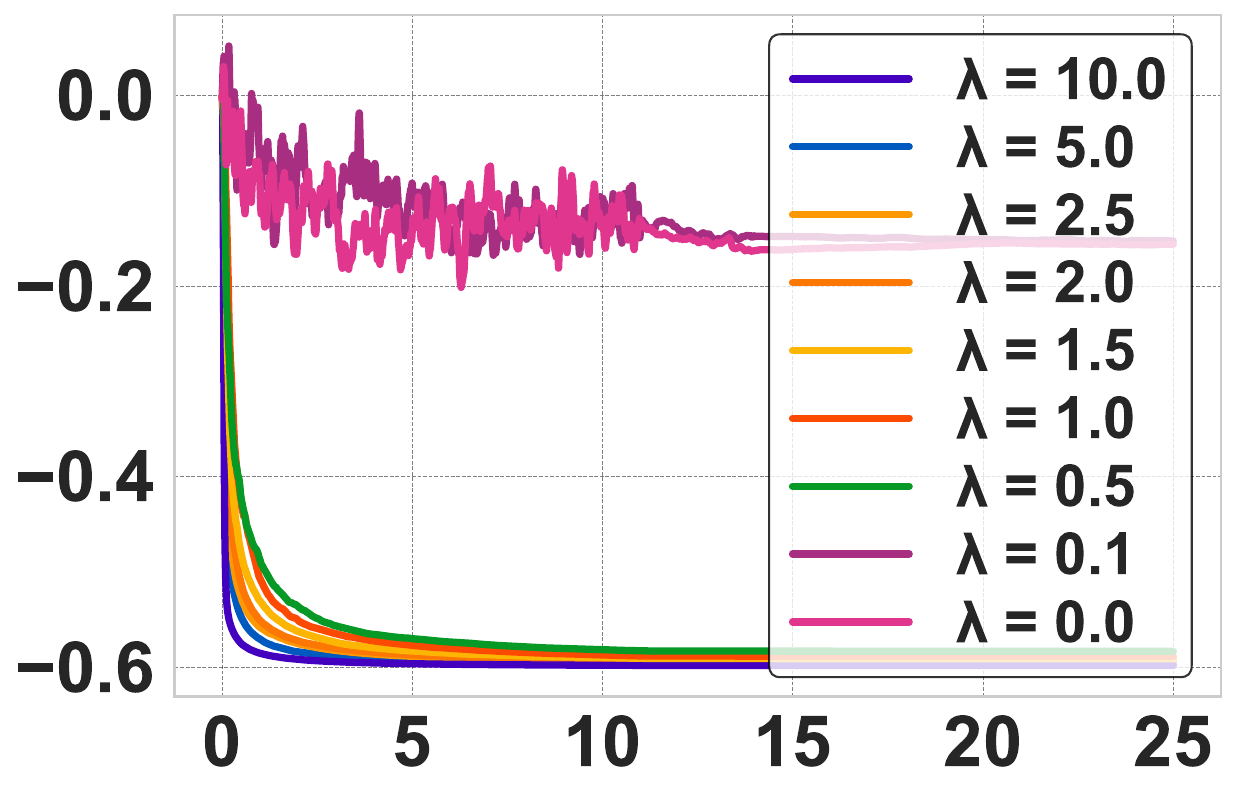}} 
    \subfigure[Across datasets]{\includegraphics[height=2.553cm]{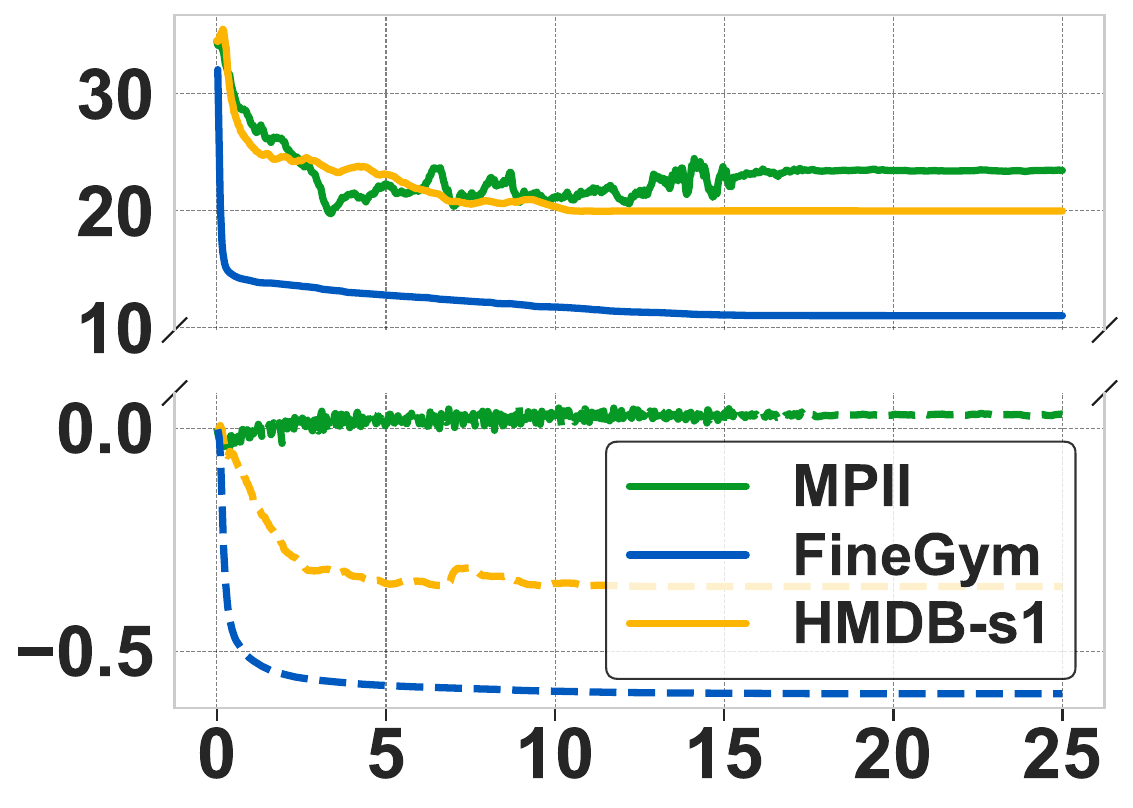}}
    \subfigure[Across backbones]{\includegraphics[height=2.553cm]{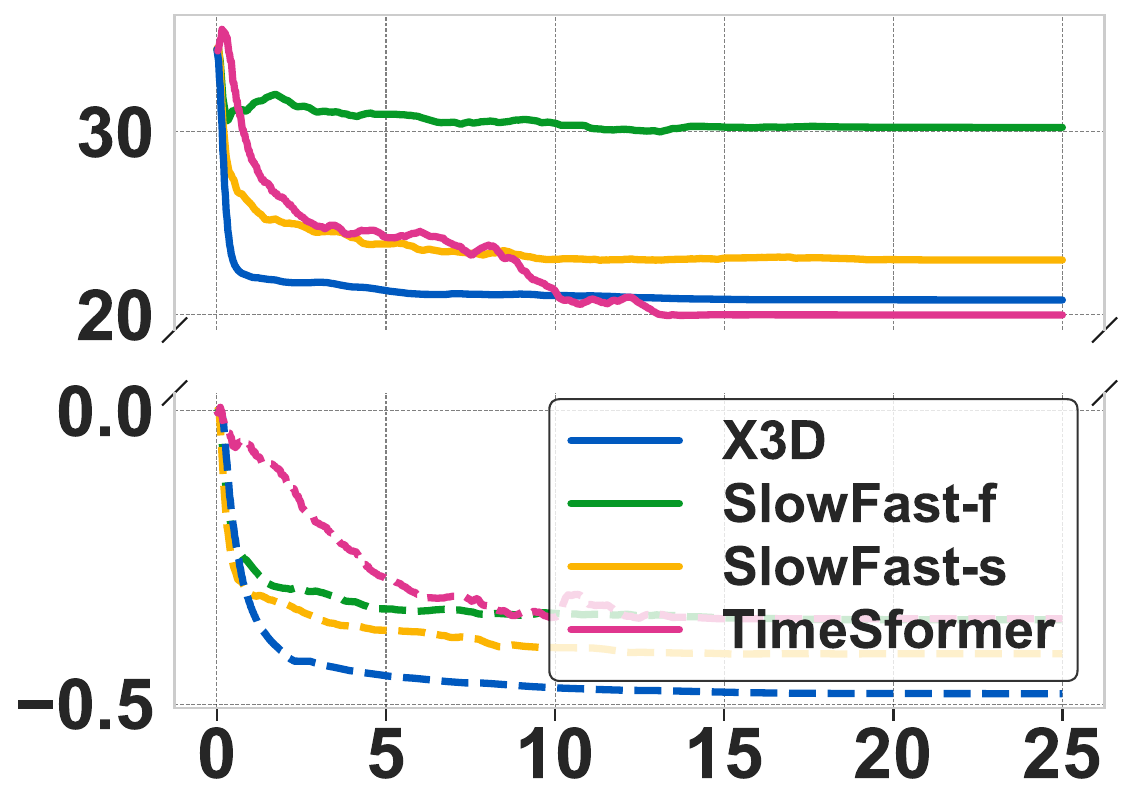}} 
    \caption{Effects of the penalty parameter $\lambda$ on (a) slope $a$ and (b) shift $b$ are evaluated using FineGym with TimeSformer pretrained on Kinetics-600. Larger $\lambda$ values cause both $a$ and $b$ to approach their lower bounds. In contrast, when smaller $\lambda$s are used, $b$ becomes noisy and fluctuates, while $a$ assumes lower values.
    % Overall, on FineGym, $a$ and $b$ tend to be lower: our PN function shifts to the left and the steepness tends to decrease, which considers some negative motions while moderately adjusting transitions. 
    % Additionally, we observe that the learning of $a$ and $b$ typically lasts around 15 epochs during fine-tuning.
    (c) and (d) show the learned $a$ (solid lines) and $b$ (dashed lines) of top performers across different datasets (with TimeSformer) and backbones (on HMDB-51 split 1). In (d), SlowFast-f and SlowFast-s denote motion prompt layer added to the fast-only (green color) and slow-only (yellow color) stream, respectively.
    }
    \label{fig:learned-pn-a-b}
    \vspace{-0.6cm}
\end{figure*}

We design the video motion prompt generation process as a single layer with two learnable parameters that amplify relevant motions while attenuating irrelevant movements. Eq.~\eqref{eq:reg} is incorporated into the original loss function $\mathcal{L}_\text{ori}$, such as cross-entropy loss for action recognition, used in models like SlowFast and TimeSformer backbones:
\begin{equation}
\mathcal{L} = \mathcal{L}_\text{ori} + \lambda \mathcal{V},
\label{eq:final-loss}
\end{equation}
where $\lambda$ is a penalty parameter that controls the strength of this regularization, balancing the trade-off between temporal smoothness and the maintenance of spatially significant motion regions. 
We simply insert our motion prompt layer between the video input and the model architecture, using Eq.~\eqref{eq:final-loss} as the loss function to learn the VMPs as new inputs. The entire model can be learned in an end-to-end manner or fine-tuned on specific layers, including the learning of our motion prompt layer.

\section{Experiment}
\label{sec:exp}

% shake, Uneven Bar, Balance Beam, Pushup
% The consecutive frames exhibit significant discrepancies due to modern video processing sampling strategies.

\subsection{Setup}
\label{sec:data-setup}

\begin{figure}[tbp]
\begin{center}
\includegraphics[width=\textwidth]{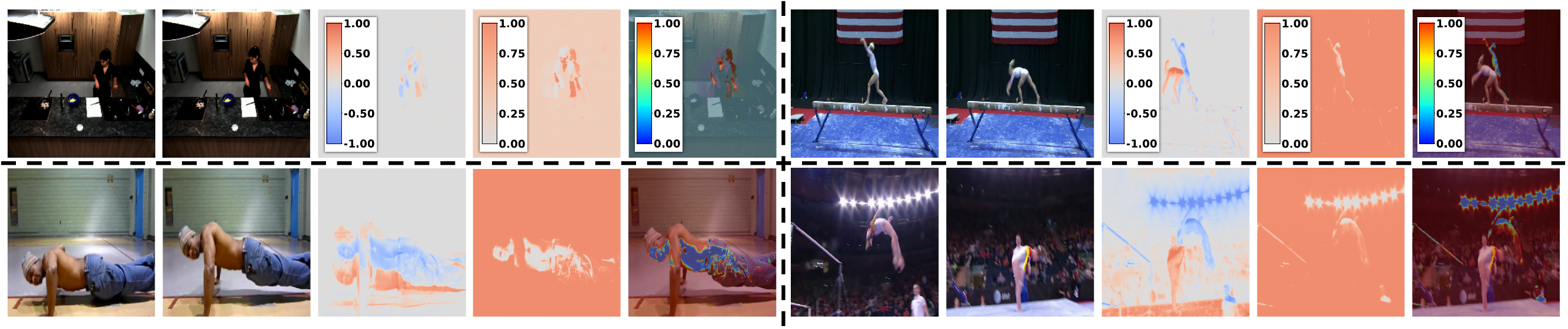}
\caption{
Visualizations of consecutive frames (columns 1-2 and 6-7), frame differencing maps (columns 3 and 8), attention maps (columns 4 and 9), and motion prompts (columns 5 and 10). Top left shows \textit{shake} (MPII), bottom left shows \textit{pushup} (HMDB-51), and top and bottom right show \textit{balance beam} and \textit{uneven bar} (FineGym). Frame differencing maps are noisy; our attention maps, guided by learned slope and shift, are cleaner. Motion prompts (columns 5 and 10) contain richer motion information than the original frames (columns 2 and 7). We notice that for static camera, the attention map shows light orange, indicating that background information is not important for the action (\eg, \textit{shake} in MPII). For moving cameras, the background information is important, hence it appears darker red, receiving higher attention scores. Additional visualizations are in Appendix~\ref{sec:appl-vis}.
}
\label{fig:vis-motion-prompt}
\end{center}
\vspace{-0.6cm}
\end{figure}

\noindent\textbf{Dataset.} For generic action recognition, we choose the popular and challenging HMDB-51~\citep{kuehne2011hmdb}, which features significant camera and background motion. For fine-grained action recognition, we select the large-scale MPII Cooking 2 (2,881,616 frames, resolution $1624\!\times\!1224$)~\citep{rohrbach15ijcv} and FineGym~\citep{shao2020finegym} (Gym99 v1.1: 26,320/8,521 for train/val set, respectively) datasets. % For anomaly detection, we use the widely recognized UCF-Crime~\cite{sultani2018real}, XD-Violence~\cite{Wu2020not}, and UBnormal~\cite{ubnormal2022} datasets.
% FineGym and MPII Cooking 2 are popular datasets for fine-grained action recognition. 
FineGym focuses on human actions performed in a gym environment, capturing a wide range of activities with fine granularity (99 classes). In contrast, the MPII Cooking 2 dataset specializes in cooking-related actions (67 classes).
We adhere to their standard evaluation protocols in our experiments.

\noindent\textbf{Implementation.} The motion prompt layer is initialized with a normal distribution, having a mean of $1\!\! \times\!\! 10^{-5}$ and a standard deviation of 1. The penalty parameter for temporal attention variation regularization is selected from the range [1e-4, 10]. We use SlowFast~\citep{feichtenhofer2019slowfast}, C2D, I3D~\citep{i3d_net}, X3D~\citep{feichtenhofer2020x3d}, and TimeSformer~\citep{gberta_2021_ICML} as backbones.
All experiments use SGD as the optimizer (\eg, with momentum 0.9). The learning rate (\eg, 0.005), weight decay (\eg, 0.0001), decay strategy (\eg, step decay, or cosine decay with a warm-up), and the number of sampled video frames per video follow those specified in the original papers. Note that our layer requires an additional frame to ensure that the resulting motion prompts have exactly the same length as the original input video frames (see Eq.~\eqref{eq:motion-prompt}).
We fine-tune models pretrained on Kinetics-400~\citep{kay2017kinetics} (or Kinetics-600~\citep{long2020learning}) as our baseline. All experiments are conducted using the Tesla V100 GPU. Below, we present our evaluations and analysis. 

\begin{figure}[tbp]
\begin{center}
\includegraphics[width=\textwidth]{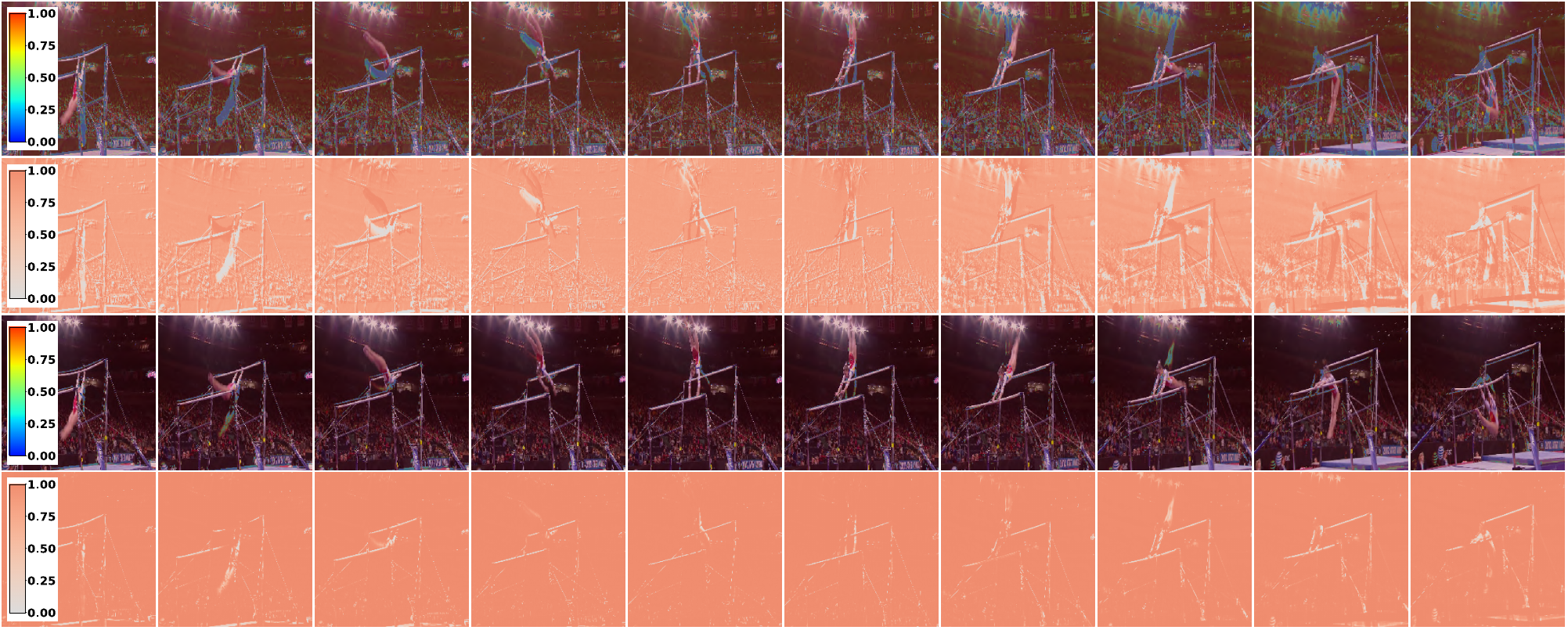}
\caption{Effects of the regularization term. We use the \textit{uneven bar} action from FineGym for visualization. The first two rows show motion prompts and attention maps without regularization ($\lambda\!=\!0$). % , learned $a\!=\!9.61$ and $b\!=\!-0.15$). 
The last two rows show results with regularization ($\lambda\!=\!2.5$). %, learned $a\!=\!11.04$ and $b\!=\!-0.59$). 
Regularization removes unnecessary motion details, resulting in smoother and cleaner attention maps. More visualizations are in Appendix~\ref{sec:appl-vis}.}\label{fig:motion-prompt-attn}
\end{center}
\vspace{-0.6cm}
\end{figure}

\subsection{Evaluation}
\label{sec:main-eval}

\noindent\textbf{Analysis of learnable slope and shift.} We present the learning process of $a$ and $b$ versus the number of fine-tuning epochs in Fig.~\ref{fig:learned-pn-a-b} with varying regularization penalty parameter $\lambda$. We use FineGym with TimeSformer pretrained on Kinetics-600 for fine-tuning with our motion prompts. As shown in the figure, choosing bigger $\lambda$ results in both slope and shift parameters quickly approaching their lower bounds. Using smaller $\lambda$ or set $\lambda$ to 0 results in the noisy and fluctuated learning process for $b$ (red and purple lines in Fig.~\ref{fig:learned-pn-a-b} (b)), and $a$ tends to be slightly bigger (green line in Fig.~\ref{fig:learned-pn-a-b} (a)).
Overall, on FineGym, $a$ and $b$ tend to be small, %in other words, our PN function tends to shift to the left, and the steepness tends to be smaller, 
that is to consider some negative motions in the frame differencing maps while ensuring a smooth transition rather than an increase in steepness. This is reasonable as FineGym is captured by moving cameras, hence all positive motions should be considered with varying degrees of attention. 
% We also notice that the learning for both $a$ and $b$ lasts around 15 epochs for fine-tuning, the potential reasons are (i) learning rate decays to a smaller value (ii) the stabilization of the learning process with better motion prompts obtained. 
The optimal value of $\lambda$ is 2.5, and the learned values are $a = 11.04$ and $b = -0.59$, resulting in a performance gain of 0.8\% compared to the baseline. % It is noteworthy that the improvement is obtained from just two additional learned parameters.

\noindent\textbf{Learned $a$ and $b$ across various datasets and backbones.} We visualize the pairs of $a$ and $b$ from top performers per dataset using the TimeSformer backbone in Fig.~\ref{fig:learned-pn-a-b} (c). We notice that MPII Cooking 2 tends to have higher $a$ and $b$ values. This is attributed to the dataset being captured by a static camera, making it easier to extract motions related to cooking activities, \eg, with a steep slope. On FineGym, both slope and shift tend to be smaller compared to HMDB-51 split 1 (HMDB-s1). This is because FineGym focuses specifically on gymnastic activities, where significant camera motions occur due to player localization and tracking. 
In Fig.~\ref{fig:learned-pn-a-b} (d), we observe that the learned $a$ and $b$ vary significantly across different backbones. Moreover, using motion prompts on the SlowFast fast-only (SlowFast-f) stream results in a steeper slope compared to the SlowFast slow-only (SlowFast-s) stream. This is because the fast-only stream samples more frames, offering richer and smoother temporal information that facilitates easier access to motions of interest.

\begin{table}[tbp]
\centering
\setlength{\tabcolsep}{0.02pt}
\caption{Evaluations are conducted on (\textit{left}) HMDB-51, and (\textit{right}) FineGym, MPII Cooking 2, using SlowFast, C2D, I3D, X3D and TimeSformer as backbones. For SlowFast, we explore three variants by adding motion prompts into the slow-only stream, fast-only stream, and both slow and fast streams. \textit{K600} denotes that the Kinetics-600 pretrained model is used. We highlight improvements in red.}
\begin{tabular}{cc}
\resizebox{0.5\linewidth}{!}{
\begin{tabular}{lcccccccc}
\toprule
    \multirow{2}{*}{Model} & \multicolumn{6}{c}{HMDB-51} & \multicolumn{2}{c}{\multirow{2}{*}{Mean}} \\
    \cline{2-7}
    & \multicolumn{2}{c}{Split 1} & \multicolumn{2}{c}{Split 2} & \multicolumn{2}{c}{Split 3} & & \\ 
\midrule
    SlowFast & 75.4 &  & 76.2 &  & 76.9 &  & 76.2 & \\
    +\textbf{VMPs} (slow-only) & \textbf{76.8} & $^\textbf{\textcolor{red}{$\uparrow$1.4}}$ & \textbf{77.0} & $^\textbf{\textcolor{red}{$\uparrow$0.8}}$ & \textbf{77.3} & $^\textbf{\textcolor{red}{$\uparrow$0.4}}$ & \textbf{77.0} & $^\textbf{\textcolor{red}{$\uparrow$0.8}}$ \\
    +\textbf{VMPs} (fast-only) & \textbf{76.5} & $^\textbf{\textcolor{red}{$\uparrow$1.1}}$ & \textbf{77.4} & $^\textbf{\textcolor{red}{$\uparrow$1.2}}$ & \textbf{77.1} & $^\textbf{\textcolor{red}{$\uparrow$0.2}}$ & \textbf{77.0} & $^\textbf{\textcolor{red}{$\uparrow$0.8}}$ \\
    +\textbf{VMPs} (slow\&fast) & \textbf{76.2} & $^\textbf{\textcolor{red}{$\uparrow$0.8}}$ & \textbf{76.7} & $^\textbf{\textcolor{red}{$\uparrow$0.5}}$ & \textbf{77.1} & $^\textbf{\textcolor{red}{$\uparrow$0.2}}$ & \textbf{76.6} & $^\textbf{\textcolor{red}{$\uparrow$0.4}}$ \\
    \hline
    C2D & 67.7 & & 66.9 & & 66.1 & & 66.9 \\
    +\textbf{VMPs} & \textbf{69.4} & $^\textbf{\textcolor{red}{$\uparrow$1.7}}$ & \textbf{68.3} & $^\textbf{\textcolor{red}{$\uparrow$1.4}}$ & \textbf{66.9} & $^\textbf{\textcolor{red}{$\uparrow$0.8}}$ & \textbf{68.2} & $^\textbf{\textcolor{red}{$\uparrow$1.3}}$ \\
    \hline
    I3D & 70.1 & & 69.7 & & 69.2 & & 69.7 \\
    +\textbf{VMPs} & \textbf{70.5} & $^\textbf{\textcolor{red}{$\uparrow$0.4}}$ & \textbf{70.5} & $^\textbf{\textcolor{red}{$\uparrow$0.8}}$ & \textbf{70.2} & $^\textbf{\textcolor{red}{$\uparrow$1.0}}$ & \textbf{70.4} & $^\textbf{\textcolor{red}{$\uparrow$0.7}}$ \\
    \hline
    X3D & 75.0 & & 72.6 & & 73.4 & & 73.7 \\
    +\textbf{VMPs} & \textbf{75.8} & $^\textbf{\textcolor{red}{$\uparrow$0.8}}$ & \textbf{73.2} & $^\textbf{\textcolor{red}{$\uparrow$0.6}}$ & \textbf{73.6} & $^\textbf{\textcolor{red}{$\uparrow$0.2}}$ & \textbf{74.2} & $^\textbf{\textcolor{red}{$\uparrow$0.5}}$ \\
    \hline
    TimeSformer & 70.0 & & 72.1 & & 70.8 & & 71.0 \\
    +\textbf{VMPs} & \textbf{72.7} & $^\textbf{\textcolor{red}{$\uparrow$2.7}}$ & \textbf{73.8} & $^\textbf{\textcolor{red}{$\uparrow$1.7}}$ & \textbf{70.9} & $^\textbf{\textcolor{red}{$\uparrow$0.1}}$ & \textbf{72.5} & $^\textbf{\textcolor{red}{$\uparrow$1.5}}$ \\
    \hline
    TimeSformer (\textit{K600}) & 72.7 & & 73.1 & & 72.2 & & 72.7 \\
    +\textbf{VMPs} & \textbf{74.2} & $^\textbf{\textcolor{red}{$\uparrow$1.5}}$ & \textbf{74.3} & $^\textbf{\textcolor{red}{$\uparrow$1.2}}$ & \textbf{72.9} & $^\textbf{\textcolor{red}{$\uparrow$0.7}}$ & \textbf{73.8} & $^\textbf{\textcolor{red}{$\uparrow$1.1}}$ \\
\bottomrule
\end{tabular}
} & 
\resizebox{0.493\linewidth}{!}{%
\begin{tabular}{lccccccccc}
\toprule
    \multirow{2}{*}{Model} &  \multicolumn{4}{c}{FineGym} & & \multicolumn{4}{c}{MPII Cooking 2} \\ 
    \cline{2-5} \cline{7-10}
    & \multicolumn{2}{c}{Top-1} & \multicolumn{2}{c}{Top-5} & &  \multicolumn{2}{c}{Top-1} & \multicolumn{2}{c}{Top-5} \\
\midrule
    SlowFast & \textbf{89.8} & & 99.2 & & & 52.9 & & \textbf{86.1} & \\
    +\textbf{VMPs} (slow-only) & 89.7 & $^{\textcolor{blue}{\downarrow0.1}}$ & 99.2 & & & \textbf{55.5} & $^\textbf{\textcolor{red}{$\uparrow$2.6}}$ & 84.5 & $^{\textcolor{blue}{\downarrow1.6}}$ \\
    +\textbf{VMPs} (fast-only) & \textbf{90.3} & $^\textbf{\textcolor{red}{$\uparrow$0.5}}$ & \textbf{99.3} & $^\textbf{\textcolor{red}{$\uparrow$0.1}}$ & & \textbf{55.2} & $^\textbf{\textcolor{red}{$\uparrow$2.3}}$ & 84.0 & $^{\textcolor{blue}{\downarrow2.1}}$ \\
    +\textbf{VMPs} (slow\&fast) & \textbf{90.1} & $^\textbf{\textcolor{red}{$\uparrow$0.3}}$ & \textbf{99.3} & $^\textbf{\textcolor{red}{$\uparrow$0.1}}$ & & \textbf{56.8} & $^\textbf{\textcolor{red}{$\uparrow$3.9}}$ & \textbf{86.6} & $^\textbf{\textcolor{red}{$\uparrow$0.5}}$ \\
    \hline
    C2D & 79.7 & & 97.5 & & & 48.8 & & \textbf{83.4} & \\
    +\textbf{VMPs} & \textbf{81.3} & $^\textbf{\textcolor{red}{$\uparrow$0.6}}$ & \textbf{97.8} & $^\textbf{\textcolor{red}{$\uparrow$0.3}}$ & & \textbf{50.9} & $^\textbf{\textcolor{red}{$\uparrow$2.1}}$ & 82.1 & $^{\textcolor{blue}{\downarrow1.3}}$ \\
    \hline
    I3D & 82.4 & & 98.3 & & & 53.1 & & 82.9 & \\
    +\textbf{VMPs} & \textbf{84.7} & $^\textbf{\textcolor{red}{$\uparrow$2.3}}$ & \textbf{98.4} & $^\textbf{\textcolor{red}{$\uparrow$0.1}}$ & & \textbf{56.1} & $^\textbf{\textcolor{red}{$\uparrow$3.0}}$ & \textbf{85.7} & $^\textbf{\textcolor{red}{$\uparrow$2.8}}$ \\
    \hline
    X3D & 83.0 & & 98.4 & & & 48.4 & & \textbf{80.8} & \\
    +\textbf{VMPs} & \textbf{83.8} & $^\textbf{\textcolor{red}{$\uparrow$0.8}}$ & \textbf{98.6} & $^\textbf{\textcolor{red}{$\uparrow$0.2}}$ & & \textbf{49.1} & $^\textbf{\textcolor{red}{$\uparrow$0.7}}$ & 80.6 & $^{\textcolor{blue}{\downarrow0.2}}$ \\
    \hline
    TimeSformer & 83.5 & & 98.5 & & & 50.0 & & 79.4 & \\ 
    +\textbf{VMPs} & \textbf{83.8} & $^\textbf{\textcolor{red}{$\uparrow$0.3}}$ & \textbf{98.6} & $^\textbf{\textcolor{red}{$\uparrow$0.1}}$ & & \textbf{55.2} & $^\textbf{\textcolor{red}{$\uparrow$5.2}}$ & \textbf{82.5} & $^\textbf{\textcolor{red}{$\uparrow$3.1}}$ \\
    \hline
    TimeSformer (\textit{K600}) & 83.6 & & \textbf{98.7} & & & 50.6 & & 81.8 & \\ 
    +\textbf{VMPs} & \textbf{84.4} & $^\textbf{\textcolor{red}{$\uparrow$0.8}}$ & 98.5 & $^{\textcolor{blue}{\downarrow0.2}}$ & & \textbf{56.6} & $^\textbf{\textcolor{red}{$\uparrow$6.0}}$ & \textbf{84.4} & $^\textbf{\textcolor{red}{$\uparrow$2.6}}$ \\
\bottomrule
\end{tabular}
}
\end{tabular}
\label{tab:combined}
\end{table}

\noindent\textbf{Visualizations of attention maps and motion prompts.} We visualize frame difference maps, learned attention maps, and motion prompts in Fig.~\ref{fig:vis-motion-prompt}. We also include the original video frames for comparison. As shown in the figure, we observe discrepancies between consecutive frames, attributable to the frame sampling strategy commonly used in video processing tasks.
The frame differencing maps show noticeable noise; blue indicates negative motions while orange shows positive motions, especially in videos captured by moving cameras like HMDB-51 and FineGym. Conversely, in static camera scenes such as MPII Cooking 2, the background appears clean. Consequently, in the generated attention maps, the background is depicted in lighter orange, suggesting lower attention scores and less importance relative to the action.
In contrast, in scenarios with moving cameras, background motions appear more significant, reflected by darker red shades in the attention maps, indicating higher attention scores. Interestingly, the generated motion prompts reveal rich action information, compared to original frames.

\noindent\textbf{With and without temporal attention variation regularization.} %We explore the use of temporal attention variation regularization in the generation of attention maps and motion prompts. 
Fig.~\ref{fig:motion-prompt-attn} shows a comparison of with and without the use of temporal attention variation regularization. We observe that without the regularization term, the generated attention maps are quite noisy, especially in the background. However, with regularization, the attention maps contain much less noise. This demonstrates that our regularization term contributes to generating smooth and clean attention maps, thereby improving the quality of motion prompts. 
We also observe that the attention maps exhibit several interesting patterns: (i) they highlight the motion regions in the current frame, (ii) they capture potential movements from the previous frame, and (iii) they attend to background scenes affected by camera motions. These observations indicate that our attention maps, guided by only two learnable parameters, effectively highlight visual contents of interest while capturing dynamics over short periods of time.
% We also observe that the regularization penalty parameter $\lambda$ varies across different datasets, for example, $\lambda = 2.5$ for FineGym and $0.001$ for HMDB-51. 
More visualizations and discussions can be found in Appendix~\ref{sec:appl-vis}.

\noindent\textbf{Generic action recognition.} Our evaluations on HMDB-51 are summarized in Table~\ref{tab:combined} ({\textit{left}}). Using TimeSformer as the backbone and integrating our motion prompt layer, we achieve accuracy improvements of 1.5\%, 1.2\%, and 0.7\% for split 1, 2 and 3, respectively. On average, this results in a performance gain of 1.1\%. 
% Our VMPs show consistent improvements across various recent action recognition backbones, including SlowFast and X3D.
Our VMPs show consistent improvements across various recent action recognition backbones.

\noindent\textbf{Fine-grained action recognition.}  In Table~\ref{tab:combined} ({\textit{right}}), we report performance gains on MPII Cooking 2, with Top-1 mean average precision improvements of 3.9\% for SlowFast, 2.1\% for C2D, 3.0\% for I3D, 0.7\% for X3D, and 6.0\% for TimeSformer. For FineGym, TimeSformer shows a 0.8\% increase in Top-1 accuracy. The TimeSformer backbone consistently outperforms the X3D backbone on both datasets, indicating that model performance and the benefits of VMPs depend on the backbone's ability to handle motion.

% \noindent\textbf{Video anomaly detection.}

% \begin{table}[ht]
% \centering
% \begin{tabular}{clcclccl}
% \hline
% \multirow{2}{*}{Method} &  & \multicolumn{5}{c}{UCF-Crime} &  \\ \cline{3-7}
%                         &  & AUC		     & AUC\_sub     &  & AP	     & AP\_sub    &  \\ \hline
% MULDE                   &  &   87.24            &    71.71            &  &   36.26             &  38.13            &  \\ \hline
% + prompt (ours)                   &  &               &               &  &               &              &  \\ \hline
% \end{tabular}
% \caption{The performance of MULDE~\cite{zhou2023batchnorm} on UCF-Crime and XD-Violence.}
% \label{table3}
% \end{table}

% \subsection{Ablation}
% \label{sec:ablation}

% \noindent\textbf{End-to-end tuning \vs motion prompt layer tuning}

% \noindent\textbf{}

% \subsection{Discussion}
% \label{sec:dis}

% \noindent\textbf{The roles of motions.}

% \noindent\textbf{The roles of attentions.}

% \noindent\textbf{The roles of prompts.}

% \noindent\textbf{Performance \vs speed.}

\section{Conclusion}
\label{sec:concl}

We introduce \textit{video motion prompts} to enhance action recognition. We use a modified Sigmoid activation function with learnable slope and shift as a power normalization function on frame differencing maps to activate motions as attention maps. Additionally, we introduce the \textit{temporal attention variation regularization} term to generate more accurate and smooth motion prompts. We formalize the entire process as a single \textit{motion prompt layer} acting as an adapter, resulting in improved performance across various benchmarks and backbones.

\acks{Qixiang Chen conducted this research under the supervision of Lei Wang for his final year honors research project at ANU. 
He is a recipient of The Active Intelligence Research Challenge Award. 
This work was also supported by the NCI Adapter Scheme Q4 2023, the NCI National AI Flagship Merit Allocation Scheme, and the National Computational Merit Allocation Scheme 2024 (NCMAS 2024), with computational resources provided by NCI Australia, an NCRIS-enabled capability supported by the Australian Government.}

\bibliography{acml24}

% \pagebreak

\appendix

\section{Related Work}
\label{sec:related}

Below, we review closely related work on motion extraction, attention mechanisms, prompts, and adapter layers for video processing. We also highlight the significant differences between our work and these studies.

\noindent\textbf{Motion.} Optical flow computed between consecutive video frames is a widely used secondary input for video processing~\citep{two_stream, i3d_net, wang2024flow}, complementing the use of conventional videos~\citep{wang2018temporal, wang2021tdn}. Dynamic images, which record spatio-temporal information in a single frame, were introduced in \citep{Bilen_2016_CVPR}. A channel sampling method that combines the R, G, or B channels of consecutive frames into a single frame for better motion capture was proposed in \citep{kim2022capturing}. Recently, Taylor videos have been introduced to capture dominant motions in videos for action recognition~\citep{wang2024taylor}. 

Unlike these approaches, our motion prompts are (i) lightweight, with only two extra learnable parameters, (ii) dependent on motions guided by frame differencing maps, and (iii) driven by attention mechanisms that highlight the spatio-temporal motion regions over time.

\noindent\textbf{Attention.} To improve feature representations, attention mechanisms capture the relationships between tokens. Attention mechanisms have been efficiently incorporated into transformers, including self-attention~\citep{vaswani2017attention, dosovitskiy2021an} and cross-attention~\citep{DBLP:journals/corr/abs-2106-05786, https://doi.org/10.48550/arxiv.2204.00452, DBLP:journals/corr/abs-2103-14899, Wei_2020_CVPR, wang20233mformer}, among others. 

Our attention mechanism, however, differs from traditional approaches. It is lightweight and does not require the learning of attention matrices. We use a sequence of frame differencing maps to record video dynamics, along with a newly introduced regularization term, to learn spatio-temporally smooth attention maps. Consequently, our attention mechanism is motion-dependent rather than dataset-dependent. Furthermore, compared to existing works, we use an activation function as power normalization to modulate motions, with learnable slope and shift parameters for controlling motion strengths and thresholding. This makes our attention mechanism more transparent and interpretable.

\noindent\textbf{Prompt.} Prompt engineering has gained significant interest with advancements in image and video processing. Representative works include the use of prompt templates~\citep{radford2021learning}, textual prompts for video content description~\citep{hu2022mage}, incorporating video features into language models as prompts~\citep{yang2022zero}, and learnable continuous prompt vectors as virtual tokens~\citep{ju2022prompting}. Visual prompts include methods such as visual prompt tuning~\citep{jia2022visual}, fine-grained visual prompting~\citep{yang2024fine}, among others.

To the best of our knowledge, none of these works consider using a sequence of refined motions as motion prompts to enhance video processing tasks. Our motion prompts are defined as a sequence of video frames with highlighted spatio-temporally smooth motion regions per frame. 
Our motion prompts are learnable and form a plug-and-play motion prompt layer. They are optimized using the original loss function.

\noindent\textbf{Adapter.} There are several layers and mechanisms in neural networks that can be considered as adapters between the data input and the model itself. Embedding layers, commonly used in NLP and recommendation systems, convert categorical data or tokens into dense vectors of fixed size~\citep{hrinchuk2019tensorized,wu2019stochastic}. Positional encoding, used in transformer models, adds positional information to input embeddings to retain the order of elements in sequences~\citep{vaswani2017attention, dosovitskiy2021an}. Normalization layers, such as Batch Normalization and Layer Normalization, stabilize and accelerate training by normalizing input data~\citep{ioffe2015batch,ba2016layer}. 
Adapter layers, often used in fine-tuning pre-trained models, add task-specific layers between existing layers to adapt pre-trained models to new tasks~\citep{Lee2020Mixout:,ding2022delta}. These adapter layers and mechanisms serve as intermediaries that process and transform input data, making it compatible with the model architecture. This enhances the model's ability to learn from and make predictions on the input data effectively. 

Our motion prompt layer functions as an adapter as well, bridging the gap between `blind motion extraction' and motion prompt-guided motion extraction.

\section{Preliminary}
\label{sec:appl-pre}

Below, we refer to the preliminary works used in the paper.

\noindent\textbf{Activation function.} Activation functions such as Sigmoid, Tanh, ReLU, and Softmax are among the most popular and commonly used in neural networks. These functions can be either linear or non-linear, depending on their formulation and the context in which they are applied. A review of activation functions in deep learning can be found in \citep{nwankpa2018activation, dubey2022activation}. A logistic function is a common $S$-shaped (sigmoid) curve with the equation:
\begin{equation}
    f(x) = \frac{L}{1 + e^{-a(x - b)}}
\end{equation}
where $L$ is the carrying capacity, $a$ is the logistic growth rate (the steepness of the curve), and $b$ is the point at which the output transitions from below $L/2$ to above $L/2$. For values of $x$ much less than $b$, $f(x)$ is close to 0 (or the lower bound), and for values of $x$ much greater than $b$, $f(x)$ is close to $L$ (or the upper bound). This characteristic makes $b$ act as a kind of threshold in the Sigmoid function. The standard logistic function is when $L=1$, $a=1$, and $b=0$.

Both the Sigmoid and Softmax functions introduce non-linearity; however, the Softmax function additionally provides a means to interpret the neural network's output as probabilities. In attention mechanisms, the Softmax function is applied to a set of scores (often called attention scores or logits) to produce a probability distribution over the elements in the sequence. This ensures that the importance (attention) weights sum to 1.

\noindent\textbf{A mathematical view of attention.} The Vision Transformer (ViT) applies the attention mechanism to image processing by first dividing an image into $n$ patches and then treating these patches as input tokens $\mX \in \mbr{n\times d}$. The self-attention mechanism involves three main components: query $\mQ$, key $\mK$ and value $\mV$ matrices, which are computed as linear projections of the input matrix $\mX$. The self-attention scores are then computed as the scaled dot-product of the query and key matrices, and the resulting attention matrix is then used to weight the value matrix. The mechanism can be described as follows:
\begin{align}
    & \mA = \text{softmax}\left(\frac{\mathbf{Q} \mathbf{K}^\top}{\sqrt{d_k}}\right), \label{eq:softm}\\
    & \mZ = \mA \mV, \label{eq:attn-v}
\end{align}
where $\mA \in \mbr{n\times n}$ is the attention matrix representing the attention weights, and $\mZ \in \mbr{n\times d_k}$ is the output of the self-attention layer. $\mQ=\mX\mW^Q$, $\mK=\mX\mW^K$, and $\mV=\mX\mW^V$. $\mW^Q, \mW^K, \mW^V \in \mbr{d\times d_k}$ are learned projection matrices. In practice, multi-head attention is used to allow the model to jointly attend to information from different representation subspaces at different positions. This mechanism enables the ViT to effectively capture the relationships between different parts of an image \citep{dosovitskiy2021an}. In our work, we design a lightweight attention mechanism to highlight the motions of interest from a sequence of frame differencing maps.

\noindent\textbf{Power normalization family.} Power Normalization (PN) is used to adjust the power or amplitude of signals to a standard or desired level. It is commonly used in signal and image processing, as well as in statistical methods such as non-linear pooling of features~\citep{jegou2009burstiness,koniusz2021power} and optical flow correction~\citep{wang2024flow}. We apply power normalizing functions to enhance or reduce motions in a sequence of frame differencing maps. A review of well-behaved PN functions such as Gamma, MaxExp, AsinhE (Arcsin hyperbolic function) and SigmE (Logistic a.k.a. Sigmoid functions) can be found in~\citep{koniusz2021power}. 

\section{Boundedness and Differentiability}
\label{sec:appl-proof}

Below we proof the boundedness and differentiability for both $a(m)$ and $b(n)$.

\begin{tcolorbox}[ width=1.0\linewidth, colframe=blackish, colback=beaublue, boxsep=0mm, arc=3mm, left=1mm, right=1mm, top=1mm, bottom=1mm]

\noindent\textbf{Boundedness.} The hyperbolic tangent function $\tanh(m)$ has a range of $(-1, 1)$. Therefore, $\beta\left|\tanh(m)\right|$ has a range of $(0, \beta)$. Adding a positive constant $\epsilon > 0$, the range of $\beta\left|\tanh(m)\right| + \epsilon$ is $(\epsilon, \beta + \epsilon)$. Thus, $a(m)$, which is $\frac{\alpha}{\beta\left|\tanh(m)\right| + \epsilon}$, ranges from $\frac{\alpha}{\beta + \epsilon}$ to $\frac{\alpha}{\epsilon}$. This means $a(m)$ is bounded between $\frac{\alpha}{\beta + \epsilon}$ and $\frac{\alpha}{\epsilon}$. 

Similarly, $\tanh(n)$ has a range of $(-1, 1)$ and $b(n) = \gamma \tanh(n)$ has a range of $(-\gamma, \gamma)$. This means $b(n)$ is bounded.

\noindent\textbf{Differentiability.} To show differentiability, we compute the derivative of $a(m)$ with respect to $m$: $a'(m) = -\frac{\alpha g'(m)}{[g(m)]^2}$,
where $g(m) = \beta\left|\tanh(m)\right| + \epsilon$.
The derivative of the hyperbolic tangent function is 
$\tanh'(m) = 1 - \tanh^2(m)$.
The derivative of $|\tanh(m)|$ is generally given by
$\frac{d}{dm}|\tanh(m)| = \text{sgn}(\tanh(m)) \cdot \tanh'(m) = \text{sgn}(\tanh(m)) \cdot (1 - \tanh^2(m))$. Therefore, $g'(m)$ is:
$g'(m) = \beta \cdot \text{sgn}(\tanh(m)) \cdot (1 - \tanh^2(m))$.
Thus, the derivative of $a(m)$ is:
$a'(m) = -\frac{\alpha \cdot \beta \cdot \text{sgn}(\tanh(m)) \cdot (1 - \tanh^2(m))}{(\beta\left|\tanh(m)\right| + \epsilon)^2}$,
where $\text{sgn}(\tanh(m))$ is the sign function, which is 1 for $\tanh(m) \geq 0$ and $-1$ for $\tanh(m) < 0$.

Similarly, we compute the derivative of $b(n)$ with respect to $n$: $b'(n) = \gamma \cdot \tanh'(n) = \gamma \cdot (1 - \tanh^2(n))$. Since $\tanh(n)$ is differentiable, $b(n)$ is also differentiable, and its derivative is given by: $b'(n) = \gamma \cdot (1 - \tanh^2(n))$.

\end{tcolorbox}

\section{Parameter Constraints and Sensitivity Analysis}
\label{sec:suppl-proof}

In this section, we explore the relationships among $\alpha$, $\beta$ and $\gamma$. We then present a sensitivity analysis of how these parameters affect our power normalization function,  and illustrate the selection process for these parameters.

\noindent\textbf{Constraints on $\alpha$, $\beta$, and $\gamma$.} Given the function $f(\mD) = \frac{1}{1 + e^{-\left(\frac{\alpha}{\beta|\tanh(m)| + \epsilon}\right) (\mD - \gamma \tanh(n))}}$, we need to ensure that $f(\mD)$ has a minimum value of 0 and a maximum value of 1 over the interval $[-1, 1]$. For simplicity, assume $0 < \gamma < 1$, $\alpha>0$ and $\beta>0$.

To satisfy $f(1) \approx 1$:
\begin{align}
    & \left(\frac{\alpha}{\beta|\tanh(m)| + \epsilon}\right) (1 - \gamma \tanh(n)) \to +\infty \nonumber \\
    & \left(\frac{\alpha}{\beta|\tanh(m)| + \epsilon}\right) (1 - \gamma \tanh(n)) \gg 1
\end{align}

In the worst-case scenario where $|\tanh(m)| = 1$ and $\tanh(n) = 1$, it follows:
\begin{align}
    & \left(\frac{\alpha}{\beta + \epsilon}\right) (1 - \gamma) \gg 1 \nonumber \\
    & \quad \quad \quad \quad \!\! \alpha (1 - \gamma) \gg \beta + \epsilon \nonumber \\
    & \quad \quad \quad \quad \quad \quad \quad \!\! \alpha \gg \frac{\beta + \epsilon}{1 - \gamma} \label{eq:cond1}
\end{align}

To ensure $f(-1) \approx 0$:
\begin{align}
    & \left(\frac{\alpha}{\beta|\tanh(m)| + \epsilon}\right) (-1 - \gamma \tanh(n)) \to -\infty \nonumber \\
    & \left(\frac{\alpha}{\beta|\tanh(m)| + \epsilon}\right) (-1 - \gamma \tanh(n)) \ll -1
\end{align}

In the worst-case scenario where $|\tanh(m)| = 0$ and $\tanh(n) = -1$, it follows:
\begin{align}
    & \frac{\alpha}{\epsilon} \cdot (-1 + \gamma) \ll -1 \nonumber \\
    & \quad \quad \!\!\!\!\!\! \alpha (-1 + \gamma) \ll -\epsilon \nonumber \\
    & \quad \quad \quad \quad \quad \quad  \!\!\!\!\!\!\! \alpha \gg \frac{-\epsilon}{-1 + \gamma} \label{eq:cond2}
\end{align}

From Eq.~\eqref{eq:cond1} and~\eqref{eq:cond2}, to ensure a practical and feasible relationship between $\alpha$, $\beta$, and $\gamma$, we derive:
\begin{align}
    & \alpha \gg \max \biggl\{\frac{\beta + \epsilon}{1 - \gamma}, \frac{-\epsilon}{-1 + \gamma} \biggr\} \nonumber \\
    & \alpha = k \cdot \max \biggl\{\frac{\beta + \epsilon}{1 - \gamma}, \frac{-\epsilon}{-1 + \gamma} \biggr\}
    \label{eq:constraints}
\end{align}
where $k \gg 1$ is a constant ensuring that $\alpha$ is sufficiently large.

\noindent\textbf{Sensitivity analysis on $\alpha$, $\beta$, and $\gamma$.} In this section, we analyze the sensitivity of the function $f(\mD)$ with respect to the parameters $\alpha$, $\beta$, and $\gamma$. Given the function $f(\mD) = \frac{1}{1 + e^{-\left(\frac{\alpha}{\beta|\tanh(m)| + \epsilon}\right) (\mD - \gamma \tanh(n))}}$, we compute the partial derivatives to understand how small changes in these parameters affect the value of $f(\mD)$. First, let $g(\mD) = \left(\frac{\alpha}{\beta|\tanh(m)| + \epsilon}\right) (\mD - \gamma \tanh(n))$, we calculate the partial derivative of $f(\mD)$ with respect to $\alpha$:
\begin{align}
    & \frac{\partial f(\mD)}{\partial \alpha} = \frac{\partial}{\partial \alpha} \left( \frac{1}{1 + e^{-g(\mD)}} \right), \nonumber \\
    & \frac{\partial f(\mD)}{\partial \alpha} = \frac{e^{-g(\mD)}}{\left(1 + e^{-g(\mD)}\right)^2} \cdot \frac{\partial g(\mD)}{\partial \alpha} \nonumber
\end{align}

Now, calculate $\frac{\partial g(\mD)}{\partial \alpha}$:

\begin{equation}
    \!\!\!\!\!\!\!\!\!\!\!\!\!\!\!\!\!\! \frac{\partial g(\mD)}{\partial \alpha} = \frac{\mD - \gamma \tanh(n)}{\beta|\tanh(m)| + \epsilon} \nonumber
\end{equation}

Thus,
\begin{equation}
    \quad \quad \quad \quad\! \frac{\partial f(\mD)}{\partial \alpha} = \frac{e^{-g(\mD)}}{\left(1 + e^{-g(\mD)}\right)^2} \cdot \frac{\mD - \gamma \tanh(n)}{\beta|\tanh(m)| + \epsilon} \label{eq:alpha}
\end{equation}

Similarly, we calculate the partial derivative of $f(\mD)$ with respect to $\beta$ and $\gamma$:

\begin{align}
    & \frac{\partial f(\mD)}{\partial \beta} = \frac{e^{-g(\mD)}}{\left(1 + e^{-g(\mD)}\right)^2} \cdot \left(-\frac{\alpha(\mD - \gamma \tanh(n))|\tanh(m)|}{(\beta|\tanh(m)| + \epsilon)^2}\right)  \\
    & \frac{\partial f(\mD)}{\partial \gamma} = \frac{e^{-g(\mD)}}{\left(1 + e^{-g(\mD)}\right)^2} \cdot \left( -\frac{\alpha \tanh(n)}{\beta|\tanh(m)| + \epsilon} \right)
\end{align}

The parameter $\alpha$ scales the exponent inside the Sigmoid function of $f(\mD)$, defined as $g(\mD)$. Increasing $\alpha$ steepens the curve of $f(\mD)$, causing it to approach 1 more rapidly for positive $\mD - \gamma \tanh(n)$, and approach 0 more rapidly for negative $\mD - \gamma \tanh(n)$. Conversely, $\alpha$ makes $f(\mD)$ change more gradually. 
Note that while $\alpha$ affects how quickly $f(\mD)$ transitions from 0 to 1, it does not directly alter the learning process of $m$ and $n$. However, the steepness can indirectly affect the gradients during optimization.
The parameter $\beta$ affects the sensitivity of $f(\mD)$ by influencing the denominator inside $g(\mD)$. As $\beta$ increases, the term $\beta|\tanh(m)| + \epsilon $ in the denominator increases, making $g(\mD)$ smaller. Consequently, $f(\mD)$ changes more gradually. Decreasing $\beta$ makes the denominator smaller, causing $g(\mD)$ to increase and thereby making $f(\mD)$ transition more sharply. Similar to $\alpha$, $\beta$ affects the sensitivity of $f(\mD)$ but does not directly control the learning of $m$ and $n$. % It adjusts how $f(\mD)$ responds to variations in $m$ and $n$ but does not affect the gradients of $m$ and $n$ themselves.
The parameter $\gamma$ controls the horizontal shift of $f(\mD)$. Increasing $\gamma$ shifts $g(\mD)$ to the right by increasing $\gamma \tanh(n)$, thereby reducing $f(\mD)$ for a given $\mD$. Conversely, decreasing $\gamma$ shifts $g(\mD)$ to the left, increasing $f(\mD)$ for a given $\mD$. Therefore, $\gamma$ adjusts the position at which $f(\mD)$ transitions from 0 to 1, affecting the predictions but not directly influencing the learning process of $m$ and $n$. 

Below we show the derivatives with respect to $m$ and $n$:

\begin{align}
    & \frac{\partial f(\mD)}{\partial m} = \frac{e^{-g(\mD)}}{\left(1 + e^{-g(\mD)}\right)^2} \cdot \frac{\alpha \beta \tanh'(m) \cdot (\mD - \gamma \tanh(n))}{(\beta |\tanh(m)| + \epsilon)^2} \label{eq:der-m}\\ 
    & \frac{\partial f(\mD)}{\partial n} = \frac{e^{-g(\mD)}}{\left(1 + e^{-g(\mD)}\right)^2} \cdot \frac{-\alpha \gamma \cdot \tanh'(n)}{\beta |\tanh(m)| + \epsilon} \label{eq:der-n}
\end{align}

While $\alpha$, $\beta$ and $\gamma$ play crucial roles in shaping $f(\mD)$ and its behavior, their impact on the learning of $m$ and $n$ is indirect. The learning process for $m$ and $n$ depends more directly on how well 
$f(\mD)$ fits the video data and the gradients of $m$ and $n$ with respect to the loss function. Therefore, while tuning $\alpha$, $\beta$ and $\gamma$ is important for optimizing $f(\mD)$, their influence on the learnability of $m$ and $n$ is secondary to the fitting and optimization process itself. 

\noindent\textbf{Parameter selection and analysis.} Considering the constraints (Eq.~\eqref{eq:constraints}), sensitivity analysis (Eq.~\eqref{eq:alpha} --~\eqref{eq:der-n}) and optimisation (Eq.~\eqref{eq:final-loss} as the loss function), we simply choose $\beta=0.45$ and $\gamma=0.6$ (small constant $\epsilon=0.1$ as in the main paper):
\begin{align}
    & \alpha = k \cdot \max \biggl\{\frac{0.45 + 0.1}{1 - 0.6}, \frac{-0.1}{-1 + 0.6} \biggr\} \nonumber \\
    & \quad \!\!= k \cdot \max \{1.375, 0.25\} \nonumber \\
    & \quad \!\! = 1.375k 
\end{align}
We set $\alpha=5$ ($k\approx3.6$).
\begin{tcolorbox}[ width=1.0\linewidth, colframe=blackish, colback=beaublue, boxsep=0mm, arc=3mm, left=1mm, right=1mm, top=1mm, bottom=1mm]
\noindent\textbf{The roles of $\alpha$, $\beta$, and $\gamma$.} We now provide a detailed analysis of the selected $\alpha$, $\beta$, and $\gamma$ values. 
With $\alpha=5$ and $\beta=0.45$ in our power normalization function (Eq.~\eqref{eq:final-sig}): if the learnable $|\tanh(m)|=0$, the slope $a(m)$ (Eq.~\eqref{eq:ab-sigmoid}) becomes the steepest, leading to significant changes around the learnable shift $b(n)$; conversely, if $|\tanh(m)|=1$, the slope is the most gentle. This pair of parameters allows flexible adjustment of the slope, which defines the strength of motion modulation, through the learnable and well-bounded $|\tanh(m)|$, resulting in a wide slope range from 9.09 to 50 (see Fig.~\ref{fig:pn-comp} (b)).

On the other hand, parameter $\gamma=0.6$ controls the maximum shift for the threshold (Eq.~\eqref{eq:ab-sigmoid}, and also see Fig.~\ref{fig:pn-comp} (c)). The learnable $\tanh(n)$, ranging from -1 to 1, determines the degree (\eg, $\tanh(n)=1$ indicates a maximum rightward shift of 0.6) and the direction (\eg, positive for right, negative for left) of shift. 
% If the pixel values in $\mD$ are greater than $0.6$, the motion should ideally be considered as useful and enhanced. Conversely, pixel values lower than $0.6$ should be dampened.

\noindent\textbf{The roles of the $\tanh$ function.} Therefore, the $\tanh(\cdot)$ function in our power normalization achieves several objectives: (i) it bounds the parameter search space of slope and shift between -1 and 1, allowing for free learning of $m$ and $n$, (ii) it effectively adjusts the strengths of motion modulation in slope, where $|\tanh(\cdot)|=0$ indicates the steepest slope and 1 indicates the gentlest slope, (iii) it controls the direction of shift based on the sign of $\tanh(\cdot)$, where a negative value indicates a left shift and a positive value indicates a right shift, and (iv) it adapts the threshold for motion modulation based on data properties, \eg, frame differencing maps, are also range from -1 and 1, allowing the shift to consider both positive and negative motions.
\end{tcolorbox}

We notice that this set of parameters gives reasonably good performance on both generic and fine-grained action recognition datasets, so we use it for evaluations across various action recognition datasets and model architectures.

\section{Additional Visualisations \& Discussions}
\label{sec:appl-vis}

\subsection{Exploring the Impact of Video Motion Prompts (VMPs)}

\begin{table}[tbp]
\centering
\setlength{\tabcolsep}{6.18pt}
% \caption{Variant study of finetuning on MPII Cooking 2 using TimeSformer. \textcolor{cyan}{\ding{100}} denotes frozen, and \textcolor{red}{\ding{89}} denotes finetuning/training. For example, [Pretrained\textcolor{cyan}{\ding{100}}\textcolor{red}{\ding{89}}\textcolor{cyan}{\ding{100}}\textcolor{red}{\ding{89}}+VMPs\textcolor{red}{\ding{89}}] represents the use of a Kinetics-600 pretrained model (Pretrained\textcolor{cyan}{\ding{100}}), which is then finetuned on MPII Cooking 2 (Pretrained\textcolor{cyan}{\ding{100}}\textcolor{red}{\ding{89}}, baseline). Then, this finetuned model (Pretrained\textcolor{cyan}{\ding{100}}\textcolor{red}{\ding{89}}\textcolor{cyan}{\ding{100}}) undergoes end-to-end finetuning (Pretrained\textcolor{cyan}{\ding{100}}\textcolor{red}{\ding{89}}\textcolor{cyan}{\ding{100}}\textcolor{red}{\ding{89}}), including our motion prompt layer (+VMPs\textcolor{red}{\ding{89}}). We highlight improvements in red. Fig.~\ref{fig:layer-weights} in Appendix~\ref{sec:appl-vis} shows insightful comparisons of VMP-based finetuning models across various layers of TimeSformer.}
\caption{Variant study of fine-tuning on MPII Cooking 2 using a TimeSformer model pretrained on Kinetics-600. We evaluate model variants, including pretrained and baseline (finetuned without VMPs) models, with and without the use of VMPs. This setup explores all potential fine-tuning combinations. \textcolor{red}{\ding{89}} and \textcolor{cyan}{\ding{100}} denote fine-tuning and frozen states, respectively. `\textcolor{red}{\ding{89}}Baseline' denotes a second round of fine-tuning. We highlight improvements in red. Fig.~\ref{fig:layer-weights} shows insightful comparisons of VMPs-based finetuning models across various layers of TimeSformer.}
% \resizebox{0.95\linewidth}{!}{\begin{tabular}{lcccccc}
% \toprule
% &  \multirow{2}{*}{[Pretrained\textcolor{cyan}{\ding{100}}]} & [Pretrained\textcolor{cyan}{\ding{100}} & Baseline & [Pretrained\textcolor{cyan}{\ding{100}}\textcolor{red}{\ding{89}} & [Pretrained\textcolor{cyan}{\ding{100}}\textcolor{red}{\ding{89}}\textcolor{cyan}{\ding{100}} & [Pretrained\textcolor{cyan}{\ding{100}}\textcolor{red}{\ding{89}}\textcolor{cyan}{\ding{100}}\textcolor{red}{\ding{89}}\\
% &  & +\textbf{VMPs}\textcolor{red}{\ding{89}}] & ([Pretrained\textcolor{cyan}{\ding{100}}\textcolor{red}{\ding{89}}]) & +\textbf{VMPs}\textcolor{red}{\ding{89}}] & +\textbf{VMPs}\textcolor{red}{\ding{89}}] & +\textbf{VMPs}\textcolor{red}{\ding{89}}] \\
% \midrule
% Top-1 & 36.6 & \textbf{37.1}$^{\textbf{\textcolor{red}{$\uparrow$0.5}}}$ & 50.6 & \textbf{56.6}$^{\textbf{\textcolor{red}{$\uparrow$6.0}}}$ & \textbf{56.2}$^{\textbf{\textcolor{red}{$\uparrow$5.6}}}$ & \textbf{57.1}$^{\textbf{\textcolor{red}{$\uparrow$6.5}}}$ \\
% Top-5 & \textbf{66.9} & 66.2 $^{\textcolor{blue}{\downarrow0.7}}$ & 81.8 & \textbf{84.4}$^{\textbf{\textcolor{red}{$\uparrow$2.6}}}$ & \textbf{84.3}$^{\textbf{\textcolor{red}{$\uparrow$2.5}}}$ & \textbf{83.7}$^{\textbf{\textcolor{red}{$\uparrow$1.9}}}$ \\ 
% \bottomrule
\resizebox{0.95\linewidth}{!}{\begin{tabular}{lcccccc}
\toprule
    & \multirow{2}{*}{Pretrained} & \textbf{VMPs} +  & \multirow{2}{*}{Baseline} & \textbf{VMPs} + & \textbf{VMPs} + & \textbf{VMPs} + \\
    & & \textcolor{cyan}{\ding{100}}Pretrained & & \textcolor{red}{\ding{89}}Pretrained & \textcolor{cyan}{\ding{100}}Baseline & \textcolor{red}{\ding{89}}Baseline \\
\midrule
    Top-1 & 36.6 & \textbf{37.1}$^{\textbf{\textcolor{red}{$\uparrow$0.5}}}$ & 50.6 & \textbf{56.6}$^{\textbf{\textcolor{red}{$\uparrow$6.0}}}$ & \textbf{56.2}$^{\textbf{\textcolor{red}{$\uparrow$5.6}}}$ & \textbf{57.1}$^{\textbf{\textcolor{red}{$\uparrow$6.5}}}$ \\
    Top-5 & \textbf{66.9} & 66.2$^{\textcolor{blue}{\downarrow0.7}}$ & 81.8 & \textbf{84.4}$^{\textbf{\textcolor{red}{$\uparrow$2.6}}}$ & \textbf{84.3}$^{\textbf{\textcolor{red}{$\uparrow$2.5}}}$ & \textbf{83.7}$^{\textbf{\textcolor{red}{$\uparrow$1.9}}}$ \\ 
\bottomrule
\end{tabular}}
\label{tab:vmp-eval}
\end{table}

\noindent\textbf{The impact of VMPs on fine-tuning.} We explore four sets of experiments using Eq.~\eqref{eq:final-loss} as the loss function: (i) [VMPs+\textcolor{cyan}{\ding{100}}Pretrained] freezing the Kinetics-600 pretrained model and training only our motion prompt layer, (ii) [VMPs+\textcolor{red}{\ding{89}}Pretrained] finetuning the Kinetics-600 pretrained model, including the motion prompt layer, (iii) [VMPs+\textcolor{cyan}{\ding{100}}Baseline] freezing the finetuned baseline model and training only our motion prompt layer, and (iv) [VMPs+\textcolor{red}{\ding{89}}Baseline] end-to-end finetuning of the finetuned baseline model, including the motion prompt layer. We choose MPII Cooking 2 and use TimeSformer as the backbone. 

\begin{figure}[tbp]
\begin{center}
\includegraphics[width=\textwidth]{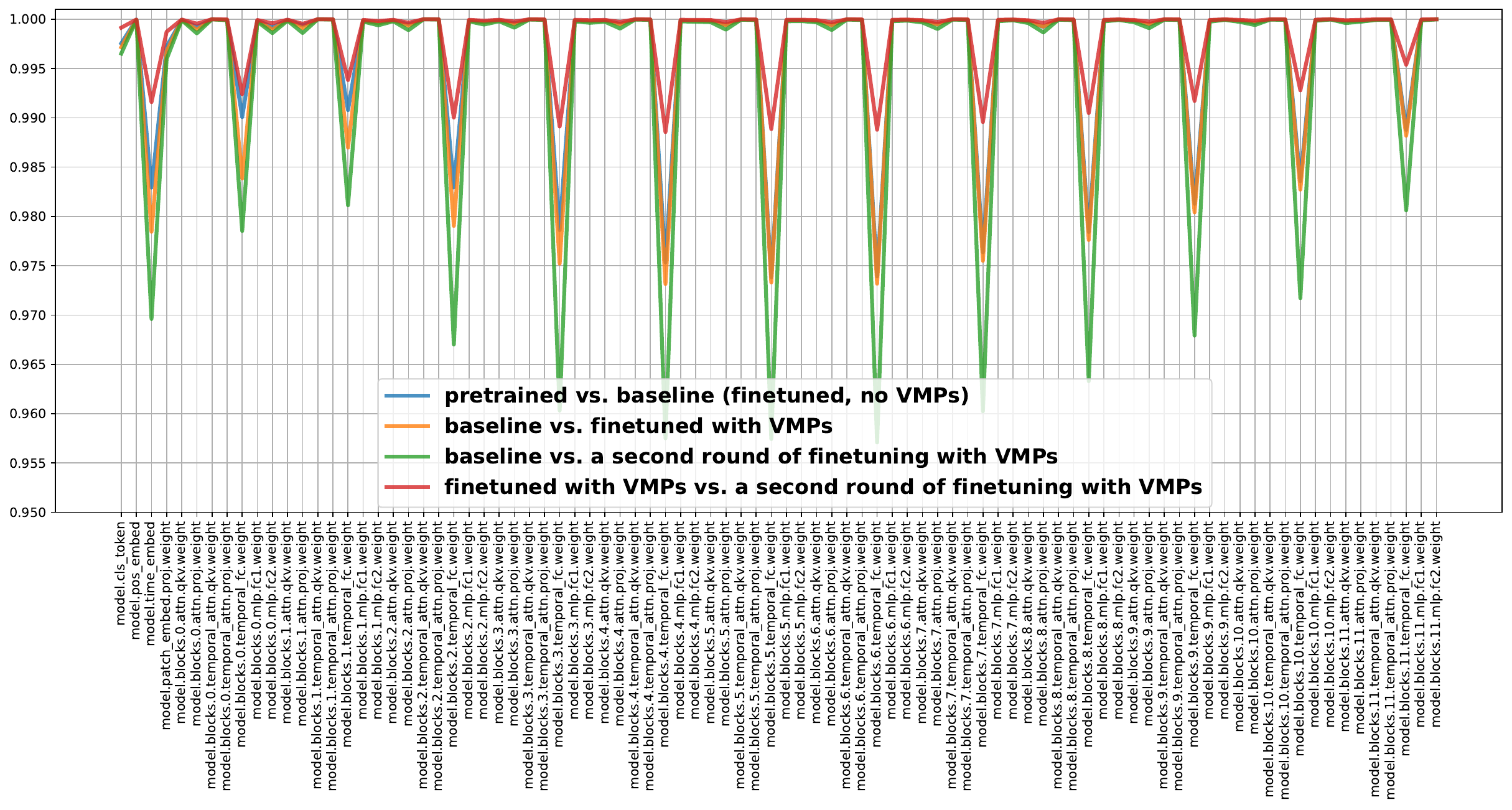}
\caption{Roles of VMPs in model finetuning via per-layer weight similarity comparison. We use TimeSformer pretrained on Kinetics-600 as the backbone, and finetuned on MPII Cooking 2 with or without VMPs. The vertical axis shows the similarity scores, ranging from 0 to 1, with higher scores indicating greater similarity between model weights. The horizontal axis displays the names of the backbone's layers. Note that the final projection layer has been removed to better display the other layers.
We use cosine similarity to evaluate the vectorized per-layer weights across different experimental setups: (i) (blue color) pretrained on Kinetics-600 \vs MPII Cooking 2 finetuned (baseline, no VMPs), (ii) (orange color) baseline \vs finetuned with VMPs, (iii) (green color) baseline \vs end-to-end VMP-based finetuning of the baseline model, and (iv) (red color) finetuned with VMPs \vs second-round VMP-based finetuning. 
Our VMP-based finetuning demonstrates significant effects across different aspects of model architecture. Firstly, it affects the weights of attention projection layers, using only two learnable parameters to subtly enhance focus. Secondly, VMPs impact FC layers dedicated to temporal information embedding, enriching the video processing task with nuanced motion cues. Thirdly, they affect initial layers such as tokens and time embeddings, acting as a adapter to refine the focus on motion and enhance the embeddings of motion-guided tokens and temporal concepts. These findings show the effectiveness of VMPs in optimizing model for action recognition tasks.}\label{fig:layer-weights}
\end{center}
\end{figure}

The evaluations are summarized in Table~\ref{tab:vmp-eval}.
We observe that using VMPs outperforms [Pretrained] by 0.5\% in terms of Top-1 accuracy.
% Interestingly, by freezing the Kinetics-600 pretrained model and learning only our VMPs, we obtain a 2.30\% Top-1 accuracy (compared to 1.54\% for random prediction on this dataset). Note that our VMPs have only 2 learnable parameters.
We also observe that fine-tuning the Kinetics-600 pretrained model with VMPs ([VMPs+\textcolor{red}{\ding{89}}Pretrained]), freezing the baseline model weights while training only the VMPs ([VMPs+\textcolor{cyan}{\ding{100}}Baseline]), and performing a second round of fine-tuning on the baseline model with VMPs ([VMPs+\textcolor{red}{\ding{89}}Baseline]) outperformed the baseline model (finetuned on MPII Cooking 2 without VMPs) by 6\%, 5.6\%, and 6.5\%, respectively.
% Our end-to-end finetuning of the finetuned baseline model with VMPs achieves the best Top-1 performance.
We include an intriguing plot (Fig.~\ref{fig:layer-weights}) that compares per-layer weight similarity among these models, showing the effects of VMP-based finetuning across various layers and/or blocks of TimeSformer.

\begin{table}[tbp]
\centering
\setlength{\tabcolsep}{1.0pt}
\caption{Evaluations of TimeSformer pretrained on different datasets for fine-tuning, with and without the motion prompt layer, on MPII Cooking 2. We highlight improvements in red.}
\resizebox{\linewidth}{!}{
    \begin{tabular}{l c ccccc c ccccc c ccccc c ccccc c}
    \toprule
        \multirow{2}{*}{Model} && \multicolumn{5}{c}{HowTo100M} && \multicolumn{5}{c}{SSv2} && \multicolumn{5}{c}{Kinetics-400} && \multicolumn{5}{c}{Kinetics-600} \\
        \cline{3-7} \cline{9-13} \cline{15-19} \cline{21-25}
        && \multicolumn{2}{c}{Top-1} && \multicolumn{2}{c}{Top-5} && \multicolumn{2}{c}{Top-1} && \multicolumn{2}{c}{Top-5} && \multicolumn{2}{c}{Top-1} && \multicolumn{2}{c}{Top-5} && \multicolumn{2}{c}{Top-1} && \multicolumn{2}{c}{Top-5} & \\
    \midrule
        TimeSformer && 50.4 &&& 82.2 &&& 53.7 &&& 81.4 &&& 50.0 &&& 79.4 &&& 50.6 &&& 81.8 &\\ 
        +\textbf{VMPs} && \textbf{54.5} & $^\textbf{\textcolor{red}{$\uparrow$4.1}}$ && \textbf{82.5} & $^\textbf{\textcolor{red}{$\uparrow$0.3}}$ && \textbf{55.6} & $^\textbf{\textcolor{red}{$\uparrow$1.9}}$ && \textbf{83.7} & $^\textbf{\textcolor{red}{$\uparrow$2.3}}$ && \textbf{55.2} & $^\textbf{\textcolor{red}{$\uparrow$5.2}}$ && \textbf{82.5} & $^\textbf{\textcolor{red}{$\uparrow$3.1}}$ && \textbf{56.6} & $^\textbf{\textcolor{red}{$\uparrow$6.0}}$ && \textbf{84.4} & $^\textbf{\textcolor{red}{$\uparrow$2.6}}$ & \\
    \bottomrule
    \end{tabular}
}
\label{tab:pretrain-all}
\end{table}

\noindent\textbf{VMPs in the context of various pretrained models.} Table~\ref{tab:pretrain-all} presents the results. The SSv2 pretrained model achieved the best results after being fine-tuned on the MPII Cooking 2 dataset. However, fine-tuning with the motion prompt layer significantly boosts performance for both Top-1 and Top-5 accuracy, particularly for models with lower baseline performance. Notably, the model pretrained on the Kinetics-600 dataset exhibits the most significant improvement (a 6\% increase in top-1 accuracy) when fine-tuning with the motion prompt layer. Overall, consistent improvements are observed across all pretrained models with the addition of the motion prompt layer.

\begin{table}[tbp]
\centering
\setlength{\tabcolsep}{5.0pt}
\caption{Cross-dataset evaluation with and without video motion prompts. Both top-1 and top-5 performance are reported, with improvements highlighted in red.}
\label{tab:zero-shot}
\resizebox{0.56\linewidth}{!}{
\begin{tabular}{lcccc}
\toprule
     & \multicolumn{2}{c}{FineGym to MPII} & \multicolumn{2}{c}{MPII to FineGym} \\
\cmidrule(lr){2-3} \cmidrule(lr){4-5}
     & Baseline & \textbf{+VMPs} & Baseline & \textbf{+VMPs} \\
\midrule
    Top-1 & 34.1 & \textbf{34.3}$^{\textbf{\textcolor{red}{$\uparrow$0.2}}}$ & \textbf{47.2} & 46.5$^{\textcolor{blue}{\downarrow0.7}}$ \\
    Top-5 & 63.0 & \textbf{64.1}$^{\textbf{\textcolor{red}{$\uparrow$1.1}}}$ & \textbf{84.3} & 82.6$^{\textcolor{blue}{\downarrow1.7}}$ \\
\bottomrule
\end{tabular}}
\label{tab:cross-eval}
\end{table}

\noindent \textbf{Evaluating VMPs across different datasets.} We evaluate the generalizability of finetuned TimeSformer+VMPs models on MPII-cooking 2 and FineGym datasets using unseen videos in Table~\ref{tab:cross-eval}. For the model finetuned on MPII-cooking 2, we use FineGym as the test set, training only the final fully connected (FC) layer for 5 epochs while keeping the other layer weights frozen. We also perform the evaluation in the reverse scenario. 

% As shown in the Table~\ref{tab:zero-shot}, the model trained on a moving camera scene (FineGym) and then adapted to a static camera scene (MPII Cooking 2) shows improved performance with the addition of VMPs. The Top-1 accuracy increases by 0.2\%, and the Top-5 accuracy improves by 1.1\%, demonstrating the ability of VMPs to enhance generalization from dynamic to static camera environments.

% Interestingly, when the model is trained on a static camera scene (MPII Cooking 2) and adapted to an unseen moving camera scene (FineGym), a slight performance drop is observed. Top-1 accuracy decreases by 0.7\%, and Top-5 accuracy declines by 1.7\%. This performance drop may be attributed to the challenges of adapting attention mechanisms learned from static camera scenes to moving camera scenarios, which introduce more complex spatiotemporal dynamics.

As shown in the table, the model trained on a moving camera scene and then adapted to a static camera scene (FineGym to MPII) demonstrates improved performance, with Top-1 and Top-5 accuracy being 0.24\% and 1.15\% higher, respectively. 

Interestingly, when the model is trained on a static camera scene and then adapted to an unseen moving camera scene (MPII to FineGym), performance slightly decreases, with Top-1 and Top-5 accuracy being 0.71\% and 1.78\% lower, respectively. This is likely due to the difficulty in adapting learned attention mechanisms from static camera videos to moving camera scenarios.

\begin{figure}[tbp]
\begin{center}
\includegraphics[width=\textwidth]{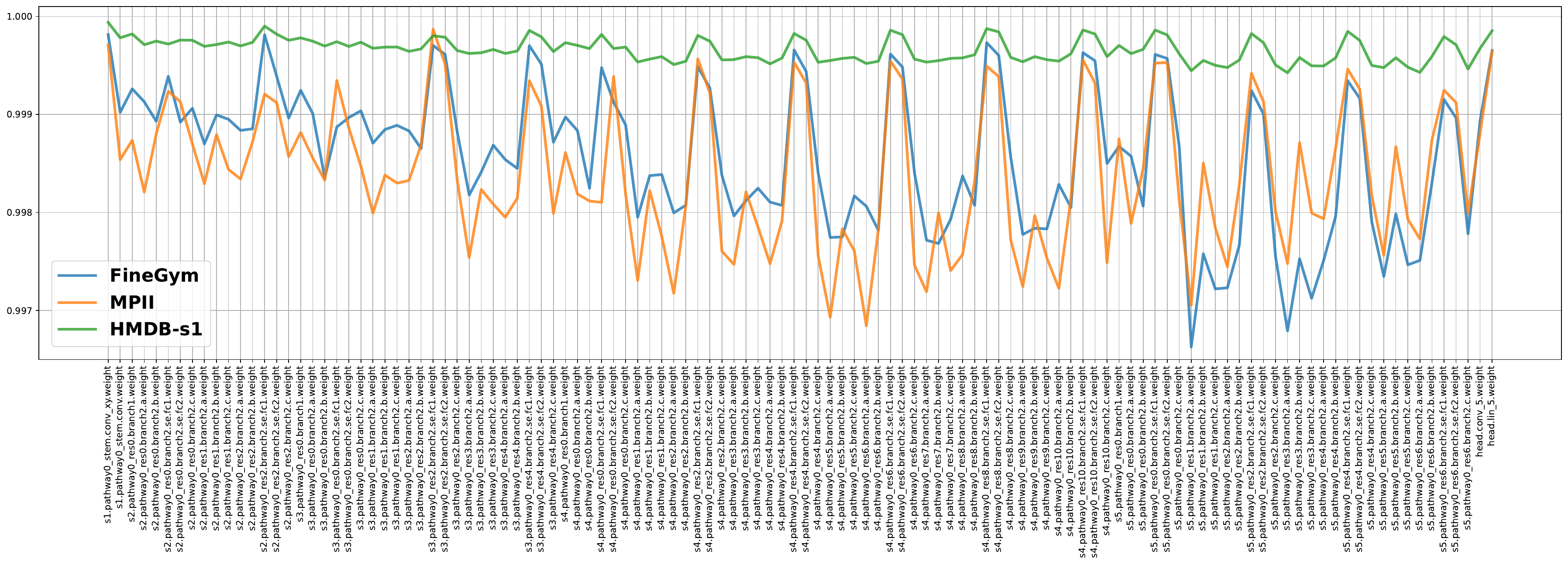}
\caption{Roles of VMPs in model finetuning via per-layer weight similarity comparison. We use X3D pretrained on Kinetics-600 as the backbone and finetune it on FineGym, MPII Cooking 2 (MPII), and HMDB-51 split 1 (HMDB-s1). We measure the per-layer weight similarity between baseline model (no VMPs) and model finetuned with our VMPs per dataset. The vertical axis depicts similarity scores ranging from 0 to 1, where higher scores indicate greater similarity between model weights. The horizontal axis lists the names of X3D's layers. Note that the final projection layer has been removed to better display the other layers. 
Our VMP-based finetuning demonstrates significant effects across different layers of X3D. We observe that on HMDB-s1, both the baseline and the models finetuned with VMPs show high similarities in weights between pairs of layers compared to FineGym and MPII. This difference arises because FineGym and MPII are designed for fine-grained action recognition, whereas Kinetics-600 shares similar action types with HMDB-s1, hence minor adjustments in weights are reasonable.
Interestingly, our VMPs notably influence the temporal modeling blocks of X3D. These findings highlight the effectiveness of VMPs in optimizing model for 3D CNN-based action recognition.
}\label{fig:layer-weights-x3d}
\end{center}
\end{figure}

\begin{figure}[tbp]
\begin{center}
\includegraphics[width=\textwidth]{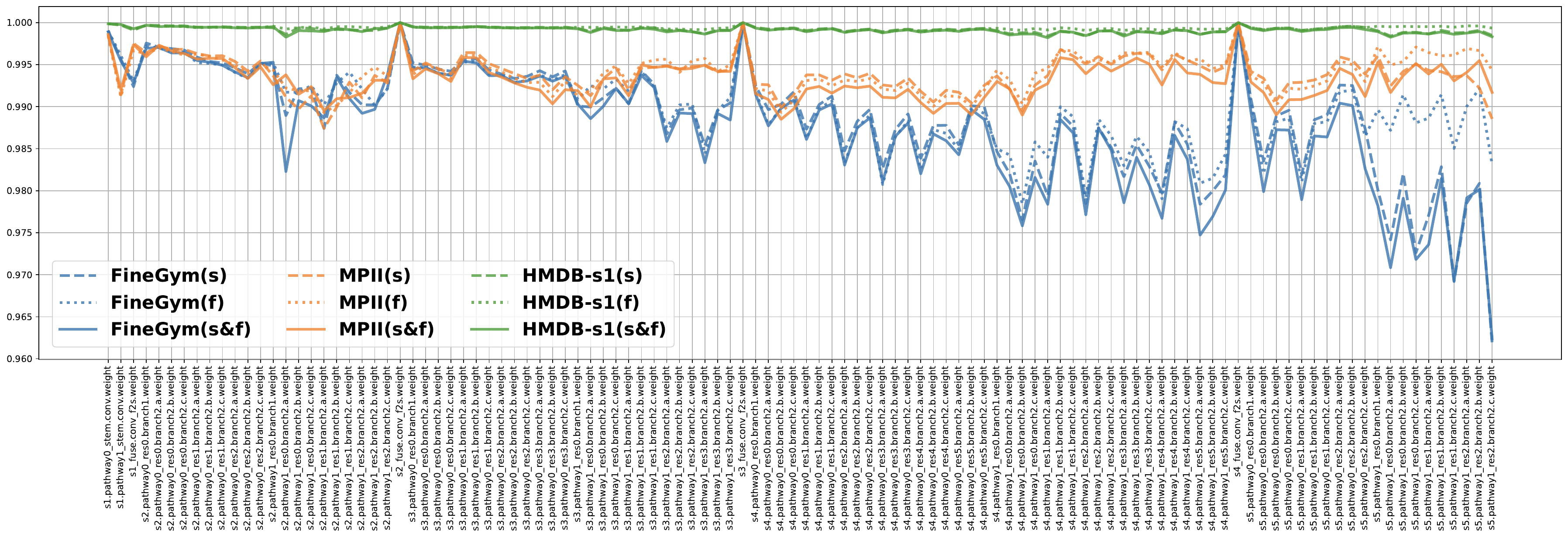}
\caption{Roles of VMPs in model finetuning via per-layer weight similarity comparison. We use SlowFast pretrained on Kinetics-600 as the backbone and finetune it on FineGym, MPII Cooking 2 (MPII), and HMDB-51 split 1 (HMDB-s1). We measure the per-layer weight similarity between baseline model (no VMPs) and model finetuned with our VMPs per dataset. We explore three variants: adding VMPs to (i) the slow stream only (s), (ii) the fast stream only (f), and (iii) both the slow and fast streams (s\&f). The vertical axis depicts similarity scores ranging from 0 to 1, where higher scores indicate greater similarity between model weights. The horizontal axis lists the names of SlowFast's layers. Note that the final projection layer has been removed to better display the other layers. 
Our VMP-based finetuning demonstrates significant effects across different layers of SlowFast. We observe that on HMDB-s1, both the baseline and the models finetuned with VMPs show high similarities in weights between pairs of layers compared to FineGym and MPII. This difference arises because FineGym and MPII are designed for fine-grained action recognition, whereas Kinetics-600 shares similar action types with HMDB-s1, hence minor adjustments in weights are reasonable.
Interestingly, our VMPs notably influence the temporal modeling blocks of SlowFast. These findings highlight the effectiveness of VMPs in optimizing model for 3D CNN-based action recognition.
}\label{fig:layer-weights-slowfast}
\end{center}
\end{figure}

\noindent\textbf{Analyzing VMPs in model fine-tuning through per-layer weight similarity.} We choose TimeSformer pretrained on Kinetics-600 as our backbone. We use cosine similarity to compare the per-layer weights between the baseline model (pretrained on Kinetics-600 and then finetuned on MPII Cooking 2) and each model variant finetuned with our VMPs. The purpose of this comparison is to assess the effects of our VMPs on model finetuning across different layers of the backbone network.

We conduct a detailed analysis on four pairs of experiments: (i) pretrained on Kinetics-600 \vs model finetuned on MPII Cooking 2 (baseline, no VMPs), (ii) baseline \vs finetuned with our VMPs, (iii) baseline \vs end-to-end finetuning of the finetuned baseline model with our VMPs (a second round of VMP-based finetuning), and (iv) finetuned with our VMPs \vs a second round of finetuning with VMPs. Fig.~\ref{fig:layer-weights} summarizes our results. Table~\ref{tab:vmp-eval} in the main paper presents the quantitative results on variant study of finetuning.

\begin{tcolorbox}[ width=1.0\linewidth, colframe=blackish, colback=beaublue, boxsep=0mm, arc=3mm, left=1mm, right=1mm, top=1mm, bottom=1mm]
Interestingly, we observe that our VMP-based finetuning: (i) influences the weights of each attention projection layer. It is worth noting that our VMPs only consist of two learnable parameters and subtly adjust the attention mechanism to enhance focus. (ii) impacts the weights of FC layers used for embedding temporal information. This highlights how our VMPs enrich the video processing task with nuanced motion cues. (iii) affects initial layers such as tokens and time embeddings. This demonstrates that our VMPs act as an adapter, refining not only the input video data's focus on motion but also the embeddings of motion-guided tokens and temporal concepts.
\end{tcolorbox}
We also observe that the similarity measures for the layer weights between the model finetuned with VMPs and the variant of a second round of finetuning with VMPs are higher than those of the other three experiments. Notably, both the model finetuned with VMPs and the second round of finetuning with VMPs achieve superior performance. This indicates that models tuned with VMPs tend to show very similar learned weights per layer, as evidenced by significantly higher similarities compared to the other three experiments. 

Fig.~\ref{fig:layer-weights-x3d} and ~\ref{fig:layer-weights-slowfast} present additional plots on all three action recognition datasets using X3D and SlowFast backbones. We also observed that our VMPs influence the temporal modeling blocks in these 3D CNN-based architectures.

\begin{table}[tbp]
\centering
\setlength{\tabcolsep}{5.0pt}
\caption{Evaluation of the VMPs layer's effectiveness under various video degradation conditions, including salt-and-pepper noise and reduced resolution from compression. This table presents the performance of the TimeSformer model pretrained on Kinetics-600, comparing scenarios with and without the motion prompt layer in both static and moving camera settings. The results show the improvements in action recognition performance facilitated by our motion prompts in the presence of degraded video quality.}
\label{tab:quality_eval}
\resizebox{0.91\linewidth}{!}{
\begin{tabular}{lccccccc}
\toprule
    &  & \multicolumn{2}{c}{Origin} & \multicolumn{2}{c}{Pepper-salt noise} & \multicolumn{2}{c}{Compression} \\ 
    \cline{3-8}
    &  & Baseline & \textbf{+VMPs} & Baseline & \textbf{+VMPs} & Baseline & \textbf{+VMPs} \\ 
\midrule
\multirow{2}{*}{FineGym}  
    & Top-1 & 83.6 & \textbf{84.4}$^{\textbf{\textcolor{red}{$\uparrow$0.8}}}$ & \textbf{83.1} & \!\!\!\textbf{83.1}$\quad$ & 79.7 & \textbf{80.5}$^{\textbf{\textcolor{red}{$\uparrow$0.8}}}$ \\ 
    & Top-5 & \textbf{98.7} & 98.5$^{\textcolor{blue}{\downarrow0.2}}$ & \textbf{98.4} & 98.3$^{\textcolor{blue}{\downarrow0.1}}$ & \textbf{97.7} & 97.6$^{\textcolor{blue}{\downarrow0.1}}$ \\ 
\midrule
\multirow{2}{*}{MPII}      
    & Top-1 & 50.6 & \textbf{56.6}$^{\textbf{\textcolor{red}{$\uparrow$6.0}}}$ & 51.2 & \textbf{53.2}$^{\textbf{\textcolor{red}{$\uparrow$2.0}}}$ & 48.1 & \textbf{52.8}$^{\textbf{\textcolor{red}{$\uparrow$4.7}}}$ \\ 
    & Top-5 & 81.8 & \textbf{84.4}$^{\textbf{\textcolor{red}{$\uparrow$2.6}}}$ & 82.0 & \textbf{83.5}$^{\textbf{\textcolor{red}{$\uparrow$1.5}}}$ & 78.6 & \textbf{80.1}$^{\textbf{\textcolor{red}{$\uparrow$1.5}}}$ \\
\bottomrule
\end{tabular}}
\label{tab:video-qual}
\end{table}

\noindent\textbf{VMPs in addressing extreme variations in video quality and camera movements.} We evaluate the effectiveness of the VMPs layer under various conditions by synthesizing low-quality videos in two distinct ways. First, we introduce salt-and-pepper noise, with a probability of 0.01 for each pixel to display either salt or pepper characteristics. Second, we apply a compression algorithm that reduces the original video quality by half, then upsample it back to the original input size to reduce resolution. This dual approach allows us to assess the performance enhancements provided by the VMPs layer in handling different types of video degradation.

We finetune the TimeSformer pretrained on Kinetics-600 with and without our motion prompt layer. The model without the motion prompt layer serves as the baseline. Below, we present experimental results and analysis regarding video quality and camera movements. Table~\ref{tab:video-qual} summarizes the results. We observe that our video motion prompts lead to improved performance in both static and moving camera scenes, even in the presence of low-resolution and noisy videos.

Our video motion prompts are computed based on the frame differencing maps; therefore, (i) color variations, such as dark and bright pixels, (ii) video quality, and (iii) camera movements can affect the quality of the motion prompts. Our future work will focus on exploring the use of motion prompts at the patch level. For instance, we plan to divide video frames into equal-sized image patches and apply motion prompts to each patch individually. Additionally, we will investigate integrating our motion prompts as intermediate layers within the network to modulate feature maps or feature vectors. This approach aims to address issues related to extreme variations in videos and camera movements.

% \begin{table}[ht]
% \centering
% \setlength{\tabcolsep}{3.0pt}
% \caption{Performance comparison for different actions and objects}
% \resizebox{0.95\linewidth}{!}{
% \begin{tabular}{lcccccc}
% \toprule
% Action & Model & Carrot & Cucumber & Onion & Tomato & Accuracy \\
% \midrule
% \multirow{2}{*}{cut dice}  & Baseline & 0.0 & 0.0 & 0.0 & - & 0.0 \\ 
%                            & +VMPs    & 0.0 & \textbf{21.1} & 0.0 & - & \textbf{6.0} \\ 
% \midrule
% \multirow{2}{*}{gather}    & Baseline & 0.0 & 0.0 & \textbf{50.0} & - & 11.1 \\ 
%                            & +VMPs    & \textbf{33.3} & 0.0 & \textbf{50.0} & - & \textbf{33.3} \\ 
% \midrule
% \multirow{2}{*}{slice}     & Baseline & 0.0 & 0.0 & \textbf{50.0} & 0.0 & 7.7 \\ 
%                            & +VMPs    & \textbf{80.0} & \textbf{40.0} & \textbf{50.0} & 0.0 & \textbf{53.9} \\ 
% \midrule
% \multirow{2}{*}{wash}      & Baseline & \textbf{100.0} & \textbf{100.0} & - & \textbf{100.0} & \textbf{100.0} \\ 
%                            & +VMPs    & \textbf{100.0} & \textbf{100.0} & - & \textbf{100.0} & \textbf{100.0} \\
% \bottomrule
% \end{tabular}}
% \end{table}

\begin{figure*}[tbp]
    \includegraphics[width=\textwidth]{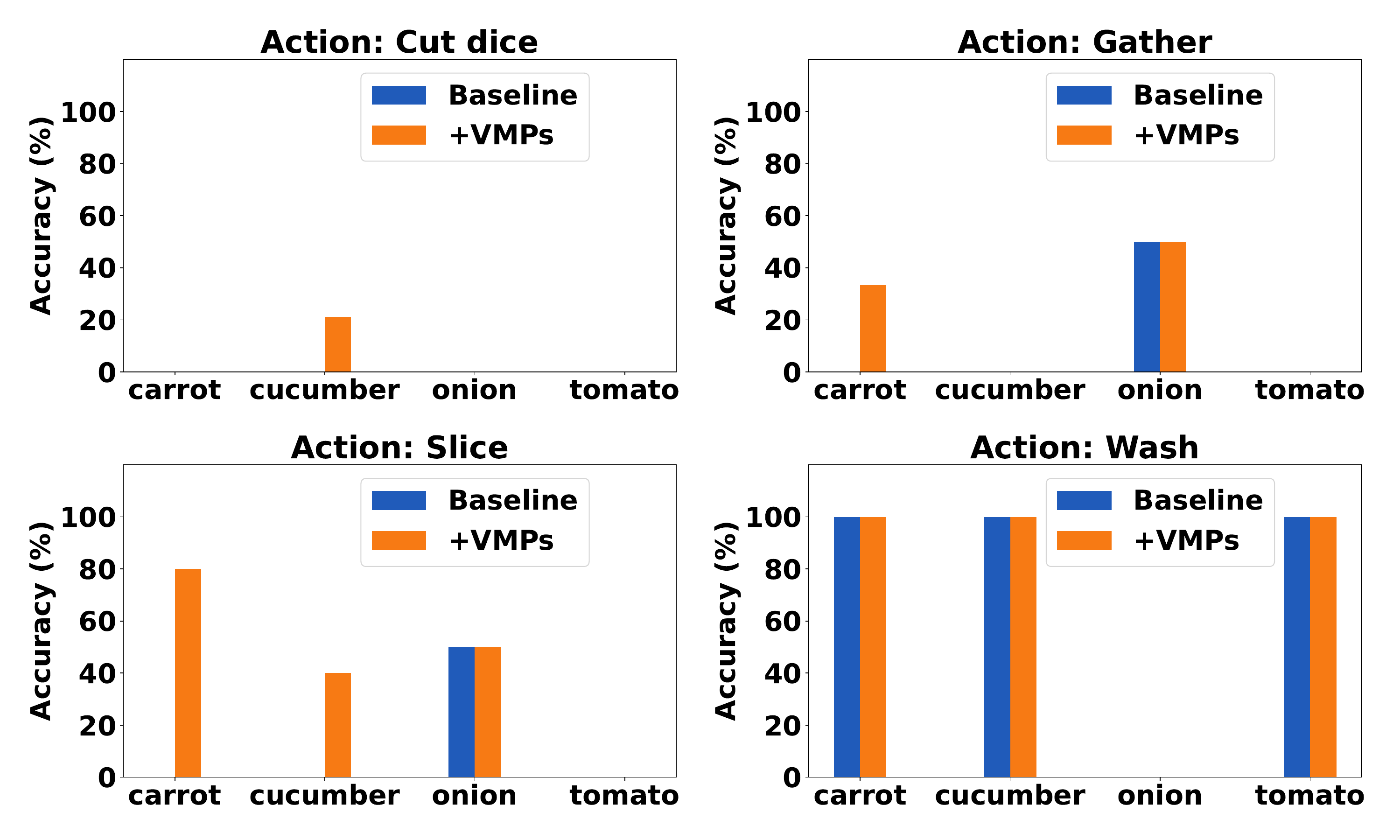}
    \vspace{-0.3cm}
    \caption{Action recognition performance comparison on selected object-centric actions from the MPII Cooking 2 dataset, with and without the use of video motion prompts (VMPs). The figure illustrates the improvement in accuracy for actions such as \textit{cut dice}, \textit{gather}, and \textit{slice}, particularly when combining motion information with object silhouettes and appearances. % The use of VMPs consistently enhances the model’s ability to recognize fine-grained actions involving specific objects.
    }
    \label{fig:obj_action}
\end{figure*}

\begin{table}[tbp]
\centering
\caption{Performance evaluation of action recognition tasks involving specific objects from the MPII Cooking 2 dataset. This table presents the accuracy of the TimeSformer model pretrained on Kinetics-600 and fine-tuned on MPII Cooking 2, comparing results with and without the integration of VMPs. The accuracy in the third column reflects the proportion of correctly classified videos per action class, while the last column summarizes overall performance across the entire test set. Results indicate that the combination of motion and object information enhances action recognition for actions such as \textit{cut dice}, \textit{gather}, and \textit{slice}.}
\resizebox{0.52\linewidth}{!}{
\begin{tabular}{lcccc}
\toprule
Action & Model & Sample acc. & Total acc. \\
\midrule
\multirow{2}{*}{cut dice}  & Baseline & 0.0 & 18.9 \\ 
                           & +VMPs    & \textbf{6.0} & \textbf{27.3} \\ 
\midrule
\multirow{2}{*}{gather}    & Baseline & 11.1 & 29.6 \\ 
                           & +VMPs    & \textbf{33.3} & \textbf{44.0} \\ 
\midrule
\multirow{2}{*}{slice}     & Baseline & 7.7 & 6.9 \\ 
                           & +VMPs    & \textbf{53.9} & \textbf{17.2} \\ 
\midrule
\multirow{2}{*}{wash}      & Baseline & \textbf{100.0} & 71.7 \\ 
                           & +VMPs    & \textbf{100.0} & \textbf{78.7} \\
\bottomrule
\end{tabular}}
\label{tab:obj_action_eval}
\end{table}

\noindent \textbf{VMPs in understanding visual and motion concepts.} Video models rely on both spatial information, such as textures and foreground objects, and temporal information, including motion and evolving dynamics over time.
Motion alone is often insufficient for tasks such as action recognition and is unlikely to perform well independently. Typically, motion occurs alongside object silhouettes and appearances; thus, combining both motion and object information can enhance performance.

To validate the correlation between motion and object, we select a subset of four actions involving specific objects from the MPII Cooking 2 dataset. We use the TimeSformer backbone pretrained on Kinetics-600 and fine-tuned on MPII Cooking 2, evaluating both with and without VMPs. Each subset focuses on a specific object across the selected actions.
Below, we present our experimental results. The accuracy in the third column is calculated as the ratio of the total number of correctly classified videos (across all selected object categories) to the total number of sampled videos per action class. The last column displays the overall performance on the entire test set for these actions.

Fig.~\ref{fig:obj_action} shows the performance on a subset of test samples. We observe that, for the selected objects, using VMPs generally improves performance on actions such as \textit{cut dice}, \textit{gather} and \textit{slice}.

Table~\ref{tab:obj_action_eval} shows that combining both motion and object information leads to improved action recognition performance.

\subsection{Analysis of Regularization, Complexity, and Motion Modulation}

\begin{table}[tbp]
\centering
\setlength{\tabcolsep}{2.0pt}
\caption{Effects of $\lambda$ in the regularization term on (\textit{left}) MPII Cooking 2 (static camera) and (\textit{right}) FineGym (moving camera). We experiment with two backbones (TimeSformer and SlowFast), with the top performance highlighted in bold.}
\label{tab:reg-eval}
\begin{tabular}{cc}
\resizebox{0.488\linewidth}{!}{
\begin{tabular}{lccccc}
\toprule
 & $0$ & $0.1$ & $0.5$ & $1$ & $2$ \\
\midrule
TimeSformer               & \textbf{56.6} & 56.2 & 55.7 & 55.6 & 55.6 \\
SlowFast (slow-only)      & \textbf{55.5} & 53.2 & 53.5 & 52.9 & -     \\
SlowFast (fast-only)      & 53.1          & 53.1 & \textbf{55.2} & 52.8 & -     \\
SlowFast (slow\&fast)     & 54.4          & 54.3 & \textbf{56.8} & 53.6 & -     \\
\bottomrule
\end{tabular}} &
\resizebox{0.5\linewidth}{!}{
\begin{tabular}{lccccc}
\toprule
 & $0$ & $0.5$ & $2.0$ & $2.5$ & $5$ \\
\midrule
TimeSformer           & 81.9 & 83.4          & 84.1          & \textbf{84.4} & 83.3 \\
SlowFast (slow-only)  & 89.3 & 88.2          & 88.3          & 88.6          & \textbf{89.7} \\
SlowFast (fast-only)  & 89.6 & \textbf{90.3} & 90.3          & 90.0          & 90.2 \\
SlowFast (slow\&fast) & 89.7 & 88.5          & \textbf{90.1} & 88.6          & 88.7 \\
\bottomrule
\end{tabular}}
\end{tabular}
\label{tab:lambda-eval}
\end{table}

\noindent\textbf{The role of regularization in static and moving camera scenarios.} We examine the effects of applying a regularization term on models fine-tuned on both static and moving camera datasets in Table~\ref{tab:lambda-eval}. In static camera settings (MPII Cooking 2), smaller regularization values ($\lambda$) generally result in better performance, with the highest accuracy observed when $\lambda=0$. However, for moving camera scenarios (FineGym), larger $\lambda$ values improve performance by smoothing attention maps, addressing temporal discontinuities caused by continuous changes in camera viewpoint while tracking moving subjects.

In the static camera scenario (MPII Cooking 2), the regularization term appears to have a limited impact, with the models achieving optimal performance at smaller $\lambda$ values. Specifically, TimeSformer performs best when no regularization is applied ($\lambda=0$), and for the SlowFast variants, moderate values of $\lambda$ tend to decrease performance.

Conversely, in the moving camera scenario (FineGym), the regularization term becomes more important. Larger $\lambda$ values contribute to smoothing the attention maps, which is crucial in videos where the camera continuously changes its viewpoint. This helps mitigate temporal discontinuity, leading to improved performance. For instance, TimeSformer reaches its peak performance at $\lambda=2.5$, while SlowFast (slow-only) and SlowFast (fast-only) also benefit from higher regularization values, showing the importance of adjusting the regularization term according to the nature of the video data.

% we show the effects of this regularization term on different datasets and backbones. We observe that in static camera scenarios, the regularization term plays a smaller role, as smaller lambda values generally yield better performance. 

% However, in moving camera scenarios, larger lambda values lead to better performance, with the regularization term playing a crucial role in smoothing attention maps. This is because videos captured with a moving camera tend to suffer from temporal discontinuity due to continuous viewpoint changes while tracking the human subject.

\begin{table}[tbp]
\centering
\setlength{\tabcolsep}{5.0pt}
\caption{Computational costs with and without the use of the motion prompt layer (evaluated on HMDB-51 split 1). The experiments are conducted with a batch size of 8, and the reported times (in seconds) reflect the processing time per batch. Each experiment is run 10 times, and we report the mean and standard deviation. The results show that the motion prompt layer adds a negligible computational overhead, as seen in the `Extra' columns.}
\resizebox{\linewidth}{!}{
    \begin{tabular}{ll cccc l cc}
    \toprule
        \multirow{2}{*}{\textbf{Model}} && \multicolumn{4}{c}{\textbf{Training}}  && \multicolumn{2}{c}{\textbf{Testing}} \\
        \cline{3-6}\cline{8-9}
        && \textbf{Forward} & \textbf{Extra}  & \textbf{Backward}  & \textbf{Extra}  && \textbf{Forward} & \textbf{Extra} \\
    \midrule
        SlowFast && 0.028 $\!\!\pm\!\!$ 0.023 &        & 0.269 $\!\!\pm\!\!$ 0.027 &        && 0.005 $\!\!\pm\!\!$ 0.027 & \\
        \multicolumn{2}{l}{\textbf{+VMPs} (slow-only)} & 0.029 $\!\!\pm\!\!$ 0.037 & +0.001 & 0.271 $\!\!\pm\!\!$ 0.017 & +0.002 && 0.005 $\!\!\pm\!\!$ 0.019 & +0.000\\
        \multicolumn{2}{l}{\textbf{+VMPs} (fast-only)} & 0.030 $\!\!\pm\!\!$ 0.022 & +0.002 & 0.290 $\!\!\pm\!\!$ 0.023 & +0.021 && 0.009 $\!\!\pm\!\!$ 0.018 & +0.004\\
        \multicolumn{2}{l}{\textbf{+VMPs} (slow\&fast)}& 0.030 $\!\!\pm\!\!$ 0.023 & +0.002 & 0.293 $\!\!\pm\!\!$ 0.017 & +0.024 && 0.011 $\!\!\pm\!\!$ 0.021 & +0.006\\
        \hline
        C2D && 0.007 $\!\!\pm\!\!$ 0.009 &        & 0.076 $\!\!\pm\!\!$ 0.005 &        && 0.003 $\!\!\pm\!\!$ 0.016\\
        \textbf{+VMPs} && 0.008 $\!\!\pm\!\!$ 0.011 & +0.001 & 0.079 $\!\!\pm\!\!$ 0.007 & +0.003 && 0.004 $\!\!\pm\!\!$ 0.012 & +0.001 \\
        \hline
        I3D && 0.007 $\!\!\pm\!\!$ 0.010 &        & 0.106 $\!\!\pm\!\!$ 0.004 &        && 0.003 $\!\!\pm\!\!$ 0.019 &\\
        \textbf{+VMPs} && 0.008 $\!\!\pm\!\!$ 0.011 & +0.001 & 0.121 $\!\!\pm\!\!$ 0.008 & +0.015 && 0.005 $\!\!\pm\!\!$ 0.013 & +0.002 \\
        \hline
        X3D  && 0.019 $\!\!\pm\!\!$ 0.015 &        & 0.178 $\!\!\pm\!\!$ 0.005 &        && 0.023 $\!\!\pm\!\!$ 0.009 &\\
        \textbf{+VMPs} && 0.020 $\!\!\pm\!\!$ 0.014 & +0.001 & 0.179 $\!\!\pm\!\!$ 0.005 & +0.001 && 0.024 $\!\!\pm\!\!$ 0.009 & +0.001\\
        \hline
        TimeSformer  && 0.141 $\!\!\pm\!\!$ 0.011 &        & 0.253 $\!\!\pm\!\!$ 0.001 &        && 0.010 $\!\!\pm\!\!$ 0.020 &        \\
        \textbf{+VMPs}  && 0.161 $\!\!\pm\!\!$ 0.011 & +0.020 & 0.255 $\!\!\pm\!\!$ 0.001 & +0.002 && 0.011 $\!\!\pm\!\!$ 0.019 & +0.001\\
    \bottomrule
    \end{tabular}
}
\label{tab:cost}
\end{table}

\noindent \textbf{Computational complexity analysis.} The computational cost for generating the frame differencing map is $\mathcal{O}(H\times W)$, where H and W represent the frame height and width, respectively. The learnable PN acts element-wise on the frame differencing map, resulting in a computational complexity of 
$\mathcal{O}(H\times W)$ as well. The element-wise multiplication between the generated attention map and the original frame also incurs a cost of $\mathcal{O}(H\times W)$. Therefore, the total computational cost of the motion prompt layer remains $\mathcal{O}(H\times W)$. Table~\ref{tab:cost} summarizes our results.

\noindent\textbf{Attention maps and motion prompts.} Below we show more visualisations of motion prompts and attention maps on MPII Cooking 2 and HMDB-51.

\begin{figure}[tbp]
\begin{center}
\includegraphics[width=\textwidth]{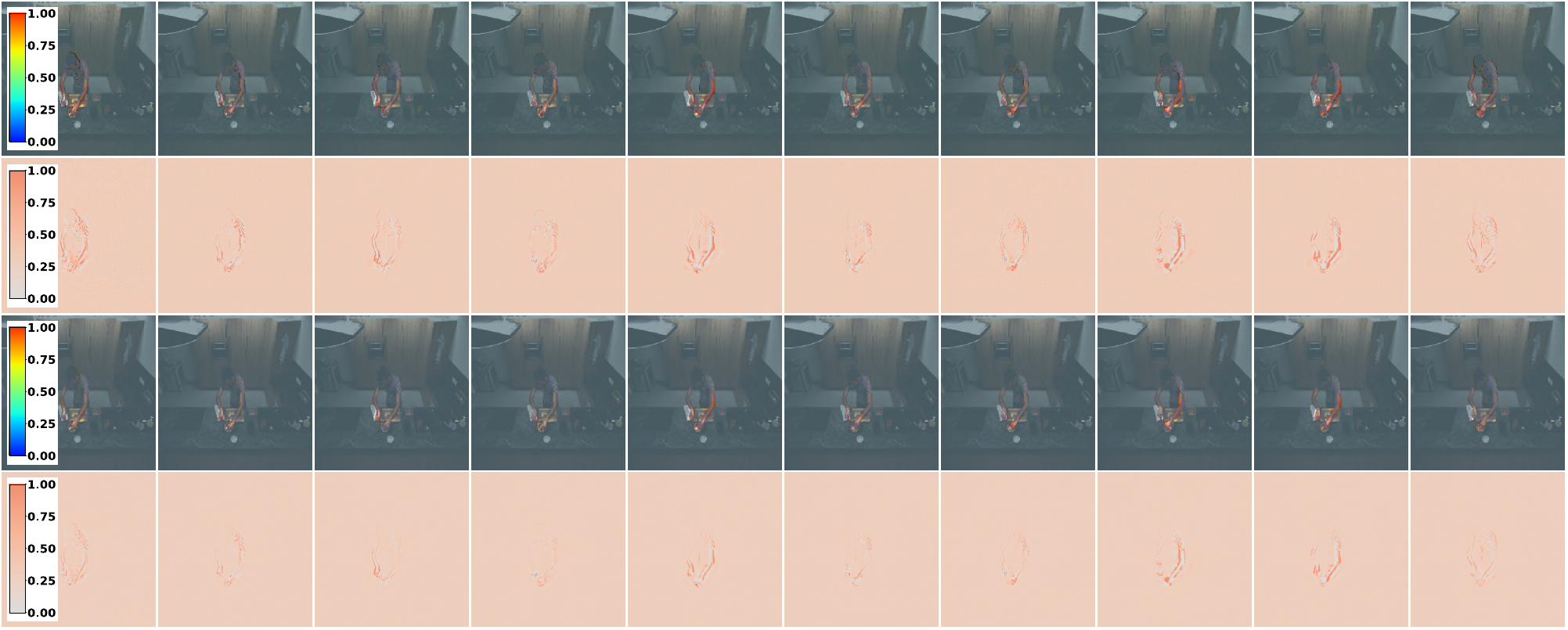}
\caption{Effects of the regularization term. We use the \textit{arrange} action from MPII Cooking 2 for visualization. The first two rows show motion prompts and attention maps without regularization ($\lambda\!=\!0$, learned $a\!=\!23.45$ and $b\!=\!0.03$). The last two rows show results with regularization ($\lambda\!=\!2$, learned $a\!=\!10.90$ and $b\!=\!0.08$). Regularization removes unnecessary motion details, resulting in smoother and cleaner attention maps.
}\label{fig:motion-prompt-attn-mpii1}
\end{center}
\end{figure}

\begin{figure}[tbp]
\begin{center}
\includegraphics[width=\textwidth]{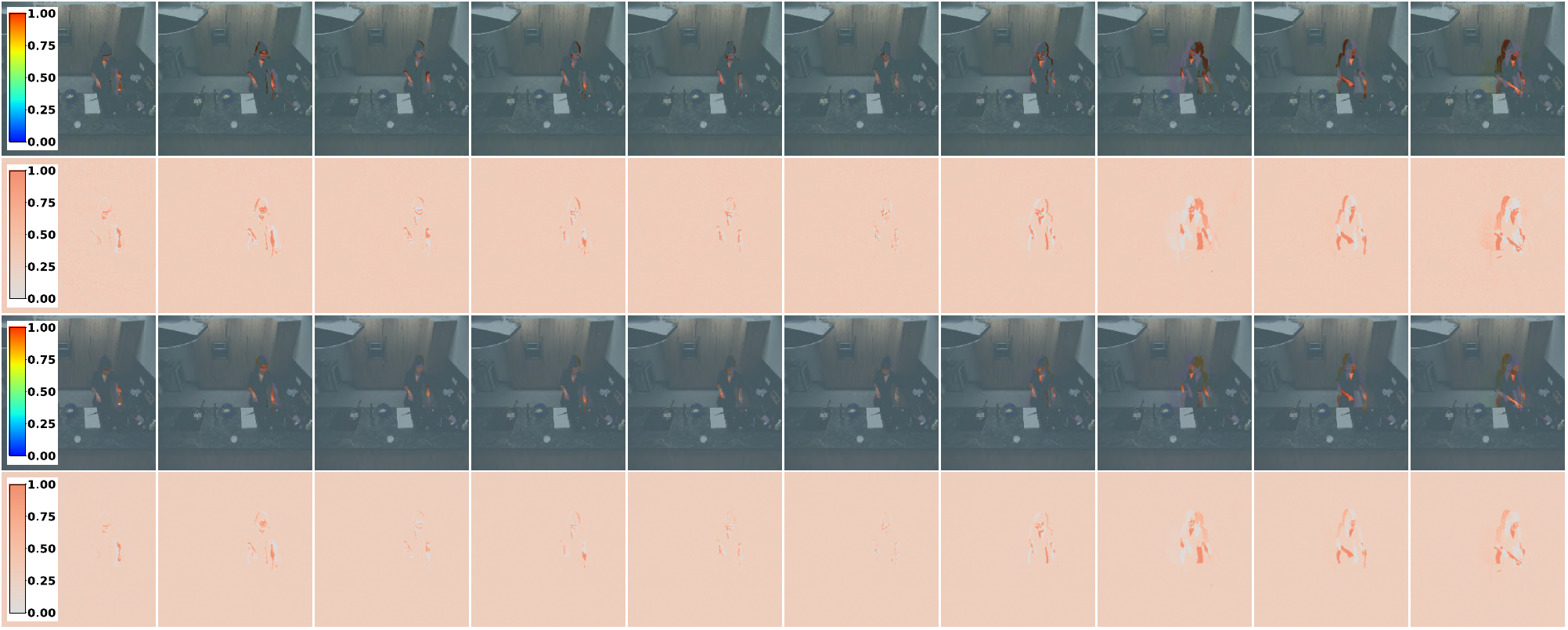}
\caption{Effects of the regularization term. We use the \textit{shake} action from MPII Cooking 2 for visualization. The first two rows show motion prompts and attention maps without regularization.
% ($\lambda\!=\!0$, learned $m\!=\!0.2571$ and $n\!=\!0.0525$\qx{($a\!=\!23.4509$ and $b\!=\!0.0315$)}). 
The last two rows show results with regularization.% ($\lambda\!=\!2$, learned $m\!=\!1.0916$ and $n\!=\!0.1343$\qx{($a\!=\!10.8966$ and $b\!=\!0.0801$)}). % Regularization removes unnecessary motion details, resulting in smoother and cleaner attention maps.
}\label{fig:motion-prompt-attn-mpii2}
\end{center}
\end{figure}

\begin{figure}[tbp]
\begin{center}
\includegraphics[width=\textwidth]{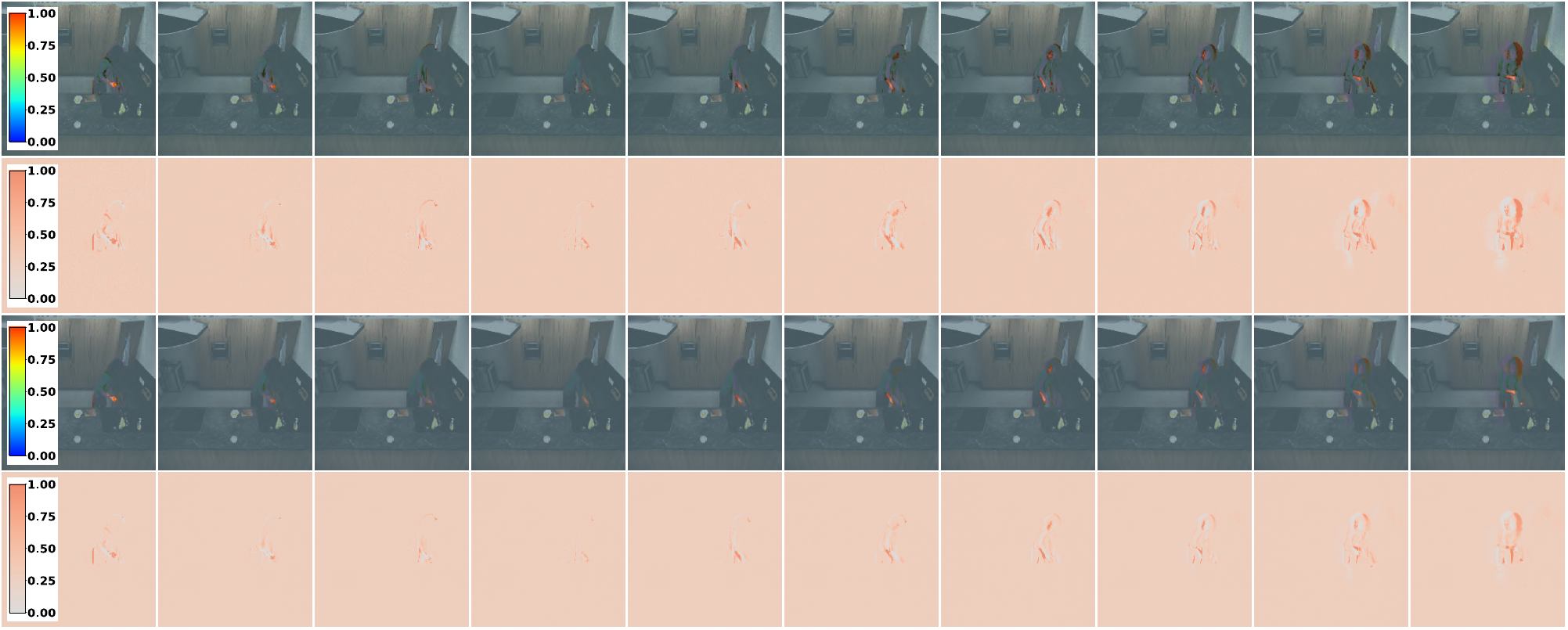}
\caption{Effects of the regularization term. We use the \textit{throw in garbage} action from MPII Cooking 2 for visualization. The first two rows show motion prompts and attention maps without regularization. %  ($\lambda\!=\!0$, learned $m\!=\!0.2571$ and $n\!=\!0.0525$\qx{($a\!=\!23.4509$ and $b\!=\!0.0315$)}). 
The last two rows show results with regularization. % ($\lambda\!=\!2$, learned $m\!=\!1.0916$ and $n\!=\!0.1343$\qx{($a\!=\!10.8966$ and $b\!=\!0.0801$)}). Regularization removes unnecessary motion details, resulting in smoother and cleaner attention maps.
}\label{fig:motion-prompt-attn-mpii3}
\end{center}
\end{figure}

\begin{figure}[tbp]
\begin{center}
\includegraphics[width=\textwidth]{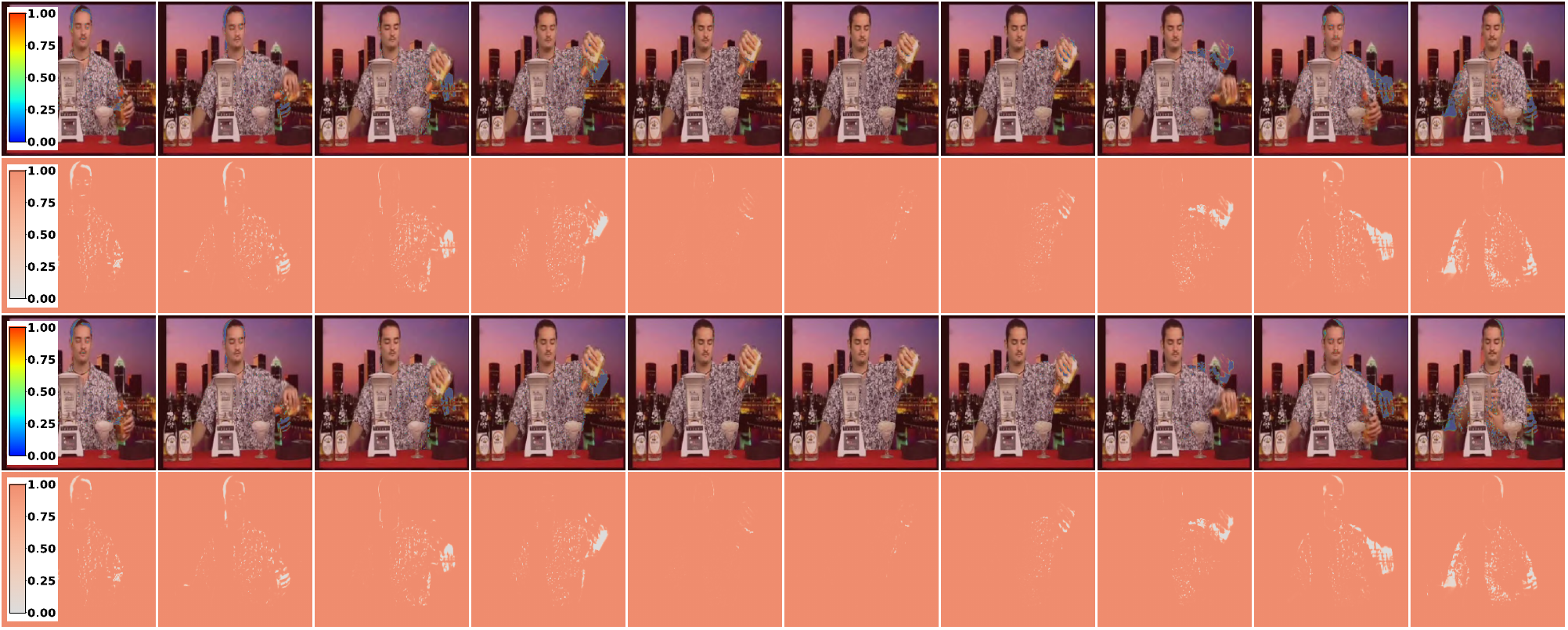}
\caption{Effects of the regularization term. We use the \textit{pour} action from HMDB-51 for visualization. The first two rows show motion prompts and attention maps without regularization ($\lambda\!=\!0$, learned $a\!=\!20.32$ and $b\!=\!-0.36$). The last two rows show results with regularization ($\lambda\!=\!0.001$, learned $a\!=\!21.70$ and $b\!=\!-0.39$). % Regularization removes unnecessary motion details, resulting in smoother and cleaner attention maps.
We observe that in HMDB-51, the attention maps and motion prompts with and without the regularization term are very similar. This behavior is likely because (i) the learned slope and shift parameters for both cases are very close, (ii) the optimal penalty parameter $\lambda$ is small (0 \vs 0.001), and (iii) HMDB-51 is a noisy dataset where camera motions are often more significant than human motions.
}\label{fig:motion-prompt-attn-hmdb1}
\end{center}
\end{figure}

\begin{figure}[tbp]
\begin{center}
\includegraphics[width=\textwidth]{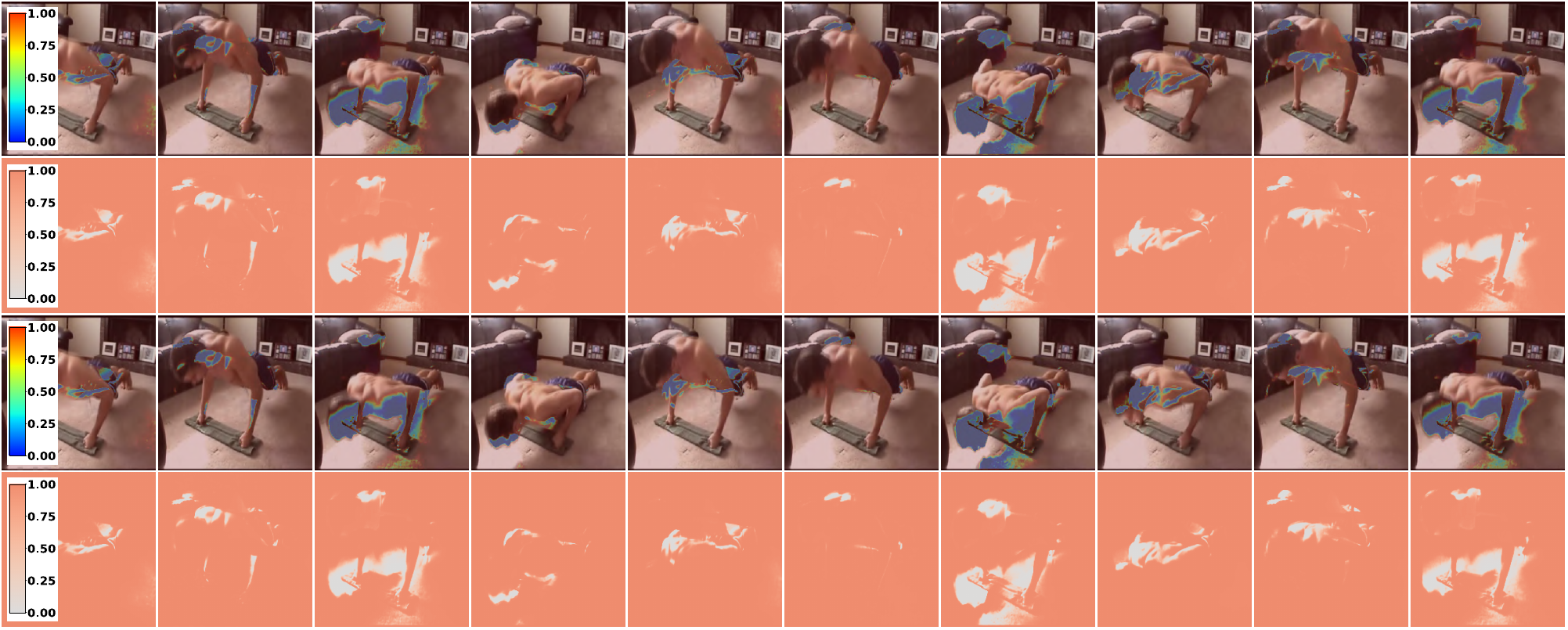}
\caption{Effects of the regularization term. We use the \textit{pushup} action from HMDB-51 for visualization. The first two rows show motion prompts and attention maps without regularization. % ($\lambda\!=\!0$, learned $m\!=\!0.3369$ and $n\!=\!-0.6905$). 
The last two rows show results with regularization.%  ($\lambda\!=\!0.001$, learned $m\!=\!0.2984$ and $n\!=\!-0.7861$). % Regularization removes unnecessary motion details, resulting in smoother and cleaner attention maps. 
}\label{fig:motion-prompt-attn-hmdb2}
\end{center}
\end{figure}

We notice that in MPII Cooking 2, the background is lighter orange, indicating lower attention scores (Fig.~\ref{fig:motion-prompt-attn-mpii1},~\ref{fig:motion-prompt-attn-mpii2} and~\ref{fig:motion-prompt-attn-mpii3}). In contrast, in HMDB-51 (Fig.~\ref{fig:motion-prompt-attn-hmdb1} and~\ref{fig:motion-prompt-attn-hmdb2}), the background is much darker red, indicating higher attention scores due to the videos being captured by moving cameras.

Additionally, we observe that in HMDB-51, the attention maps and motion prompts with and without the regularization term are very similar (Fig.~\ref{fig:motion-prompt-attn-hmdb1} and~\ref{fig:motion-prompt-attn-hmdb2}). This behavior is likely because (i) the learned slope and shift parameters for both cases are very close, (ii) the optimal regularization penalty parameter $\lambda$ is small (0 \vs 0.001), and (iii) HMDB-51 is a noisy dataset where camera motions are often more significant than human motions.
\begin{tcolorbox}[ width=1.0\linewidth, colframe=blackish, colback=beaublue, boxsep=0mm, arc=3mm, left=1mm, right=1mm, top=1mm, bottom=1mm]
We also observe that the attention maps show several interesting patterns: they (i) highlight the motion regions in the current frame, (ii) capture potential movements from the previous frame, and (iii) attend to background scenes affected by camera motions. These observations indicate that our attention maps, guided by only two learnable parameters, effectively highlight visual contents of interest while capturing dynamics over short periods of time.

In static camera scenes, such as MPII Cooking 2, the attention mechanism highlights motion regions, particularly the hand regions, which are central to cooking activities. 
In contrast, for moving camera scenes where the viewpoint continuously changes to track subjects, as in HMDB-51 and FineGym, the attention tends to highlight the background. Interestingly, we observe that our motion prompt layer also focuses on the silhouettes of human subjects and the boundaries of objects in these dynamic camera scenes.
\end{tcolorbox}

\begin{figure}[tbp]
\begin{center}
\includegraphics[width=\textwidth]{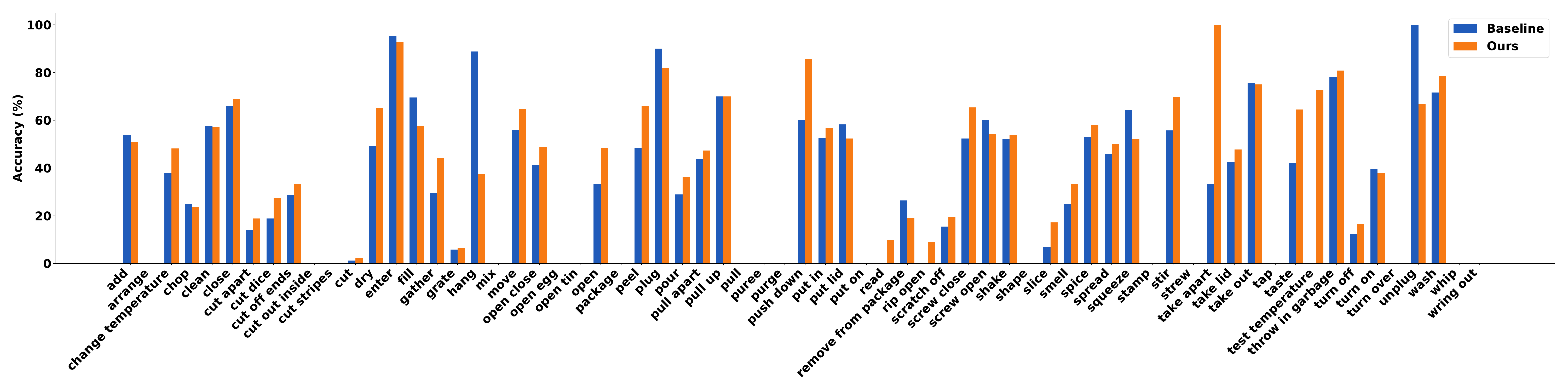}
\caption{Per-class accuracy comparison is conducted between the baseline model (pretrained on Kinetics-600 and then finetuned on MPII Cooking 2, without VMPs) and our VMP-enhanced model on MPII Cooking 2, using TimeSformer as the backbone. The integration of our VMPs results in improvements across 34 fine-grained actions out of a total of 67. Notably, with our VMPs, the model achieves accuracies for actions previously classified with 0 accuracy in the baseline model, such as \textit{read}, \textit{rip open}, and \textit{test temperature}.}\label{fig:mpii-acc}
\end{center}
\end{figure}

\begin{figure*}[tbp]
    \centering
    \subfigure[Action \textit{run} from HMDB-51.]{\includegraphics[trim=0.05cm 0.1cm 0.05cm 0.0cm, clip=true,width=0.8\textwidth]{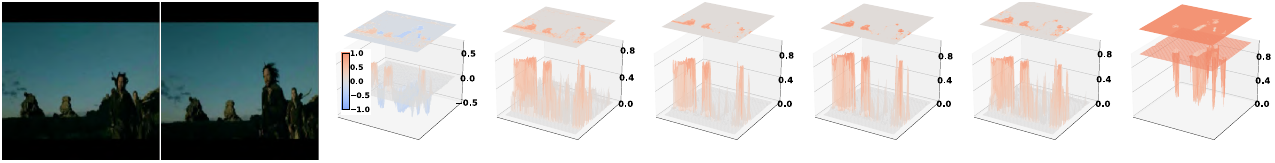}} 
    \subfigure[Action \textit{situp} from HMDB-51.]{\includegraphics[trim=0.05cm 0.1cm 0.05cm 0.0cm, clip=true,width=0.8\textwidth]{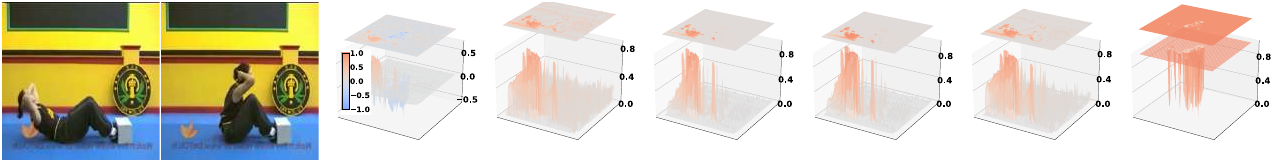}} 
    \subfigure[Action \textit{push down} from MPII Cooking 2.]{\includegraphics[trim=0.05cm 0.1cm 0.05cm 0.0cm, clip=true,width=0.8\textwidth]{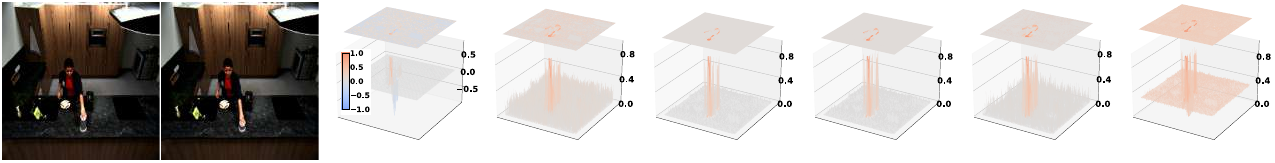}}
    \subfigure[Action \textit{whip} from MPII Cooking 2.]{\includegraphics[trim=0.05cm 0cm 0.05cm 0.0cm, clip=true,width=0.8\textwidth]{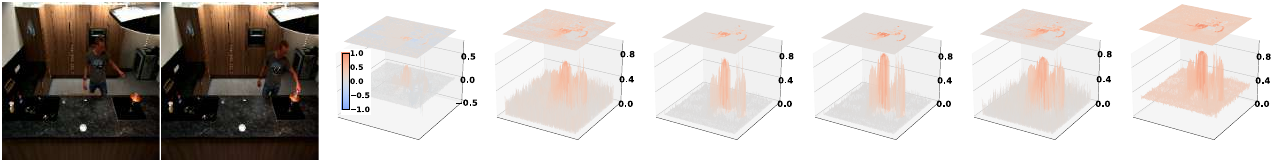}}
    \subfigure[Action (\textit{Balance Beam}) \textit{leap forward with leg change} from FineGym.]{\includegraphics[trim=0.05cm 0cm 0.05cm 0.0cm, clip=true,width=0.8\textwidth]{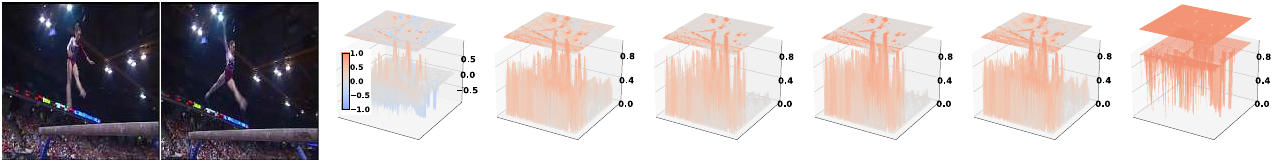}}
    \subfigure[Action (\textit{Uneven Bar}) \textit{giant circle backward} from FineGym.]{\includegraphics[trim=0.05cm 0cm 0.05cm 0.0cm, clip=true,width=0.8\textwidth]{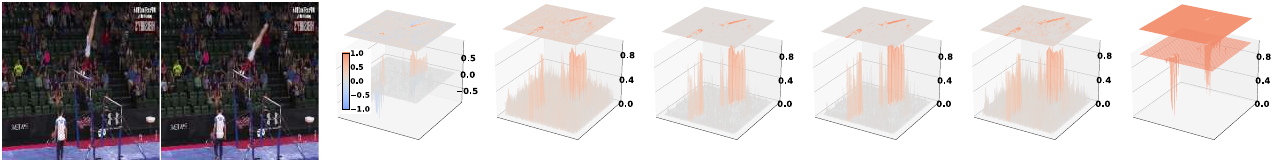}}
    \subfigure[Anomaly \textit{explosion} from UCF-Crime.]{\includegraphics[trim=0.05cm 0cm 0.05cm 0.0cm, clip=true,width=0.8\textwidth]{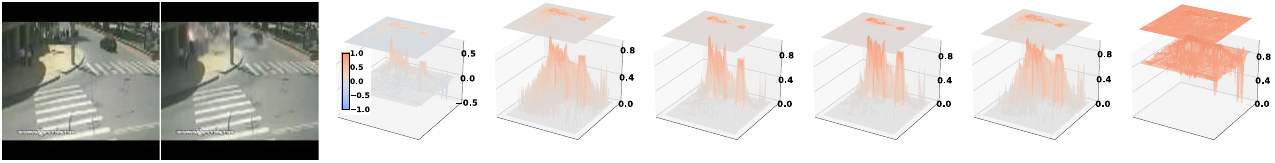}}
    \subfigure[Anomaly \textit{fighting} from UCF-Crime.]{\includegraphics[trim=0.05cm 0cm 0.05cm 0.0cm, clip=true,width=0.8\textwidth]{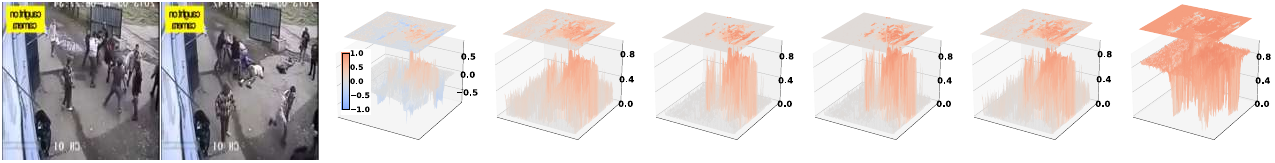}}
    \vspace{-0.3cm}
    \caption{
    We compare existing PN with our PN on motion modulation. The first two columns show consecutive video frames, and the third column displays 3D surface plots of frame differencing maps. Columns 4-7 show attention outputs in both attention maps and 3D surface plots for Gamma, MaxExp, SigmE, and AsinhE. The final column shows our PN's outputs, which capture various motions across different video types. For UCF-Crime, we apply the slope and shift learned from MPII Cooking 2, as both are captured by static cameras.
    % We compare existing PN with our PN on motion modulation. The first two columns show consecutive video frames, and the third column displays 3D surface plots of the corresponding frame differencing maps. The 4th to 7th columns show output attentions in both attention maps and 3D surface plots for Gamma, MaxExp, SigmE, and AsinhE. The last column shows outputs from our PN function, which focuses on different motions across video types, such as human actions, fine-grained actions, static and moving cameras, and anomaly detection. For UCF-Crime, we use the learned slope and shift from MPII Cooking 2, as both are captured by static cameras.
    }
    \label{fig:pns-3d-vis}
    % \vspace{-0.6cm}
\end{figure*}

\begin{figure*}[tbp]
    \centering
    \subfigure[Action \textit{run} from HMDB-51.]{\includegraphics[width=\textwidth]{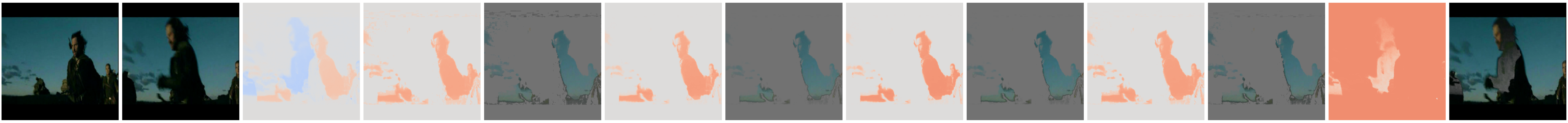}} 
    \subfigure[Action \textit{situp} from HMDB-51.]{\includegraphics[width=\textwidth]{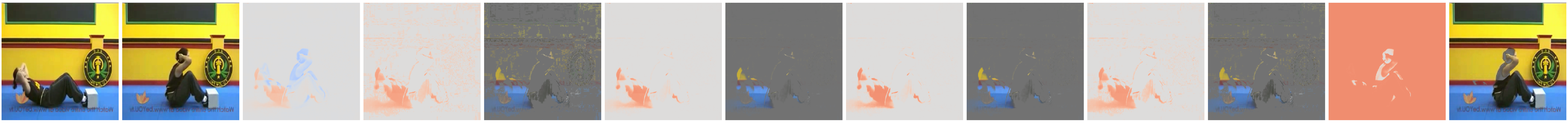}} 
    \subfigure[Action \textit{push down} from MPII Cooking 2.]{\includegraphics[width=\textwidth]{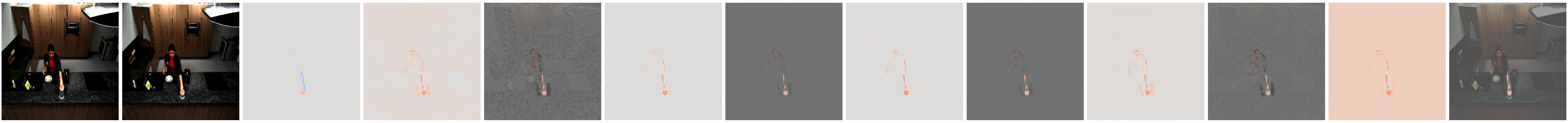}}
    \subfigure[Action \textit{whip} from MPII Cooking 2.]{\includegraphics[width=\textwidth]{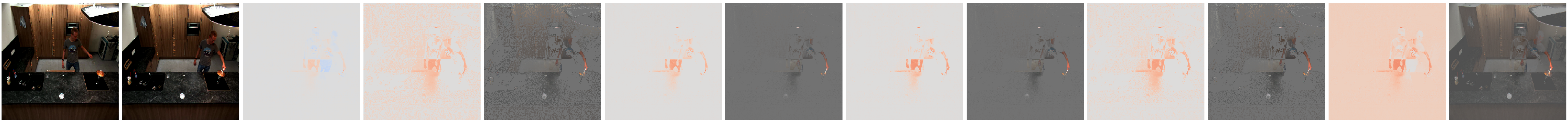}}
    \subfigure[Action \textit{(Balance Beam) leap forward with leg change} from FineGym.]{\includegraphics[width=\textwidth]{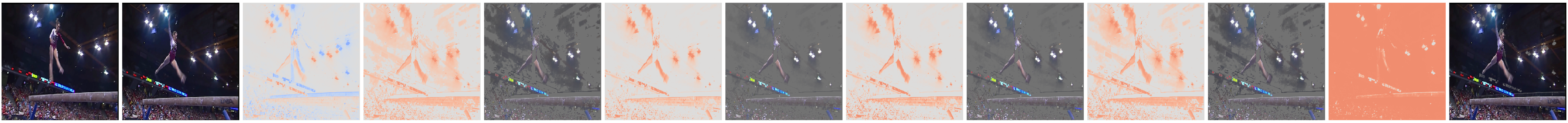}}
    \subfigure[Action \textit{(Uneven Bar) giant circle backward} from FineGym.]{\includegraphics[width=\textwidth]{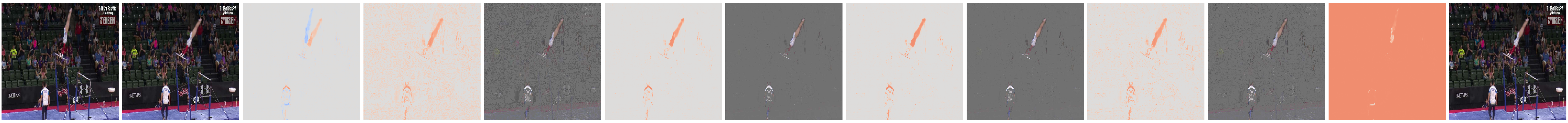}}
    \subfigure[Anomaly \textit{explosion} from UCF-Crime.]{\includegraphics[width=\textwidth]{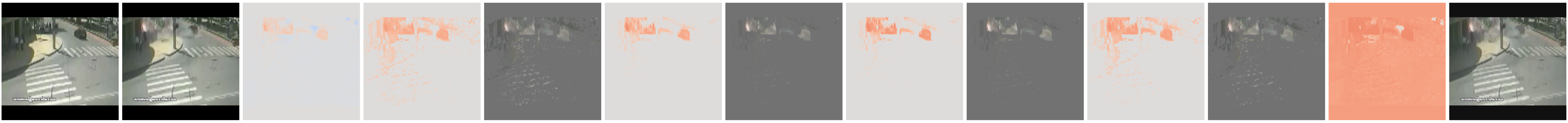}}
    \subfigure[Anomaly \textit{fighting} from UCF-Crime.]{\includegraphics[width=\textwidth]{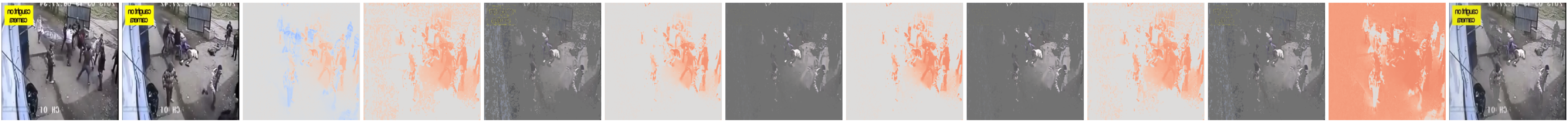}}
    \vspace{-0.3cm}
    \caption{Visualizations include original consecutive frames (first two columns), frame differencing maps (third column), pairs of attention maps and motion prompts for Gamma, MaxExp, SigmE, and AsinhE (fourth to eleventh columns). The last two columns display our attention maps and motion prompts. Our attention maps (i) depict clear motion regions, (ii) highlight motions of interest and/or contextual environments relevant to the motions, and our motion prompts capture rich motion patterns. Existing PN functions only focus on motions, often capture noisy patterns and without emphasizing contexts.}
    \label{fig:pns-3d-vis-2}
    % \vspace{-0.6cm}
\end{figure*}

\noindent\textbf{Per-class accuracy on MPII Cooking 2.} Fig.~\ref{fig:mpii-acc} shows the per-class accuracy comparison between the baseline model (pre-trained on Kinetics-600 and then finetuned on MPII Cooking 2 without VMPs) and our VMP-enhanced model on MPII Cooking 2. 
We use TimeSformer as the backbone. 
As shown in the figure, integrating our VMPs results in improvements in the accuracy of 34 fine-grained actions (out of a total of 67 actions). 
The model enhanced with VMPs is able to classify actions that are previously challenging for the baseline model (\eg, with 0 accuracy), such as \textit{read}, \textit{rip open}, and \textit{test temperature}. Furthermore, The model finetuned with VMPs also outperforms the baseline model on actions like \textit{open}, \textit{peel}, \textit{push down}, \textit{slice}, \textit{stir}, \textit{take apart}, and \textit{taste} by a large margin.

\noindent\textbf{Comparison of PN functions on motion modulation.} Our qualitative results in Fig.~\ref{fig:pns-3d-vis}. We notice that for static cameras, such as in MPII Cooking 2, our PN function focuses on motion regions, and captures more fine-grained motion patterns compared to other PN functions, which tend to capture more noisy motions.

For moving cameras, such as those in HMDB-51 and FineGym, our PN function captures the background context while the bright regions highlight the motions of interest. 
This is reasonable as most motions are relevant to surrounding objects and contexts. 
We apply our learned slope and shift parameters from MPII Cooking 2 to the UCF-Crime dataset, as both are captured by static cameras. 
Interestingly, our PN function is able to highlight the contextual environment while emphasizing the anomaly motion regions. This demonstrates that our motion prompt layer, equipped with our new PN function, is motion-dependent, attention-driven and generalizable to different video types, including anomaly detection videos. 

Fig.~\ref{fig:pns-3d-vis-2} compares visualizations of attention maps and motion prompts generated by various PN functions, including ours. As shown in the figure, existing PN functions typically focus on motions, often capturing noisy patterns and frequently disregarding the contextual environment in which the motion occurs. In contrast, our PN function captures both motions of interest and relevant contexts. For instance, anomalies are largely contextual, and our PN function effectively captures both the motions and their surrounding environments.

\end{document}

%% file: definitions.tex
\usepackage{amsmath,amsfonts,bm}

\newcommand{\idx}[1]{\mathcal{I}_{#1}}

\newcommand{\mbr}[1]{\mathbb{R}^{#1}}

\def\eqref#1{(\ref{#1})}
\def\1{\bm{1}}

\def\vs{{\bm{s}}}

\def\vx{{\bm{x}}}

\def\mA{{\bm{A}}}

\def\mD{{\bm{D}}}

\def\mF{{\bm{F}}}

\def\mK{{\bm{K}}}

\def\mQ{{\bm{Q}}}

\def\mV{{\bm{V}}}
\def\mW{{\bm{W}}}
\def\mX{{\bm{X}}}

\def\mZ{{\bm{Z}}}

\DeclareMathAlphabet{\mathsfit}{\encodingdefault}{\sfdefault}{m}{sl}
\SetMathAlphabet{\mathsfit}{bold}{\encodingdefault}{\sfdefault}{bx}{n}
\newcommand{\tens}[1]{\bm{\mathsfit{#1}}}
\def\tA{{\tens{A}}}

\def\tD{{\tens{D}}}

\def\tF{{\tens{F}}}

\def\tX{{\tens{X}}}

\def\tZ{{\tens{Z}}}

%% file: qixiang.bbl
\begin{thebibliography}{61}
\providecommand{\natexlab}[1]{#1}
\providecommand{\url}[1]{\texttt{#1}}
\expandafter\ifx\csname urlstyle\endcsname\relax
  \providecommand{\doi}[1]{doi: #1}\else
  \providecommand{\doi}{doi: \begingroup \urlstyle{rm}\Url}\fi

\bibitem[Arnab et~al.(2021)Arnab, Dehghani, Heigold, Sun, Lu{\v{c}}i{\'c}, and
  Schmid]{arnab2021vivit}
Anurag Arnab, Mostafa Dehghani, Georg Heigold, Chen Sun, Mario Lu{\v{c}}i{\'c},
  and Cordelia Schmid.
\newblock Vivit: A video vision transformer.
\newblock In \emph{ICCV}, pages 6836--6846, 2021.

\bibitem[Ba et~al.(2016)Ba, Kiros, and Hinton]{ba2016layer}
Jimmy~Lei Ba, Jamie~Ryan Kiros, and Geoffrey~E Hinton.
\newblock Layer normalization.
\newblock \emph{arXiv preprint arXiv:1607.06450}, 2016.

\bibitem[Bertasius et~al.(2021)Bertasius, Wang, and
  Torresani]{gberta_2021_ICML}
Gedas Bertasius, Heng Wang, and Lorenzo Torresani.
\newblock Is space-time attention all you need for video understanding?
\newblock In \emph{ICML}, July 2021.

\bibitem[Bilen et~al.(2016)Bilen, Fernando, Gavves, Vedaldi, and
  Gould]{Bilen_2016_CVPR}
Hakan Bilen, Basura Fernando, Efstratios Gavves, Andrea Vedaldi, and Stephen
  Gould.
\newblock Dynamic image networks for action recognition.
\newblock In \emph{CVPR}, June 2016.

\bibitem[Brauwers and Frasincar(2021)]{brauwers2021general}
Gianni Brauwers and Flavius Frasincar.
\newblock A general survey on attention mechanisms in deep learning.
\newblock \emph{IEEE Transactions on Knowledge and Data Engineering},
  35\penalty0 (4):\penalty0 3279--3298, 2021.

\bibitem[Brown et~al.(2020)Brown, Mann, Ryder, Subbiah, Kaplan, Dhariwal,
  Neelakantan, Shyam, Sastry, Askell, et~al.]{brown2020language}
Tom Brown, Benjamin Mann, Nick Ryder, Melanie Subbiah, Jared~D Kaplan, Prafulla
  Dhariwal, Arvind Neelakantan, Pranav Shyam, Girish Sastry, Amanda Askell,
  et~al.
\newblock Language models are few-shot learners.
\newblock \emph{NeurIPS}, 33:\penalty0 1877--1901, 2020.

\bibitem[Carreira and Zisserman(2018)]{i3d_net}
Jo{\~{a}}o Carreira and Andrew Zisserman.
\newblock {Quo Vadis, Action Recognition? A New Model and the Kinetics
  Dataset}.
\newblock \emph{CVPR}, pages 1--10, 2018.

\bibitem[Chen et~al.(2023)Chen, Zhang, Langren{\'e}, and
  Zhu]{chen2023unleashing}
Banghao Chen, Zhaofeng Zhang, Nicolas Langren{\'e}, and Shengxin Zhu.
\newblock Unleashing the potential of prompt engineering in large language
  models: a comprehensive review.
\newblock \emph{arXiv preprint arXiv:2310.14735}, 2023.

\bibitem[Chen et~al.(2021)Chen, Fan, and
  Panda]{DBLP:journals/corr/abs-2103-14899}
Chun{-}Fu Chen, Quanfu Fan, and Rameswar Panda.
\newblock Crossvit: Cross-attention multi-scale vision transformer for image
  classification.
\newblock \emph{CoRR}, abs/2103.14899, 2021.

\bibitem[Dalal et~al.(2006)Dalal, Triggs, and Schmid]{hof}
Navneet Dalal, Bill Triggs, and Cordelia Schmid.
\newblock {Human Detection Using Oriented Histogram of Flow and Appearance}.
\newblock \emph{ECCV}, pages 428--441, 2006.

\bibitem[Ding et~al.(2022)Ding, Qin, Yang, Wei, Yang, Su, Hu, Chen, Chan, Chen,
  et~al.]{ding2022delta}
Ning Ding, Yujia Qin, Guang Yang, Fuchao Wei, Zonghan Yang, Yusheng Su,
  Shengding Hu, Yulin Chen, Chi-Min Chan, Weize Chen, et~al.
\newblock Delta tuning: A comprehensive study of parameter efficient methods
  for pre-trained language models.
\newblock \emph{arXiv preprint arXiv:2203.06904}, 2022.

\bibitem[Dosovitskiy et~al.(2021)Dosovitskiy, Beyer, Kolesnikov, Weissenborn,
  Zhai, Unterthiner, Dehghani, Minderer, Heigold, Gelly, Uszkoreit, and
  Houlsby]{dosovitskiy2021an}
Alexey Dosovitskiy, Lucas Beyer, Alexander Kolesnikov, Dirk Weissenborn,
  Xiaohua Zhai, Thomas Unterthiner, Mostafa Dehghani, Matthias Minderer, Georg
  Heigold, Sylvain Gelly, Jakob Uszkoreit, and Neil Houlsby.
\newblock An image is worth 16x16 words: Transformers for image recognition at
  scale.
\newblock In \emph{ICLR}, 2021.

\bibitem[Dubey et~al.(2022)Dubey, Singh, and Chaudhuri]{dubey2022activation}
Shiv~Ram Dubey, Satish~Kumar Singh, and Bidyut~Baran Chaudhuri.
\newblock Activation functions in deep learning: A comprehensive survey and
  benchmark.
\newblock \emph{Neurocomputing}, 503:\penalty0 92--108, 2022.

\bibitem[Feichtenhofer(2020)]{feichtenhofer2020x3d}
Christoph Feichtenhofer.
\newblock X3d: Expanding architectures for efficient video recognition.
\newblock In \emph{Proceedings of the IEEE/CVF conference on computer vision
  and pattern recognition}, pages 203--213, 2020.

\bibitem[Feichtenhofer et~al.(2016)Feichtenhofer, Pinz, and
  Wildes]{spat_temp_resnet}
Christoph Feichtenhofer, Axel Pinz, and Richard~P. Wildes.
\newblock Spatiotemporal residual networks for video action recognition.
\newblock In \emph{NeurIPS}, pages 3468--3476, 2016.

\bibitem[Feichtenhofer et~al.(2019)Feichtenhofer, Fan, Malik, and
  He]{feichtenhofer2019slowfast}
Christoph Feichtenhofer, Haoqi Fan, Jitendra Malik, and Kaiming He.
\newblock Slowfast networks for video recognition.
\newblock In \emph{ICCV}, pages 6202--6211, 2019.

\bibitem[Guo et~al.(2022)Guo, Xu, Liu, Liu, Jiang, Mu, Zhang, Martin, Cheng,
  and Hu]{guo2022attention}
Meng-Hao Guo, Tian-Xing Xu, Jiang-Jiang Liu, Zheng-Ning Liu, Peng-Tao Jiang,
  Tai-Jiang Mu, Song-Hai Zhang, Ralph~R Martin, Ming-Ming Cheng, and Shi-Min
  Hu.
\newblock Attention mechanisms in computer vision: A survey.
\newblock \emph{CVM}, 8\penalty0 (3):\penalty0 331--368, 2022.

\bibitem[Hashiguchi and
  Tamaki(2022)]{https://doi.org/10.48550/arxiv.2204.00452}
Ryota Hashiguchi and Toru Tamaki.
\newblock Vision transformer with cross-attention by temporal shift for
  efficient action recognition, 2022.

\bibitem[He et~al.(2016)He, Zhang, Ren, and Sun]{he2016deep}
Kaiming He, Xiangyu Zhang, Shaoqing Ren, and Jian Sun.
\newblock Deep residual learning for image recognition.
\newblock In \emph{CVPR}, pages 770--778, 2016.

\bibitem[Hrinchuk et~al.(2019)Hrinchuk, Khrulkov, Mirvakhabova, Orlova, and
  Oseledets]{hrinchuk2019tensorized}
Oleksii Hrinchuk, Valentin Khrulkov, Leyla Mirvakhabova, Elena Orlova, and Ivan
  Oseledets.
\newblock Tensorized embedding layers for efficient model compression.
\newblock \emph{arXiv preprint arXiv:1901.10787}, 2019.

\bibitem[Hu et~al.(2022)Hu, Luo, and Chen]{hu2022mage}
Yaosi Hu, Chong Luo, and Zhenzhong Chen.
\newblock Make it move: Controllable image-to-video generation with text
  descriptions.
\newblock In \emph{CVPR}, 2022.

\bibitem[Ioffe and Szegedy(2015)]{ioffe2015batch}
Sergey Ioffe and Christian Szegedy.
\newblock Batch normalization: Accelerating deep network training by reducing
  internal covariate shift.
\newblock In \emph{ICML}, pages 448--456. pmlr, 2015.

\bibitem[J{\'e}gou et~al.(2009)J{\'e}gou, Douze, and
  Schmid]{jegou2009burstiness}
Herv{\'e} J{\'e}gou, Matthijs Douze, and Cordelia Schmid.
\newblock On the burstiness of visual elements.
\newblock In \emph{CVPR}, pages 1169--1176. IEEE, 2009.

\bibitem[Jia et~al.(2022)Jia, Tang, Chen, Cardie, Belongie, Hariharan, and
  Lim]{jia2022visual}
Menglin Jia, Luming Tang, Bor-Chun Chen, Claire Cardie, Serge Belongie, Bharath
  Hariharan, and Ser-Nam Lim.
\newblock Visual prompt tuning.
\newblock In \emph{ECCV}, pages 709--727. Springer, 2022.

\bibitem[Ju et~al.(2022)Ju, Han, Zheng, Zhang, and Xie]{ju2022prompting}
Chen Ju, Tengda Han, Kunhao Zheng, Ya~Zhang, and Weidi Xie.
\newblock Prompting visual-language models for efficient video understanding.
\newblock In \emph{ECCV}, pages 105--124. Springer, 2022.

\bibitem[Kay et~al.(2017)Kay, Carreira, Simonyan, Zhang, Hillier,
  Vijayanarasimhan, Viola, Green, Back, Natsev, et~al.]{kay2017kinetics}
Will Kay, Joao Carreira, Karen Simonyan, Brian Zhang, Chloe Hillier, Sudheendra
  Vijayanarasimhan, Fabio Viola, Tim Green, Trevor Back, Paul Natsev, et~al.
\newblock The kinetics human action video dataset.
\newblock \emph{arXiv preprint arXiv:1705.06950}, 2017.

\bibitem[Kim et~al.(2022)Kim, Gowda, Aodha, and Sevilla-Lara]{kim2022capturing}
Kiyoon Kim, Shreyank~N Gowda, Oisin~Mac Aodha, and Laura Sevilla-Lara.
\newblock Capturing temporal information in a single frame: Channel sampling
  strategies for action recognition.
\newblock In \emph{BMVC}. {BMVA} Press, 2022.

\bibitem[Kim et~al.(2021)Kim, Son, and Kim]{kim2021vilt}
Wonjae Kim, Bokyung Son, and Ildoo Kim.
\newblock Vilt: Vision-and-language transformer without convolution or region
  supervision.
\newblock In \emph{ICML}, pages 5583--5594. PMLR, 2021.

\bibitem[Kl{\"a}ser et~al.(2008)Kl{\"a}ser, Marszalek, and Schmid]{3D-HOG}
Alexander Kl{\"a}ser, Marcin Marszalek, and Cordelia Schmid.
\newblock {A Spatio-Temporal Descriptor Based on 3D-Gradients}.
\newblock \emph{BMCV}, pages 1--10, 2008.

\bibitem[Koniusz and Zhang(2021)]{koniusz2021power}
Piotr Koniusz and Hongguang Zhang.
\newblock Power normalizations in fine-grained image, few-shot image and graph
  classification.
\newblock \emph{IEEE TPAMI}, 44\penalty0 (2):\penalty0 591--609, 2021.

\bibitem[Kuehne et~al.(2011)Kuehne, Jhuang, Garrote, Poggio, and
  Serre]{kuehne2011hmdb}
Hildegard Kuehne, Hueihan Jhuang, Est{\'\i}baliz Garrote, Tomaso Poggio, and
  Thomas Serre.
\newblock Hmdb: a large video database for human motion recognition.
\newblock In \emph{ICCV}, pages 2556--2563. IEEE, 2011.

\bibitem[Lee et~al.(2020)Lee, Cho, and Kang]{Lee2020Mixout:}
Cheolhyoung Lee, Kyunghyun Cho, and Wanmo Kang.
\newblock Mixout: Effective regularization to finetune large-scale pretrained
  language models.
\newblock In \emph{ICLR}, 2020.

\bibitem[Lin et~al.(2021)Lin, Cheng, Wu, Yang, Shen, Wang, Song, and
  Yuan]{DBLP:journals/corr/abs-2106-05786}
Hezheng Lin, Xing Cheng, Xiangyu Wu, Fan Yang, Dong Shen, Zhongyuan Wang, Qing
  Song, and Wei Yuan.
\newblock {CAT:} cross attention in vision transformer.
\newblock \emph{CoRR}, abs/2106.05786, 2021.

\bibitem[Long et~al.(2020)Long, Yao, Qiu, Tian, Luo, and Mei]{long2020learning}
Fuchen Long, Ting Yao, Zhaofan Qiu, Xinmei Tian, Jiebo Luo, and Tao Mei.
\newblock Learning to localize actions from moments.
\newblock In \emph{Computer Vision--ECCV 2020: 16th European Conference,
  Glasgow, UK, August 23--28, 2020, Proceedings, Part III 16}, pages 137--154.
  Springer, 2020.

\bibitem[Niu et~al.(2021)Niu, Zhong, and Yu]{NIU202148}
Zhaoyang Niu, Guoqiang Zhong, and Hui Yu.
\newblock A review on the attention mechanism of deep learning.
\newblock \emph{Neurocomputing}, 452:\penalty0 48--62, 2021.
\newblock ISSN 0925-2312.
\newblock \doi{https://doi.org/10.1016/j.neucom.2021.03.091}.

\bibitem[Nwankpa et~al.(2018)Nwankpa, Ijomah, Gachagan, and
  Marshall]{nwankpa2018activation}
Chigozie Nwankpa, Winifred Ijomah, Anthony Gachagan, and Stephen Marshall.
\newblock Activation functions: Comparison of trends in practice and research
  for deep learning.
\newblock \emph{arXiv preprint arXiv:1811.03378}, 2018.

\bibitem[O'shea and Nash(2015)]{o2015introduction}
Keiron O'shea and Ryan Nash.
\newblock An introduction to convolutional neural networks.
\newblock \emph{arXiv preprint arXiv:1511.08458}, 2015.

\bibitem[Radford et~al.(2021)Radford, Kim, Hallacy, Ramesh, Goh, Agarwal,
  Sastry, Askell, Mishkin, Clark, et~al.]{radford2021learning}
Alec Radford, Jong~Wook Kim, Chris Hallacy, Aditya Ramesh, Gabriel Goh,
  Sandhini Agarwal, Girish Sastry, Amanda Askell, Pamela Mishkin, Jack Clark,
  et~al.
\newblock Learning transferable visual models from natural language
  supervision.
\newblock In \emph{ICML}, pages 8748--8763. PMLR, 2021.

\bibitem[Rohrbach et~al.(2015)Rohrbach, Rohrbach, Regneri, Amin, Andriluka,
  Pinkal, and Schiele]{rohrbach15ijcv}
Marcus Rohrbach, Anna Rohrbach, Michaela Regneri, Sikandar Amin, Mykhaylo
  Andriluka, Manfred Pinkal, and Bernt Schiele.
\newblock Recognizing fine-grained and composite activities using hand-centric
  features and script data.
\newblock \emph{IJCV}, pages 1--28, 2015.
\newblock ISSN 0920-5691.

\bibitem[Scovanner et~al.(2007)Scovanner, Ali, and Shah]{sift_3d}
Paul Scovanner, Saad Ali, and Mubarak Shah.
\newblock {A 3-Dimentional SIFT Descriptor and its Application to Action
  Recognition}.
\newblock \emph{CRCV}, pages 1--4, 2007.

\bibitem[Shao et~al.(2020)Shao, Zhao, Dai, and Lin]{shao2020finegym}
Dian Shao, Yue Zhao, Bo~Dai, and Dahua Lin.
\newblock Finegym: A hierarchical video dataset for fine-grained action
  understanding.
\newblock In \emph{CVPR}, 2020.

\bibitem[Sherstinsky(2020)]{sherstinsky2020fundamentals}
Alex Sherstinsky.
\newblock Fundamentals of recurrent neural network (rnn) and long short-term
  memory (lstm) network.
\newblock \emph{Physica D: Nonlinear Phenomena}, 404:\penalty0 132306, 2020.

\bibitem[Simonyan and Zisserman(2014)]{two_stream}
Karen Simonyan and Andrew Zisserman.
\newblock Two-stream convolutional networks for action recognition in videos.
\newblock In \emph{NIPS}, pages 568--576, 2014.

\bibitem[Srivastava et~al.(2014)Srivastava, Hinton, Krizhevsky, Sutskever, and
  Salakhutdinov]{JMLR:v15:srivastava14a}
Nitish Srivastava, Geoffrey Hinton, Alex Krizhevsky, Ilya Sutskever, and Ruslan
  Salakhutdinov.
\newblock Dropout: A simple way to prevent neural networks from overfitting.
\newblock \emph{JMLR}, 15\penalty0 (56):\penalty0 1929--1958, 2014.

\bibitem[Sultani et~al.(2018)Sultani, Chen, and Shah]{sultani2018real}
Waqas Sultani, Chen Chen, and Mubarak Shah.
\newblock Real-world anomaly detection in surveillance videos.
\newblock In \emph{CVPR}, pages 6479--6488, 2018.

\bibitem[Tang et~al.(2023)Tang, Bi, Xu, Song, Liang, Wang, Zhang, An, Lin, Zhu,
  Vosoughi, Huang, Zhang, Zheng, Zhang, Luo, Luo, and Xu]{vidllmsurvey}
Yunlong Tang, Jing Bi, Siting Xu, Luchuan Song, Susan Liang, Teng Wang, Daoan
  Zhang, Jie An, Jingyang Lin, Rongyi Zhu, Ali Vosoughi, Chao Huang, Zeliang
  Zhang, Feng Zheng, Jianguo Zhang, Ping Luo, Jiebo Luo, and Chenliang Xu.
\newblock Video understanding with large language models: A survey.
\newblock \emph{arXiv preprint arXiv:2312.17432}, 2023.

\bibitem[Tran et~al.(2015)Tran, Bourdev, Fergus, Torresani, and
  Paluri]{spattemp_filters}
Du~Tran, Lubomir Bourdev, Rob Fergus, Lorenzo Torresani, and Manohar Paluri.
\newblock {Learning Spatiotemporal Features with {3D} Convolutional Networks}.
\newblock \emph{ICCV}, pages 4489--4497, 2015.

\bibitem[Vaswani et~al.(2017)Vaswani, Shazeer, Parmar, Uszkoreit, Jones, Gomez,
  Kaiser, and Polosukhin]{vaswani2017attention}
Ashish Vaswani, Noam Shazeer, Niki Parmar, Jakob Uszkoreit, Llion Jones,
  Aidan~N Gomez, {\L}ukasz Kaiser, and Illia Polosukhin.
\newblock Attention is all you need.
\newblock \emph{NeurIPS}, 30, 2017.

\bibitem[Wang and Schmid(2013)]{improved_traj}
Heng Wang and Cordelia Schmid.
\newblock {Action Recognition with Improved Trajectories}.
\newblock \emph{ICCV}, pages 3551--3558, 2013.

\bibitem[Wang and Koniusz(2023)]{wang20233mformer}
Lei Wang and Piotr Koniusz.
\newblock 3mformer: Multi-order multi-mode transformer for skeletal action
  recognition.
\newblock In \emph{CVPR}, pages 5620--5631, 2023.

\bibitem[Wang and Koniusz(2024)]{wang2024flow}
Lei Wang and Piotr Koniusz.
\newblock Flow dynamics correction for action recognition.
\newblock In \emph{ICASSP}, pages 3795--3799. IEEE, 2024.

\bibitem[Wang et~al.(2020)Wang, Huynh, and Koniusz]{lei_tip_2019}
Lei Wang, Du~Q. Huynh, and Piotr Koniusz.
\newblock A comparative review of recent kinect-based action recognition
  algorithms.
\newblock \emph{IEEE TIP}, 29:\penalty0 15--28, 2020.
\newblock ISSN 1941-0042.

\bibitem[Wang et~al.(2024)Wang, Yuan, Gedeon, and Zheng]{wang2024taylor}
Lei Wang, Xiuyuan Yuan, Tom Gedeon, and Liang Zheng.
\newblock Taylor videos for action recognition.
\newblock In \emph{ICML}, 2024.

\bibitem[Wang et~al.(2018)Wang, Xiong, Wang, Qiao, Lin, Tang, and
  Van~Gool]{wang2018temporal}
Limin Wang, Yuanjun Xiong, Zhe Wang, Yu~Qiao, Dahua Lin, Xiaoou Tang, and Luc
  Van~Gool.
\newblock Temporal segment networks for action recognition in videos.
\newblock \emph{IEEE TPAMI}, 41\penalty0 (11):\penalty0 2740--2755, 2018.

\bibitem[Wang et~al.(2021)Wang, Tong, Ji, and Wu]{wang2021tdn}
Limin Wang, Zhan Tong, Bin Ji, and Gangshan Wu.
\newblock Tdn: Temporal difference networks for efficient action recognition.
\newblock In \emph{CVPR}, pages 1895--1904, 2021.

\bibitem[Wei et~al.(2020)Wei, Zhang, Li, Zhang, and Wu]{Wei_2020_CVPR}
Xi~Wei, Tianzhu Zhang, Yan Li, Yongdong Zhang, and Feng Wu.
\newblock Multi-modality cross attention network for image and sentence
  matching.
\newblock In \emph{CVPR}, June 2020.

\bibitem[Wu et~al.(2019)Wu, Li, Hsieh, and Sharpnack]{wu2019stochastic}
Liwei Wu, Shuqing Li, Cho-Jui Hsieh, and James~L Sharpnack.
\newblock Stochastic shared embeddings: Data-driven regularization of embedding
  layers.
\newblock \emph{NeurIPS}, 32, 2019.

\bibitem[Yang et~al.(2022)Yang, Miech, Sivic, Laptev, and Schmid]{yang2022zero}
Antoine Yang, Antoine Miech, Josef Sivic, Ivan Laptev, and Cordelia Schmid.
\newblock Zero-shot video question answering via frozen bidirectional language
  models.
\newblock \emph{NeurIPS}, 35:\penalty0 124--141, 2022.

\bibitem[Yang et~al.(2024)Yang, Wang, Li, Wang, and Yang]{yang2024fine}
Lingfeng Yang, Yueze Wang, Xiang Li, Xinlong Wang, and Jian Yang.
\newblock Fine-grained visual prompting.
\newblock \emph{NeurIPS}, 36, 2024.

\bibitem[Zhu et~al.(2021)Zhu, Su, Lu, Li, Wang, and Dai]{zhu2021deformable}
Xizhou Zhu, Weijie Su, Lewei Lu, Bin Li, Xiaogang Wang, and Jifeng Dai.
\newblock Deformable {DETR}: Deformable transformers for end-to-end object
  detection.
\newblock In \emph{ICLR}, 2021.

\bibitem[Zhuang et~al.(2020)Zhuang, Qi, Duan, Xi, Zhu, Zhu, Xiong, and
  He]{zhuang2020comprehensive}
Fuzhen Zhuang, Zhiyuan Qi, Keyu Duan, Dongbo Xi, Yongchun Zhu, Hengshu Zhu, Hui
  Xiong, and Qing He.
\newblock A comprehensive survey on transfer learning.
\newblock \emph{Proceedings of the IEEE}, 109\penalty0 (1):\penalty0 43--76,
  2020.

\end{thebibliography}
